\documentclass[11pt]{article}

\usepackage[final]{acl}

\usepackage{times}
\usepackage{latexsym}
\usepackage[T1]{fontenc}
\usepackage[utf8]{inputenc}
\usepackage{microtype}
\usepackage{inconsolata}
\usepackage{graphicx}
\usepackage[font=small,labelfont=bf]{subcaption}

\usepackage{amsmath,amssymb,amsfonts}
\usepackage{tikz}
\usetikzlibrary{shapes.geometric,arrows,positioning}
\usepackage{pgfplots}
\pgfplotsset{compat=1.18}
\usepackage{xcolor}
\usepackage{colortbl}
\usepackage[normalem]{ulem}
\usepackage{tabularx}
\usepackage{array}
\usepackage{booktabs}
\usepackage{makecell}
\usepackage{multirow}
\usepackage{pifont}
\usepackage{siunitx}
\usepackage{threeparttable}
\usepackage{algorithm}
\usepackage{algpseudocode}

\usepackage[acronym, nomain, nopostdot]{glossaries}
\glsdisablehyper   
\makeglossaries
\newacronym{LLM}{LLM}{Large Language Model}
\newacronym{ADS}{ADS}{Autonomous Driving System}
\newacronym{AD}{AD}{Autonomous Driving}
\newacronym{NLP}{NLP}{Natural Language Processing}
\newacronym{VLM}{VLM}{Vision Language Model}
\newacronym{DSL}{DSL}{Domain Specific Language}
\newacronym{MPC}{MPC}{Model Predictive Control}
\newacronym{RAG}{RAG}{Retrieval-Augmented Generation}
\newacronym{OOD}{OOD}{Out-of-Distribution}
\newacronym{SOTIF}{SOTIF}{Safety of the Intended Functionality}
\newacronym{BFS}{BFS}{Breadth-First Search}
\newacronym{OSM}{OSM}{OpenStreetMap}
\newacronym{CoT}{CoT}{Chain-of-Thought}
\newacronym{ICL}{ICL}{In-Context Learning}
\newacronym{CP}{CP}{Contextual Prompt}
\newacronym{GT}{GT}{Ground Truth}
\newacronym{NL}{NL}{Natural Language}
\newacronym{qa}{QA}{Question Answering}

\newcommand{\xmark}{\ding{55}}

\newif\ifshownew
\shownewfalse

\DeclareRobustCommand{\old}[1]{}
\newif\ifarxiv
\arxivtrue
\ifarxiv\usepackage{stfloats}\fi
\newcommand{\arxivnotice}{%
  \ifarxiv
  \begin{table*}[b]
    \centering
    \fbox{\begin{minipage}{0.965\textwidth}
      \small
      Accepted to EMNLP~2026 (Main Conference). This is the authors' preprint
      version. The version of record will appear in the ACL Anthology under a
      CC~BY~4.0 licence.
    \end{minipage}}
  \end{table*}
  \fi
}

\tikzstyle{block}  = [rectangle, rounded corners, minimum width=3cm, minimum height=1cm, text centered, draw=black, fill=blue!20]
\tikzstyle{input}  = [ellipse, minimum width=2cm, minimum height=1cm, text centered, draw=black, fill=green!20]
\tikzstyle{output} = [ellipse, minimum width=2cm, minimum height=1cm, text centered, draw=black, fill=red!20]
\tikzstyle{arrow}  = [thick,->,>=stealth]

\title{PlannerForge: LLM Agents for Scenario-Based Testing of Motion Planners in Autonomous Driving\thanks{Code and data: {\url{https://github.com/TUM-AVS/PlannerForge}}}}

\author{
  \textbf{Yuan Gao\textsuperscript{1}} \quad
  \textbf{Sebastian M\"uller\textsuperscript{1}} \quad
  \textbf{Mattia Piccinini\textsuperscript{1}} \quad
  \textbf{Marc Kaufeld\textsuperscript{1}} \\
  \textbf{Yuchen Zhang\textsuperscript{1}} \quad
  \textbf{Finn Rasmus Sch\"afer\textsuperscript{1}} \quad
  \textbf{Qunying Song\textsuperscript{2}} \quad
  \textbf{Johannes Betz\textsuperscript{1}} \\[2pt]
  \textsuperscript{1}Professorship of Autonomous Vehicle Systems, TUM School of Engineering and Design, \\
  Technical University of Munich, 85748 Garching, Germany; \\
  Munich Institute of Robotics and Machine Intelligence (MIRMI) \\
  \textsuperscript{2}University College London, London, United Kingdom
}

\begin{document}
\maketitle
\arxivnotice

\begin{abstract}
Ensuring the safety of autonomous driving is a critical challenge. Scenario-based testing is a systematic process used to validate \glspl{ADS}, but it remains a fragmented modular pipeline in which scenario generation, retrieval, modification, \gls{ADS} execution, and results analysis are performed by separate tools with little interaction. \gls{LLM} agents have shown promise across \gls{ADS} sub-systems such as perception, planning, and control. However, no prior work covers the whole scenario-based testing pipeline for \glspl{ADS} with a unified \gls{LLM}-agent framework.
We present \textbf{PlannerForge}, an \gls{LLM}-agent framework that extends all scenario-based testing stages (from Scenario Generation to \gls{ADS} Assessment) and adds two further \gls{LLM}-enhanced stages: \gls{ADS} Enhancement and \gls{ADS} Benchmarking. We evaluate PlannerForge with 10 off-the-shelf \glspl{LLM} across all tasks (Generation, Selection, Modification, Module Routing, Planner Testing, and Enhancement) under 5 prompt conditions. Best-per-task scores range from 0.88 to 1.00, and open-source 20--35B backends match commercial APIs on most tasks. Open-source models such as Qwen3.6:35B match commercial APIs on three of the five tasks. {} {Chaining the modules end-to-end retains 83\% / 78\% of seed queries (commercial / open). It outperforms Scenario Factory~2.0 \citep{finkeldei2025scenariofactory} on natural-language generation (193 vs.\ 144 executable of 200) and realises 92--96\% of requested city, road and vehicle attributes. It outperforms BM25 \citep{robertson2009bm25} at rank~1 selection (92.0\% vs.\ 67.5\%) and From-Words-to-Collisions \citep{gao2025words} on physically valid edits ($\geq$94\% vs.\ 31\%). At $N{=}400$, cost-tuning lifts planner success from 50.4\% to 70.2\% and cuts collisions from 19.0\% to 8.4\%, without domain-specific fine-tuning.}
\end{abstract}

\section{Introduction}
The rapid advancement of \acrfullpl{ADS} to SAE Level 4~\cite{Waymo2018, SAE2021} hinges on rigorous validation~\cite{Betz2024}. Because real-world testing of rare edge cases is prohibitively expensive~\cite{pegasus}, the industry relies heavily on scenario-based testing in simulation~\cite{riedmaier2020survey, Song2024}. While recent advances in \acrfullpl{LLM} have begun to enhance the realism and scalability of such testing, their application has been largely confined to scenario generation~\cite{gao2026foundation}. However, an \gls{ADS} assessment pipeline requires significantly more: it must seamlessly integrate the selection of relevant cases, the execution of tests, and the analysis of results~\cite{song2026generative}.

\begin{figure}[t]
    \centering
    \includegraphics[width=1\linewidth]{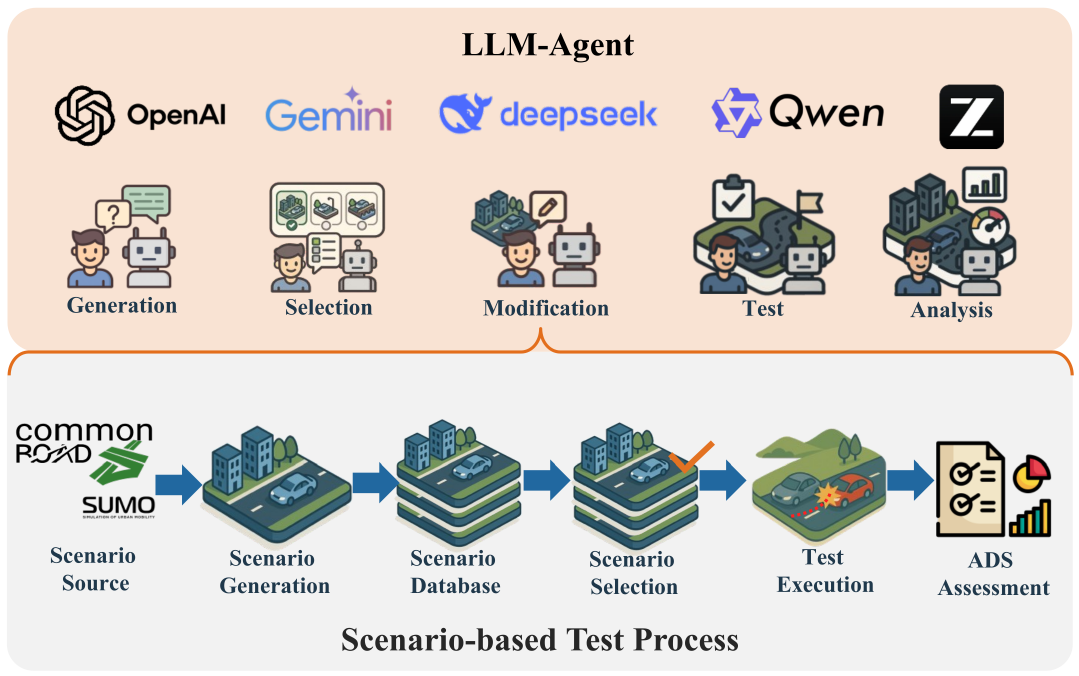}
    \captionsetup{justification=centering,singlelinecheck=false}
    \caption{\textbf{PlannerForge} overview: \gls{LLM} agents automate the scenario-based testing pipeline.}
    \label{fig:concept}
\end{figure}

In current motion-planner testing practice, these stages remain highly fragmented and manual. Scenario generation relies on GUI editors or scripts, selection depends on hand-crafted database filters, and ad-hoc scenario modification is largely unsupported. Furthermore, testing pipelines are rigidly scripted rather than guided by user intent, planner cost weights are manually tuned, and cross-planner comparisons require separately scripted batch runs with offline aggregation. To overcome these bottlenecks, there is a clear need for a unified, \gls{LLM}-powered framework that spans and automates the entire lifecycle of scenario-based testing.

In this paper, we introduce \textbf{PlannerForge}, an LLM-powered scenario-based testing framework for motion planners that closes this gap (Figure~\ref{fig:concept}). We target motion planners because they are the core decision-making module of an \gls{ADS} and their behavior is particularly sensitive to complex, safety-critical traffic scenarios. Unlike prior work that focuses narrowly on scenario generation, PlannerForge integrates the full testing pipeline: scenario \emph{generation} from real-world \gls{OSM}\footnote{\url{https://www.openstreetmap.org/}} maps, scenario retrieval from a curated open-source scenario database, scenario modification via traffic object and behavior changes, motion planner execution, and performance analysis. A chatbot-oriented interface abstracts away technical complexity while enabling the seamless integration of diverse motion planners, providing researchers and practitioners with a scalable, benchmark-ready platform for \gls{ADS} assessment.

The key contributions of this paper are:
\begin{enumerate}
    \setlength{\itemsep}{2pt}
    \setlength{\parsep}{0pt}
    \item \textbf{PlannerForge}, to our knowledge, the first {}{full-lifecycle} LLM framework that unifies all scenario-based-testing stages~\cite{riedmaier2020survey} (Scenario Source, Generation, Database, Selection, Test Execution, \gls{ADS} Assessment; together with two further \gls{LLM}-era stages, \gls{ADS} Enhancement and \gls{ADS} Benchmarking{)} in a single chatbot-driven pipeline.
    \item An \textbf{empirical evaluation} of ten off-the-shelf \glspl{LLM} backends (five commercial, five open-source, spanning reasoning and non-reasoning models) across the five core framework tasks under five prompt conditions, showing both module-level performance and the effectiveness of off-the-shelf \glspl{LLM} agents without domain-specific fine-tuning.
    \item \leavevmode{A \textbf{unified multi-planner interface} for comparative evaluation on generated and modified scenarios.}{}{}
\end{enumerate}

\section{Related Work}
\label{sec:related-work}
Scenario-based testing provides a systematic methodology for validating \glspl{ADS} by structurally evaluating operational conditions and safety-critical situations. This section reviews classical and LLM-powered approaches and positions \textbf{PlannerForge} within this landscape.

\subsection{Classical Scenario-based Testing}
Industry initiatives such as Pegasus~\cite{pegasus} and SAKURA~\cite{nakamura2022defining}, alongside foundational surveys~\cite{riedmaier2020survey}, have established an influential six-component taxonomy for scenario-based testing: (1) \emph{Scenario Source}, (2) \emph{Scenario Generation}, (3) \emph{Scenario Database}, (4) \emph{Scenario Selection}, (5) \emph{Test Execution}, and (6) \emph{ADS Assessment}.
Prior literature has extensively explored these individual components.
For \emph{Scenario Generation}, research covers {}{knowledge-driven and data-driven approaches~\cite{nalic2020scenario} as well as adversarial and deep generative methods~\cite{ding2023survey}}.
Work on \emph{Scenario Databases} includes {}reviews comparing dataset sensor modalities and annotations~\cite{ding2023survey}.
\emph{Scenario Selection} strategies typically involve knowledge-driven, data-driven, or falsification-based prioritization{}~\cite{riedmaier2020survey}.
Finally, comprehensive surveys have examined {}{scenario-based accelerated testing for ISO~21448 \gls{SOTIF}\footnote{\url{https://www.iso.org/standard/77490.html}}~\cite{tang2024scenario} and \emph{ADS Assessment} through on-road performance metrics~\cite{sharath2021literature}}.

\subsection{LLM-powered Scenario-based Testing}
With the emergence of \glspl{LLM}, the scenario-based testing process has been augmented with their reasoning capabilities across generation, analysis, and downstream execution stages.

\textbf{LLM-powered Scenario Generation:} Existing systems are split along simulator class {}{and input source}. \emph{Autonomous Driving Simulation} (CARLA~\cite{dosovitskiy2017carla}{}): ChatScene~\cite{zhang2024chatscene}, TTSG~\cite{ruan2024traffic}, Aasi et al.~\cite{aasi2024generating}, NL2Scenic~\cite{bauerfeind2025david}, {Chat2Scenic~\cite{gao2026chat2scenic},} Petrovic et al.~\cite{petrovic2024llm}, and Text2Scenario~\cite{cai2026text2scenario} synthesise safety-critical or branching \gls{OOD} scenarios from natural-language prompts{; LCTGen~\cite{tan2023language} generates language-conditioned traffic on real maps}. \emph{Crash-report reconstruction}: {}{SoVAR~\cite{guo2024sovar} and LeGEND~\cite{tang2024legend} recover simulator assets from accident reports. \emph{Traffic rules and datasets}: TARGET~\cite{deng2023target} compiles traffic rules into a \gls{DSL}; Chat2Scenario~\cite{zhao2024chat2scenario} extracts scenarios from naturalistic logs. \emph{Adversarial generation}: LLM-attacker~\cite{mei2025llm} optimises attacker trajectories in closed loop}. \emph{Traffic flow Simulation} (SUMO~\cite{SUMO2018}): ChatSUMO~\cite{li2024chatsumo} couples \glspl{LLM} with OSM~\cite{haklay2008openstreetmap} {}{import scripts}, while LLMScenario~\cite{chang2024llmscenario} composes safety-critical HighD~\cite{Krajewski2018highd} trajectories in MetaScenario~\cite{chang2022metascenario} through in-context demonstrations.

\textbf{LLM-powered \gls{ADS} Enhancement:} Recent research integrates \glspl{LLM} into autonomous driving systems {}{as planners or controllers. The Language-Agent line of work~\cite{mao2023language} is a tool-using LLM decision agent for the planner, while MPC$\times$\gls{LLM}~\cite{baumann2025enhancing} is a Model Predictive Control parameter tuner that adapts costs and constraints from natural-language context while preserving the underlying optimization. DualAD~\cite{wang2024dualad} overlays an LLM reasoning layer that issues speed decisions from textual scene encodings, while LeAD~\cite{zhang2025lead} employs a dual-rate architecture in which low-frequency \glspl{LLM} modules supplement high-frequency end-to-end systems in challenging scenarios via chain-of-thought reasoning}.

Recent surveys of \glspl{LLM} in \gls{ADS} testing and scenario generation~\cite{song2026generative, gao2026foundation} confirm that most reviewed papers focus on scenario generation{}{}. 

\subsection{Critical Summary}
Across these works, prior \gls{LLM}-powered systems remain highly fragmented, typically focusing in isolation on either \emph{Scenario Generation} or \emph{ADS Enhancement}. The critical research gap is the absence of a comprehensive, full-pipeline framework for scenario-based testing of \glspl{ADS}. To close this gap, \textbf{PlannerForge} unifies the classic six-component taxonomy~\cite{riedmaier2020survey} into a single {}{full-lifecycle} framework and extends it with two further stages: \emph{ADS Enhancement} (LLM-guided planner tuning) and \emph{ADS Benchmarking} (cross-planner comparative evaluation under shared scenarios).

\section{Problem Formulation}
\label{sec:problem}

We formalize \gls{LLM}-powered scenario-based testing as a sequence of language-to-structured-output decisions. Let $\mathcal{U}$ be the space of natural-language utterances, $\mathcal{S}$ that of 2D scenarios produced by an open-source motion-planning simulator, $\mathcal{D}\subseteq\mathcal{S}$ a curated database, $\Theta$ the space of motion-planner configurations, $\mathcal{H}$ conversation histories, $\mathcal{O}$ execution outcomes, and $\mathcal{Y}$ natural-language analyses. At dialogue turn $t$ the agent observes $x_t = (u_t, s_t, \theta_t, h_t)$, where $u_t \in \mathcal{U}$, $s_t \in \mathcal{S}$, $\theta_t \in \Theta$, and $h_t \in \mathcal{H}$.

A session begins with Generation or Selection to populate the initial scenario $s_0$:
\begin{equation}
  f_{\mathrm{gen}}\!:\mathcal{U}\to\mathcal{S} \quad \text{or} \quad f_{\mathrm{sel}}\!:\mathcal{U}\times\mathcal{D}\to\mathcal{D}
\end{equation}
Subsequent turns are dispatched by the Module Router, which at a high level selects the next phase in the scenario-based testing pipeline based on the user prompt $u_t$ and dialogue history $h_t$. Formally, it acts as an intent classifier predicting $\hat{a}_t = \arg\max_{a\in\mathcal{A}} f_{\mathrm{router}}(a\mid u_t, h_t)$ over $\mathcal{A} = \{\textsc{modify}, \textsc{tune}, \textsc{test}, \textsc{analyse}, \textsc{\gls{qa}}\}$ (the additional \textsc{qa} action returns a free-form answer without invoking any downstream operator; see \S\ref{sec:method}). The router then invokes the corresponding action-conditional maps: 
\begin{align}
  f_{\mathrm{mod}} &\!:\,\mathcal{S}\times\mathcal{U}\to\mathcal{S},\;\; & \text{where output } s'\models\Sigma_{\mathcal{D}} \notag\\
  f_{\mathrm{tune}} &\!:\,\Theta\times\mathcal{U}\to\Theta,\;\; & \text{where output } \theta'\models\Sigma_\Theta \notag\\
  f_{\mathrm{test}} &\!:\,\mathcal{S}\times\Theta\to\mathcal{O} \notag\\
  f_{\mathrm{eval}} &\!:\,\mathcal{O}\times\mathcal{U}\to\mathcal{Y} \notag
\end{align}
where $\Sigma_{\mathcal{D}}$ is the curated scenario XML dataset and $\Sigma_\Theta$ the planner configuration. Crucially, $f_{\mathrm{gen}}$, $f_{\mathrm{mod}}$, and $f_{\mathrm{tune}}$ are \emph{constrained generators}: their outputs (denoted $s'$ and $\theta'$ above) must satisfy these respective schemas. Producing schema-conformant XML is the central linguistic challenge.

\section{Methodology}
\label{sec:method}
\begin{figure*}[ht]
    \centering
    \includegraphics[width=0.99\linewidth]{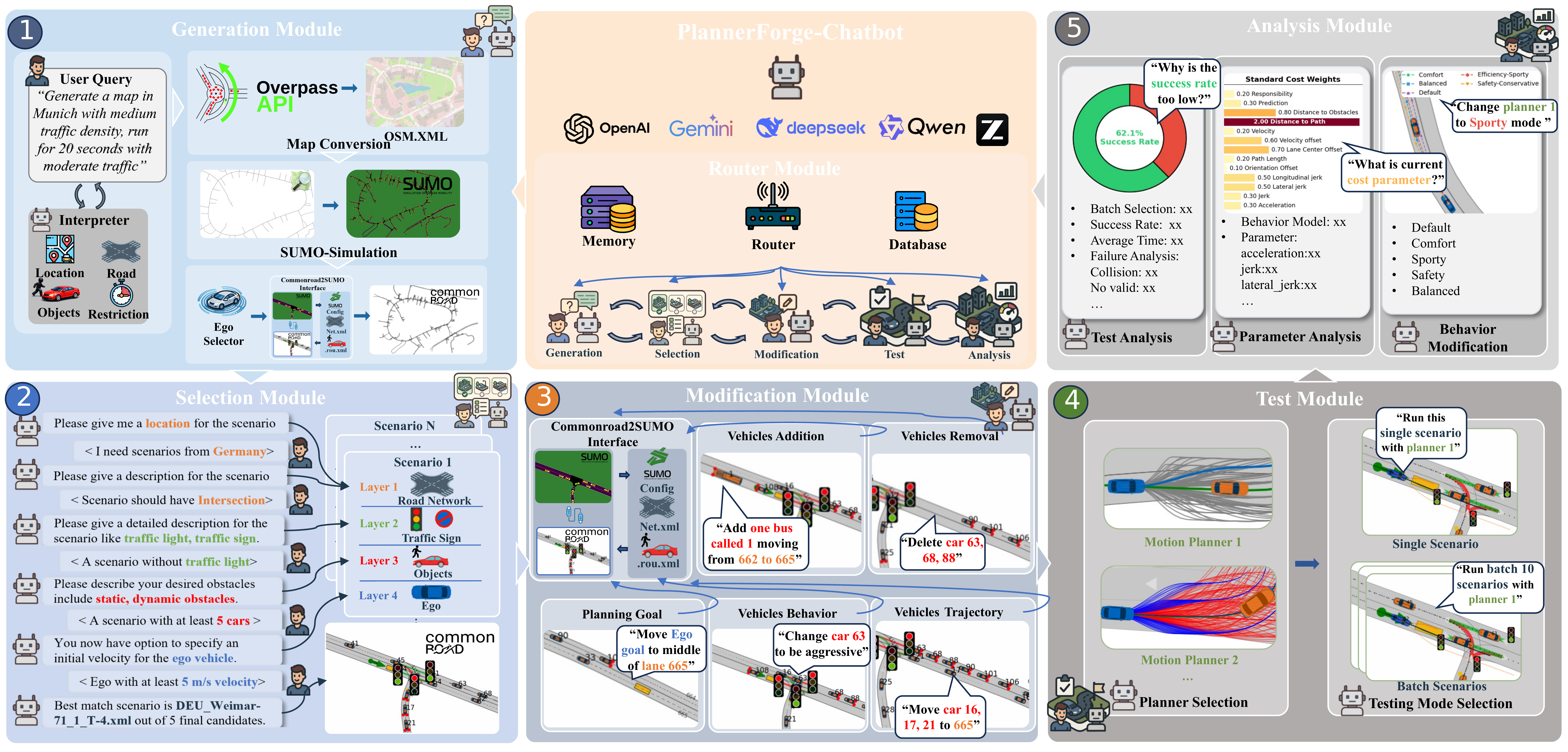}
    \caption{PlannerForge framework with six modules: \textbf{Router} (classifies user intent post-selection), \textbf{Generation} (OSM+SUMO synthesis), \textbf{Selection} (dialogue-guided retrieval from the CommonRoad DB), \textbf{Modification} (LLM-guided SUMO edits), \textbf{Testing} (Frenetix / MP-RBFN execution), and \textbf{Analysis} (LLM-powered result interpretation).}
    \label{fig:framework}
\end{figure*}
\textbf{PlannerForge} is a unified framework that integrates \glspl{LLM} across the entire scenario-based testing pipeline for motion planners, as illustrated in Figure~\ref{fig:framework}. The six modules summarised in the caption (\emph{Generation}, \emph{Selection}, \emph{Module Router}, \emph{Modification}, \emph{Testing}, and \emph{Analysis}) are detailed in the subsections below; the \emph{Module Router} (\S\ref{subsec:module-router}) acts as the intent dispatcher that unlocks flexible post-selection navigation.

\subsection{Framework Setup}
\begin{figure}[t]
    \centering
    \includegraphics[width=0.99\linewidth]{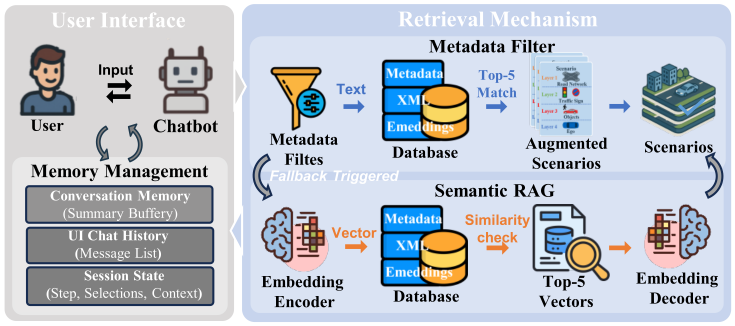}
    \caption{Chatbot front-end of PlannerForge. The Gradio UI exposes the natural-language query box, session state, conversation memory, and scenario retrieval mechanism. The LangChain backend routes user utterances to the six modules of Figure~\ref{fig:framework}.}
    \label{fig:query}
\end{figure}

The PlannerForge framework features a chatbot interface (Figure~\ref{fig:query}) built with a Gradio\footnote{\url{https://gradio.app/}} frontend and a LangChain\footnote{\url{https://www.langchain.com/}} backend. To support coherent multi-turn interactions, it manages state across three levels: \textit{Conversational Memory} (LangChain retains recent exchanges and summarises history exceeding 125k tokens), \textit{UI Chat History} (Gradio maintains an unmodified visual log of the conversation), and \textit{Session State} (in-RAM storage for user-specific context and intermediate module outputs).

\textbf{Scenario Database:} Open-source driving scenarios from CommonRoad~\cite{althoff2017commonroad} are stored as XML files augmented with a structured \texttt{<Metadata>} element covering four scenario layers (location, roadside constructs, participants, ego vehicle) and indexed in a Chroma vector database~\cite{chroma}. Further implementation details and the exact metadata schema are provided in Appendix~\ref{app:impl}.

\begin{figure}[t]
\centering
\includegraphics[width=0.99\linewidth]{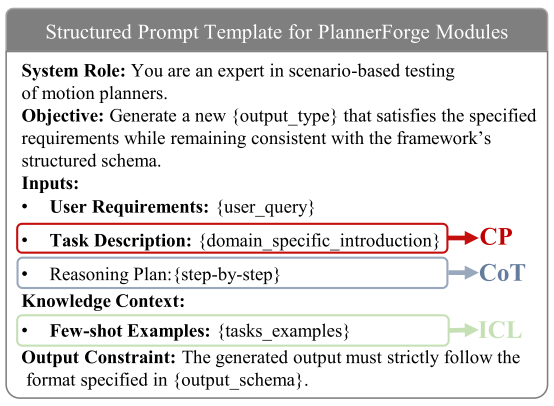}
\caption{Structured prompt template shared across PlannerForge modules.}
\label{fig:prompt-template}
\end{figure}

\textbf{Prompting Techniques.} All modules in the framework implement the following prompting techniques (Figure~\ref{fig:prompt-template}), so that pretrained \glspl{LLM} can be adjusted to our specific tasks~\cite{gao2026foundation}:
\textit{\gls{CP}} injects the structured output schema, syntactic constraints, and available operators into the prompt. For instance, the Generation module receives the JSON intent schema with required keys location, road classes, density, vehicle mix, and duration.
\textit{\gls{CoT}} structures generation into explicit reasoning steps per module. The Modification scaffold reads: identify target, enumerate route changes, preserve connectivity, and emit the SUMO edit.
\textit{\gls{ICL}} adds a few-shot demonstration examples: positive natural-language to output pairs, plus, where applicable, negative refusal examples that anchor edge-case behavior.

We denote the prompt used by module $M$ as $\mathbf{P}_{M}$ (e.g.\ $\mathbf{P}_{\textsc{gen}}, \mathbf{P}_{\textsc{sel}}, \mathbf{P}_{\textsc{router}}, \mathbf{P}_{\textsc{mod}}, \mathbf{P}_{\textsc{tune}}, \mathbf{P}_{\textsc{eval}}$); full prompts for each module are released with the code (Appendix~\ref{app:overview-prompts}).

\subsection{Module Router}
\label{subsec:module-router}
Traditional testing frameworks follow a rigid \emph{generation $\to$ selection $\to$ modification $\to$ testing $\to$ analysis} workflow. PlannerForge breaks this linearity through the \textbf{Module Router}. Following the formalisation in \S\ref{sec:problem}, after the initial scenario generation or selection, the router acts as the intent classifier $f_{\mathrm{router}}(a \mid u_t, h_t)$, mapping the user utterance $u_t$ to an action $\hat{a}_t \in \mathcal{A}$. These actions correspond to five categories: \textit{Scenario Modification} (\textsc{modify}), \textit{Parameter Tuning} (\textsc{tune}), \textit{Test Execution} (\textsc{test}), \textit{Result Analysis} (\textsc{analyse}), and \textit{General Question \& Answer} (\textsc{qa}). This is implemented via two-stage \gls{LLM} function calling: In the first stage, guided by the router prompt $P_{\text{ROUTER}}$, the LLM identifies the corresponding module and extracts the required arguments \texttt{args}, returning both as a structured JSON object. The second stage's Process Engine dispatches \texttt{args} to the corresponding module. Full dispatch pseudocode is given in Algorithm~\ref{alg:router} (Appendix~\ref{app:router-dispatch}). The router is the key architectural mechanism that distinguishes PlannerForge from prior LLM-assisted testing tools, and the per-module implementations are detailed in the following subsections.

\subsection{Scenario Generation Module}
\label{subsec:generation}

When database scenarios are insufficient, PlannerForge generates new CommonRoad scenarios from scratch via a two-stage pipeline. An \gls{LLM} parses natural-language requests into structured intents, combining real-world road topologies with procedurally simulated traffic.

\textbf{Stage 1: Map and traffic synthesis.} The user describes the desired scenario in natural language (e.g., ``Munich intersection with light traffic, focus on a turning truck''). An \gls{LLM} parses this with prompt $\mathbf{P}_{\textsc{gen}}$ into a structured JSON intent specifying location (city or bounding box), drivable road classes, traffic density, vehicle mix, and simulation duration. The bounding box drives an OpenStreetMap query via the Overpass API~\cite{haklay2008openstreetmap}; the returned road network is simulated in SUMO~\cite{SUMO2018} and then converted to CommonRoad~\cite{althoff2017commonroad} format. A microscopic SUMO simulation populates the network with vehicles, trucks, buses, and other configurable actor types, producing trajectories that respect car-following and lane-changing dynamics.

\textbf{Stage 2: Planning problem synthesis.} From the populated scenario, the user selects an ego vehicle and a goal region; the \gls{LLM} may also suggest an ego candidate using strategies such as \emph{first car}, \emph{by type}, or \emph{by index}. A planning problem is then synthesized by attaching an initial state (the ego's current pose) and a goal region (either a chosen lanelet or a forward offset along the ego's trajectory). The result is saved as a standard CommonRoad scenario file ready for downstream modification, testing, and analysis.

\subsection{Scenario Selection Module}
The Scenario Selection Module facilitates the retrieval of test cases from large-scale databases by abstracting low-level representations into an \gls{LLM}-guided natural-language dialogue.

We structure retrieval around the layer-based taxonomy introduced by Riedmaier et al.~\cite{riedmaier2020survey}, indexing scenarios across four metadata layers: \emph{location}, \emph{roadside constructs} (split into the \texttt{tags} and \texttt{road\_net} extractors below), \emph{participants}, and \emph{ego vehicle}. During a five-step dialogue (Figure~\ref{fig:framework}), the \gls{LLM} extracts a structured slot $\hat{\ell}_k$ from the user's utterance at step $k$ (geographical codes, discrete keywords, kinematic ranges) using a per-slot prompt $\mathbf{P}_{\textsc{sel}}^{(k)}$. Let $\mathcal{D}^{(0)} = \mathcal{D}$ denote the full database. For $k = 1,\ldots,5$ corresponding to (\texttt{location}, \texttt{tags}, \texttt{road\_net}, \texttt{obstacles}, \texttt{velocity}),
\begin{equation}
  \mathcal{D}^{(k)} = \mathcal{D}^{(k-1)} \cap \texttt{match}(\hat{\ell}_k),
  \label{eq:sel-filter}
\end{equation}
monotonically pruning the candidate set. The module returns $\text{top-}k(\mathcal{D}^{(5)})$ when non-empty, and otherwise falls back to a SentenceTransformer~\cite{reimers2019sentence} semantic-similarity search over $\mathcal{D}$ keyed by the concatenated dialogue $u_{1:5}$, guaranteeing retrieval by contextual meaning when exact metadata matches fail. Ablation results for the retrieval pipelines are detailed in Appendix~\ref{app:sel-results}.

\subsection{Scenario Modification Module}
CommonRoad scenarios encode fixed pre-recorded trajectories. To make edits tractable for the \gls{LLM}, we route them through the CommonRoad--SUMO interface~\cite{sumocr}: scenarios are converted to a \texttt{.net.xml} (network topology) plus \texttt{.vehicles.rou.xml} (routes and behaviour) pair, the \gls{LLM} (invoked with a per-task prompt $\mathbf{P}_{\textsc{mod}}^{\tau}$ for $\tau\in\{T, B, P, G\}$) emits a modified SUMO file, and the round-trip back to CommonRoad produces kinematically feasible trajectories. We support four edit categories: \\
(1) \textbf{Trajectory (T)} modifications redirect vehicles by updating edge sequences (two-stage prompt: the network topology is summarised into valid routes, then the edit is generated against the route file);\\
(2) \textbf{Behaviour (B)} modifications swap each vehicle's \texttt{<vType>} against six car-following presets (\textit{Aggressive}, \textit{Cautious}, \textit{Emergency}, \textit{Eco}, \textit{Balanced}, \textit{Speeder}) while preserving \texttt{vClass}; \\
(3) \textbf{Population (P)} modifications add or remove vehicle entries with type, departure, and valid routes derived from the topology summary, as shown in Figure~\ref{fig:mod-qualitative-grid} (Appendix~\ref{app:mod-results}) with vehicle removal and addition examples; \\
(4) \textbf{Goal (G)} modifications edit the planning problem by updating the goal region of the Ego Vehicle in place without a SUMO round-trip.

\subsection{Planner Testing and Enhancement Module}
This module implements the test executor $f_{\mathrm{test}}$ and parameter tuner $f_{\mathrm{tune}}$. 
The executor $f_{\mathrm{test}}$ abstracts the simulation environment, parameter parsing, and logging via a unified interface:
\begin{equation}
  f_{\mathrm{test}}(s, \theta) \triangleq \pi_P(s, \theta) = (\tau, c, m) \in \mathcal{O},
  \label{eq:planner-interface}
\end{equation}
where $\mathcal{O}$ comprises a trajectory $\tau$, collision flag $c\in\{0,1\}$, and cost log $m$. The tuner $f_{\mathrm{tune}}$ enables natural-language \gls{ADS} Enhancement. Users state qualitative presets (e.g., \textit{Safety-Conservative}) or explicit weight adjustments (e.g., ``increase \texttt{distance\_to\_obstacles}''). The \gls{LLM}, invoked with the tuning prompt $\mathbf{P}_{\textsc{tune}}$, interprets utterance $u_t$ and emits a schema-conformant YAML override, updating the configuration $\theta \to \theta'$ in place while preserving formatting. 

To enable \gls{ADS} Benchmarking {(see Appendix~\ref{app:cross-planner-qualitative})}, PlannerForge wraps two classical motion planners, sampling-based Frenetix~\cite{frenetix} and learning-based MP-RBFN~\cite{rbfn}, under this $\pi_P$ interface. 
Execution operates in \emph{single-scenario} mode for individual analysis, or \emph{batch} mode (preset sizes, query-driven, or custom sets) for parallel statistical evaluation.

\subsection{Result Analysis Module}
\label{subsec:analysis}
The Analysis Module realizes the operator $f_{\mathrm{eval}}: \mathcal{O} \times \mathcal{U} \to \mathcal{Y}$ defined in \S\ref{sec:problem}, translating batch outcomes into natural-language feedback. Given a batch of $B$ outcomes $\{o_i = (\tau_i,\, c_i,\, m_i)\}_{i=1}^{B}$ produced under planner configuration $\theta$ (each $o_i$ as defined in Eq.~\ref{eq:planner-interface}: trajectory $\tau_i$, collision flag $c_i$, per-step cost log $m_i$) and a user utterance $u_t$ (e.g.\ ``why is the success rate low?''), the module assembles the analysis prompt $\mathbf{P}_{\textsc{eval}}$ over four context blocks: (i) batch-level statistics aggregated from $\{c_i\}$ and $\{m_i\}$ (success rate, mean trajectory length, collision count, mean cost); (ii) chronological per-scenario logs; (iii) the active configuration $\theta$; and (iv) the underlying CSV log path. The \gls{LLM} returns a response $y \in \mathcal{Y}$ comprising quantitative metrics, a failure-mode breakdown (\emph{collision}, \emph{timeout}, \emph{kinematic infeasibility}), the cost configuration used, and qualitative correlations between outcomes and scenario characteristics, together with parameter-adjustment recommendations that close the loop with the tuner $f_{\mathrm{tune}}$. A complementary \textit{behavior-comparison} path retains the chronological history of past batches with their configurations $\{(\theta^{(k)}, \{o_i^{(k)}\}_i)\}_k$ and lets the \gls{LLM} reason about which cost weights changed between runs and how those changes shifted the success/failure profile.

\section{Results \& Discussion}\label{sec:results}

In this section, we present the performance of PlannerForge with {}{quantitative results}.
Five cloud API models (Qwen3.6-plus~\cite{qwen36plus}, Deepseek-v3.2~\cite{liu2025deepseek}, Glm-5~\cite{zeng2026glm}, Gemini-3-flash\footnote{\url{https://blog.google/products-and-platforms/products/gemini/gemini-3-flash/}}, Gpt-5.4-mini\footnote{\url{https://openai.com/index/introducing-gpt-5-4-mini-and-nano/}}) and open-source models (Qwen3.6:35b~\cite{qwen36_35b_a3b} (think/no-think), Gemma4:31b\footnote{\url{https://blog.google/innovation-and-ai/technology/developers-tools/gemma-4/}} (think/no-think), Gpt-oss:20b~\cite{agarwal2025gpt} (think) from Ollama\footnote{\url{https://ollama.com/}} are evaluated. Metric definitions and scoring procedures for each task are documented in the Supplementary Material.

\subsection{Quantitative Evaluation}
\label{subsec:quantitative}
\begin{figure*}[t]
    \centering
    \includegraphics[width=0.99\linewidth]{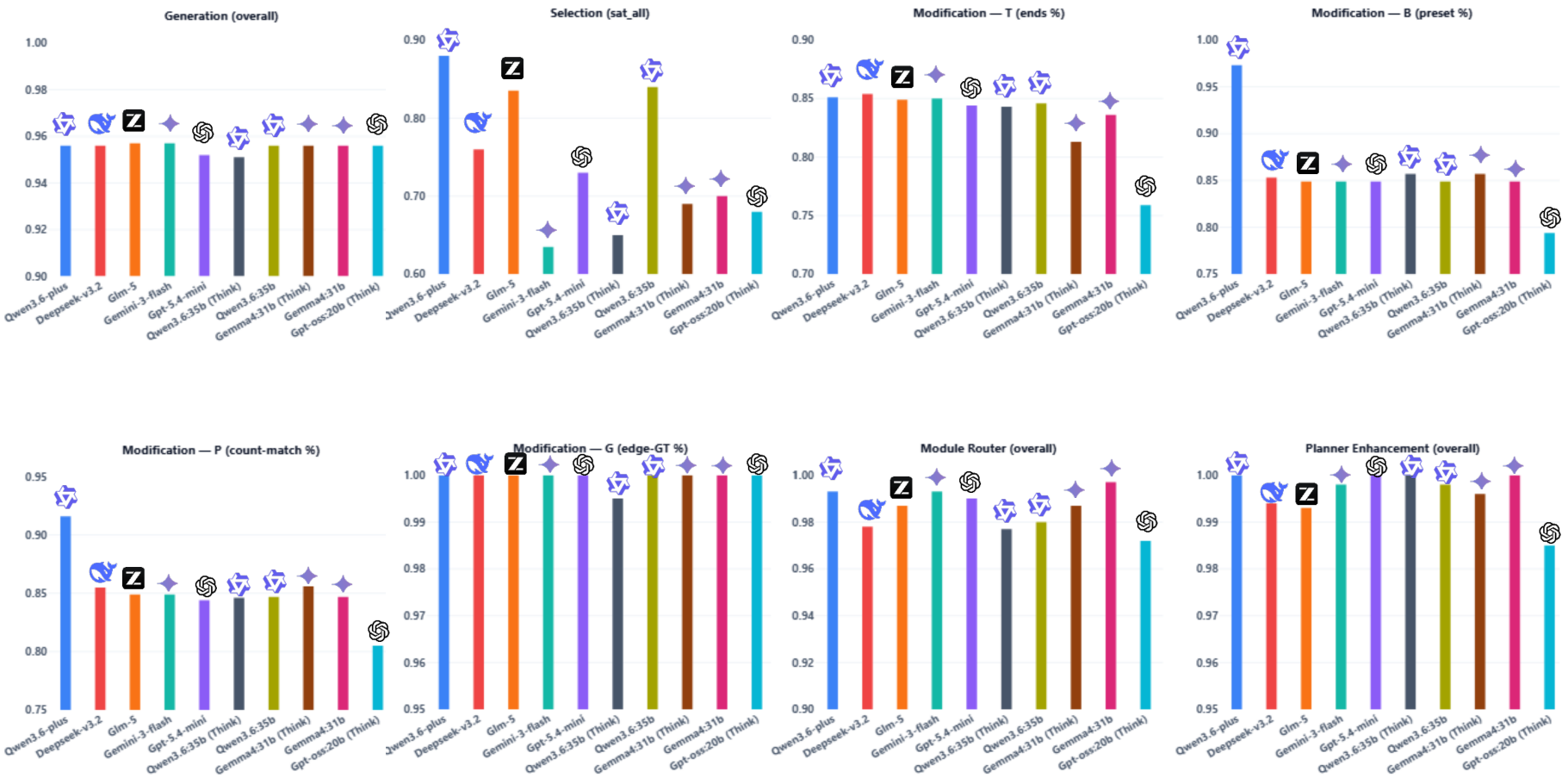}
    \caption{Best overall score per model on eight evaluated task slices (maximum across the prompt conditions).}
    \label{fig:best_per_task}
\end{figure*}

We evaluate PlannerForge on all $5$ tasks (Fig.~\ref{fig:best_per_task}), where the modification task spans the $4$ sub-tasks T/B/P/G, yielding $8$ task slices. We use $N{=}200$ queries per (sub-)task cell, $10$ model variants, and $5$ prompt conditions, from a bare \texttt{baseline} (zero shot) to \texttt{cp\_icl\_cot} (context prompting + in-context examples + chain-of-thought). For each task, we report the best-performing model under the prompt condition that achieves it, while per-cell ablations are in the appendix.

\textbf{Generation:} Glm-5 with \texttt{cp\_cot} (context prompting + chain-of-thought) achieves the best score, 0.957, improving over its bare baseline of 0.783 by +0.174. This suggests that CP+CoT is sufficient to nearly saturate intent parsing, while adding ICL provides no further gain. Because this task uses the Overpass API to interface with OpenStreetMap, the most difficult part is simply interpreting the user's free-form query into a structured intent.
\textbf{Selection:} Qwen3.6-plus with \texttt{cp\_cot} achieves the best \texttt{sat\_all} (joint satisfaction of all five retrieved scenarios) score, 0.880, compared with a baseline of 0.180 (+0.700). Selection is the hardest task because the strict five-stage filter fails whenever any extracted slot is incorrect {(per-slot extract in Appendix~\ref{app:sel-slot-extract})}. Adding ICL on top of CoT (\texttt{cp\_icl\_cot}, 0.835) can further hurt performance by encouraging over-confident slot guesses.
\textbf{Modification:} Under \texttt{cp\_icl\_cot}, Qwen3.6-plus reaches $\sim$100\% on the headline checks for all four sub-tasks: \textit{ends \%} (redirected route terminates at target edge) for T, \textit{preset \%} (behaviour vector matches preset) for B, \textit{count\_match \%} (vehicle count changes) for P, and \textit{edge-\gls{GT} \%} (extracted target lanelet matches \gls{GT}) for G. Because these modifications target SUMO configuration files, basic structural edits (T, P, G) are straightforward with high zero-shot baselines (99.0\%, 99.0\%, 88.5\%). In contrast, behavior modification (B) requires injecting precise parameter vectors, since it fails completely zero-shot (0\%) but reaches 100\% once advanced prompting provides the necessary context.
\textbf{Module Router:} Gemma4:31b with \texttt{cp\_icl\_cot} achieves 0.997, improving over its baseline of 0.722 by +0.275. In comparison, a hand-crafted regex router reaches only 45.5\% on the same corpus. This highlights that conversational intent is too diverse for rigid keyword matching, but is easily solved by LLMs via function calling given a clear JSON schema.
\textbf{Planner Testing and Enhancement:} Gpt-5.4-mini with \texttt{cp\_icl} achieves a perfect score of 1.000, compared with a baseline of 0.675 (+0.325), and outperforms a schema-constrained YAML editor baseline of 72.5\%. This shows that translating abstract user requests (e.g., ``drive safely'') into precise YAML parameter adjustments requires semantic understanding that rule-based editors lack. As downstream validation, 155 of 191 emitted YAML configurations (81.2\%) run end-to-end in Frenetix.

\begin{table*}[t]
{\ifshownew\color{red}\fi
\scriptsize
\setlength{\tabcolsep}{3pt}
\caption{End-to-end pipeline success ($N{=}200$ seed queries). Each stage consumes
the previous stage's actual output: Generation $\to$ Database $\to$ Selection $\to$
Modification $\to$ Test $\to$ Enhancement. Values are Commercial/Open (C/O), using
\texttt{qwen3.6-plus} and \texttt{qwen3.6:35b} with the \texttt{cp\_icl\_cot} prompt.
\textbf{FR} is the failure rate of that stage; \textbf{Cum.\ SR} is cumulative success
up to it. Latency and token cost are means per scenario.}
\label{tab:e2e-pipeline}
\noindent\makebox[\textwidth][c]{%
\begin{tabular*}{\textwidth}{@{\extracolsep{\fill}}lcccccc@{}}
\toprule
\textbf{Stage} & \textbf{in$\to$out (C/O)} & \textbf{FR $\downarrow$} & \textbf{Cum.\ SR $\uparrow$} & \textbf{Latency $\downarrow$} & \textbf{Token $\downarrow$} & \textbf{Hardware} \\
\midrule
\textcircled{1}~Generation   & 200$\to$192 / 200$\to$192 & 4\%/4\%   & 96\%/96\% & 21.6\,s/21.0\,s & 4.5k/4.6k & API/27G \\
\textcircled{2}~Database     & 192$\to$192              & 0\%/0\%   & 96\%/96\% & 1.8\,s/1.8\,s   & 0/0       & CPU \\
\textcircled{3}~Selection    & 192$\to$181 / 192$\to$170 & 6\%/11\%  & 91\%/85\% & 29.3\,s/26.7\,s & 8.5k/8.5k & API/27G \\
\textcircled{4}~Modification & 181$\to$165 / 170$\to$156 & 9\%/8\%   & 83\%/78\% & 44\,s/23\,s     & 29k/19k   & API/27G \\
\quad-- G (goal)             & 46$\to$46 / 43$\to$43     & 0\%/0\%   &           & 3.7\,s/2.5\,s   & 7.5k/7.5k & API/27G \\
\quad-- B (behaviour)        & 45$\to$42 / 43$\to$41     & 7\%/5\%   &           & 56.9\,s/31.2\,s & 33k/22k   & API/27G \\
\quad-- P (add/remove)       & 45$\to$38 / 42$\to$37     & 16\%/12\% &           & 56.6\,s/31.7\,s & 40k/21k   & API/27G \\
\quad-- T (trajectory)       & 45$\to$39 / 42$\to$35     & 13\%/17\% &           & 63.1\,s/29.2\,s & 34k/23k   & API/27G \\
\textcircled{5}~Test         & 165$\to$165 / 156$\to$156 & 0\%/0\%   & 83\%/78\% & 24.6\,s/23.9\,s & 0/0       & CPU \\
\textcircled{6}~Enhancement  & 165$\to$165 / 156$\to$156 & 0\%/0\%   & 83\%/78\% & 4.5\,s/2.6\,s   & 0.5k/0.5k & API/27G \\
\midrule
\textbf{$\to$~End-to-end} & \textbf{200}$\to$\textbf{165} / \textbf{200}$\to$\textbf{156} & & \textbf{83\%/78\%} & $\approx$\textbf{126\,s}/\textbf{99\,s} & $\approx$\textbf{42.5k}/\textbf{32.6k} & API/27G \\
\bottomrule
\end{tabular*}}}
\end{table*}

Per-module scores do not by themselves show that the stages compose. Table~\ref{tab:e2e-pipeline} therefore chains them on $N{=}200$ seed queries, each stage consuming the previous stage's actual output (Generation $\to$ Database $\to$ Selection $\to$ Modification $\to$ Test $\to$ Enhancement), for a commercial backend (\texttt{qwen3.6-plus}) and an open-source one (\texttt{qwen3.6:35b}) under \texttt{cp\_icl\_cot}. Both start at 96\% after Generation; Selection and Modification are the leak points, and the funnel ends at 83\% / 78\% cumulative success (commercial / open) with mean cost $\approx$126\,s / 99\,s and $\approx$42.5k / 32.6k tokens per scenario.

\textbf{Take-aways.} Schema-constrained tasks (Generation, Router, Planner) saturate with CP+CoT or CP+ICL, while semantically dense tasks (Selection, Modification) require advanced promptings. {Chained end-to-end, the pipeline retains 83\% / 78\% of seed queries (commercial / open).} Open-source 20--35B models perform slightly lower than commercial APIs but are fully capable of driving the entire pipeline. Notably, applying additional prompting techniques to open-source models with native reasoning (e.g., Think variants) disrupts their internal reasoning (Fig.~\ref{fig:best_per_task}), increasing latency and token consumption while degrading performance.

{}

\subsection{{Comparison with Prior Scenario-Testing Tools}}
\label{subsec:prior-tools}
\begin{table}[t]
{\ifshownew\color{red}\fi
\centering\scriptsize
\setlength{\tabcolsep}{2pt}
\caption{Cost-tuning across batch sizes, paired per scenario. Each batch is tuned
by three independent LLM calls (qwen3.6-plus, \texttt{cp\_icl\_cot}, temperature
0.0); we report mean\,$\pm$\,SD over the three rounds. The planner is
deterministic for a fixed (scenario, configuration) pair, so the LLM call is the
only stochastic component. All 15 rounds returned the identical configuration.}
\label{tab:cost-tuning-batches}
\resizebox{\columnwidth}{!}{%
\begin{tabular}{@{}rcccc@{}}
\toprule
\textbf{$N$} & \textbf{Success $\uparrow$ (b$\to$a)} & \textbf{$\Delta$ (pp)} & \textbf{Collision $\downarrow$ (b$\to$a)} & \textbf{$\Delta$ (pp)} \\
\midrule
50  & 46.7\%$\to$68.0$\pm$9.1\% & $+21.3\pm6.8$ & 20.7\%$\to$7.3$\pm$2.5\% & $-13.3\pm5.2$ \\
100 & 51.0\%$\to$68.6$\pm$4.2\% & $+17.6\pm2.6$ & 20.3\%$\to$9.8$\pm$0.9\% & $-10.5\pm1.7$ \\
200 & 51.3\%$\to$71.5$\pm$2.4\% & $+20.1\pm0.5$ & 20.6\%$\to$8.2$\pm$1.0\% & $-12.4\pm1.5$ \\
300 & 51.6\%$\to$69.8$\pm$0.5\% & $+18.2\pm0.7$ & 18.9\%$\to$7.8$\pm$0.6\% & $-11.1\pm1.3$ \\
400 & 50.4\%$\to$70.2$\pm$0.4\% & $+19.8\pm0.3$ & 19.0\%$\to$8.4$\pm$0.4\% & $-10.6\pm0.3$ \\
\bottomrule
\end{tabular}}}
\end{table}

\begin{table}[t]
{\ifshownew\color{red}\fi
\centering\scriptsize
\setlength{\tabcolsep}{3pt}
\caption{\textbf{Generation.} PlannerForge vs.\ the rule-based state of the art,
Scenario Factory~2.0 \citep{finkeldei2025scenariofactory}, on 200 queries across
50 cities. PlannerForge receives the full natural-language query; SF~2.0 receives
the extracted target city, its native input. \textbf{Exec.\,S}~= scenarios that
generate and execute in the planner. \xmark~= attribute not targetable.
$^\dagger$SF~2.0 is given the city directly.}
\label{tab:cmp-generation}
\resizebox{\columnwidth}{!}{%
\begin{tabular}{@{}lccccccc@{}}
\toprule
\textbf{Method} & \textbf{Time $\downarrow$} & \textbf{Exec.\,S $\uparrow$} & \textbf{City $\uparrow$} & \textbf{Road $\uparrow$} & \textbf{Vehicle $\uparrow$} & \textbf{Diverse $\uparrow$} & \textbf{Coll $\uparrow$} \\
\midrule
SF\,2.0 & \textbf{2.2\,s} & 144/200 & 72\%$^\dagger$ & \xmark & \xmark & 4 & 6.1\% \\
\textbf{PlannerForge} & 21.6\,s & \textbf{193/200} & \textbf{96.0\%} & \textbf{92.0\%} & \textbf{95.6\%} & \textbf{7} & \textbf{20.0\%} \\
\bottomrule
\end{tabular}}}
\end{table}

\begin{table}[t]
{\ifshownew\color{red}\fi
\centering\scriptsize
\setlength{\tabcolsep}{5pt}
\caption{\textbf{Selection.} PlannerForge vs.\ BM25 keyword search
\citep{robertson2009bm25} on 200 natural-language queries over a 500+ scenario
database; the retrieval backend is shared. \textbf{Satisfy@1 / Any@5}~= the
request is satisfied at rank~1 / anywhere in the top~5.}
\label{tab:cmp-selection}
\resizebox{\columnwidth}{!}{%
\begin{tabular}{@{}lcccc@{}}
\toprule
\textbf{Retrieval (top-5)} & \textbf{Latency $\downarrow$} & \textbf{Token $\downarrow$} & \textbf{Satisfy@1 $\uparrow$} & \textbf{Any@5 $\uparrow$} \\
\midrule
Keyword search / BM25 & \textbf{$<$0.01\,s} & \textbf{0} & 67.5\% & 86.0\% \\
\textbf{PlannerForge (LLM)} & 21.6\,s & 8.7k & \textbf{92.0\%} & \textbf{96.5\%} \\
\bottomrule
\end{tabular}}}
\end{table}

\begin{table}[t]
{\ifshownew\color{red}\fi
\centering\scriptsize
\setlength{\tabcolsep}{2pt}
\caption{\textbf{Modification.} PlannerForge (PF) four edit types vs.\
From-Words-to-Collisions \citep{gao2025words} on the same 200 base scenarios.
\textbf{Exec.\,S}~= planner-runnable; \textbf{Phy.\ Val.}~= physically valid share;
\textbf{New Coll.}~= valid new collisions; \textbf{min\_risk}~= mean
base$\to$modified risk (0~= collision, 5~= safe).}
\label{tab:cmp-modification}
\resizebox{\columnwidth}{!}{%
\begin{tabular}{@{}lccccccc@{}}
\toprule
\textbf{Method} & \textbf{Time $\downarrow$} & \textbf{Tokens $\downarrow$} & \textbf{Exec.\,S $\uparrow$} & \textbf{Phy.\ Val.\ $\uparrow$} & \textbf{New Coll.\ $\uparrow$} & \textbf{Goal $\downarrow$} & \textbf{min\_risk $\downarrow$} \\
\midrule
FWtC & 51\,s & 17.6k & 200/200 & 31.0\% & 16 & 24.2\% & $1.84{\to}1.69$ \\
PF (Behaviour)   & 56\,s & 24.2k & 200/200 & 98.0\% & 31 & 47.2\% & $1.84{\to}1.61$ \\
PF (Trajectory)  & 60\,s & 22.1k & 194/200 & 97.9\% & 32 & 50.5\% & $1.84{\to}1.51$ \\
PF (Participant) & 63\,s & 21.7k & 192/200 & 94.8\% & \textbf{58} & 35.7\% & $\mathbf{1.84{\to}1.20}$ \\
PF (Goal)        & \textbf{8\,s} & \textbf{7.1k} & 199/200 & \textbf{100\%} & 45 & \textbf{22.6\%} & $1.84{\to}1.36$ \\
\bottomrule
\end{tabular}}}
\end{table}

The modules above are reliable at schema-checked execution. We next show that
they also beat or lift the strongest available baseline at each stage, including
Enhancement versus the hand-set Default planner configuration.

\textbf{Generation} (Table~\ref{tab:cmp-generation}). CommonRoad has no
natural-language DSL, so alternative sourcing relies on GUI drawing. Against
Scenario Factory~2.0 \citep{finkeldei2025scenariofactory}, the rule-based state of
the art, PlannerForge is an order of magnitude slower per scenario, because it
runs an LLM where SF~2.0 runs a procedure. In exchange it delivers more
executable scenarios (193 vs.\ 144 of 200), realises 92--96\% of the requested
city, road and vehicle attributes that SF~2.0 cannot target at all, produces
seven traffic-participant classes rather than four, and induces $3.3\times$ more
planner collisions (20.0\% vs.\ 6.1\%).

\textbf{Selection} (Table~\ref{tab:cmp-selection}). The CommonRoad GUI supports
only manual parameter filters. Against BM25 keyword search
\citep{robertson2009bm25}, which is effectively free at $<$0.01\,s and zero
tokens, PlannerForge costs 21.6\,s and 8.7k tokens per query. BM25 already finds
a valid scenario somewhere in the top five for 86.0\% of queries, so the gap at
Any@5 is modest (96.5\%). The gap at rank~1 is what matters for an interactive
tool: 67.5\% vs.\ 92.0\%, because LLM slot extraction resolves paraphrase,
location ambiguity and implicit range constraints that keyword matching cannot.

\textbf{Modification} (Table~\ref{tab:cmp-modification}). Against
From-Words-to-Collisions \citep{gao2025words}, a recent LLM-based
safety-critical modification tool, the decisive difference is physical validity.
FWtC writes raw coordinates without vehicle dynamics, so roughly 70\% of its
edits are kinematically impossible, and it offers a single edit type. Because
PlannerForge routes every edit through SUMO, all four of its edit types stay
above 94\% valid, and Participant ($1.84{\to}1.20$, 58 new collisions) and Goal
($1.84{\to}1.36$, 45) stress the planner considerably harder than FWtC's valid
edits ($1.84{\to}1.69$, 16).

\textbf{Enhancement} (Table~\ref{tab:cost-tuning-batches}). The LLM retunes cost
weights against the hand-set Default configuration across five batch sizes
($N{=}50$--$400$), three independent calls each, scored paired per scenario.
Success rises in every batch ($+17.6$ to $+21.3$\,pp) and collisions fall
($10.5$ to $13.3$\,pp); at $N{=}400$, $50.4\%{\to}70.2\%$. Spread shrinks with
$N$ ($\pm6.8$\,pp at $50$ vs.\ $\pm0.3$\,pp at $400$). A Frenetix vs.\ MP-RBFN
dispatch is in Appendix~\ref{app:cross-planner-qualitative}.

Taken together, the LLM buys attribute control in Generation, rank-1 precision
in Selection, physically valid edits in Modification, and a lift over Default
in Enhancement, at a cost in seconds and tokens that the classical tools do not
pay.

\subsection{{Discussion: Transferable Insights}}
\label{subsec:discussion}
Three findings generalise to other structured-output agent tasks.
\textbf{Prompt techniques match distinct failure modes.}
\gls{CP} saturates closed vocabularies (\texttt{Gpt-5.4-mini}: Planner full-YAML
$36.1\%{\to}100\%$, Router $43.5\%{\to}91.0\%$).
\gls{ICL} is required for refusals (out-of-vocab parameters: $0\%$ baseline,
$44.4\%$ \gls{CP}, $100\%$ only with \gls{ICL}).
\gls{CoT} helps joint constraints (Selection \texttt{sat\_all}
$46.0\%{\to}72.0\%$ from \gls{CP} to \gls{CP}+\gls{CoT}).
Match the prompt to the failure mode rather than stacking every technique.
\textbf{External \gls{CoT} can conflict with native thinking.}
On Selection \texttt{sat\_all}, adding \gls{CoT} on top of \gls{CP} hurts every
reasoning-enabled model ($65.0{\to}40.0\%$, $67.0{\to}30.0\%$,
$58.0{\to}32.0\%$) while lifting non-thinking \texttt{Qwen3.6:35b}
($63.5{\to}83.0\%$). Treat reasoning models as a distinct prompting regime.
\textbf{Reliability needs an executable harness.}
Code around each \gls{LLM} call parses, schema-validates, and scores both form
and downstream execution. The prompt raises the hit rate; the harness makes the
stage dependable.

\section{Conclusion and Future Work}

{}
{We presented \textbf{PlannerForge}, a full-lifecycle \gls{LLM}-agent framework for scenario-based testing of motion planners, with a Module Router, schema-checked modules, and a unified motion planner interface. Across 80{,}000 off-the-shelf calls with no fine-tuning, modules score from 0.88 (Selection) to 1.00 (Planner Testing), with Generation at 0.957, Modification near 100\% on the headline checks, and the Router at 0.997. End-to-end chaining from Generation through Enhancement retains 83\% / 78\% of seed queries (commercial / open). Against Scenario Factory~2.0, BM25, and From-Words-to-Collisions, it is more attribute-faithful in generation (193 vs.\ 144 executable), more precise at rank~1 selection (92.0\% vs.\ 67.5\%), and physically valid in modification ($\geq$94\% vs.\ 31\%). On planner safety-critical performance, generated scenarios induce 20.0\% collisions versus 6.1\% for Scenario Factory~2.0, and cost-tuning against Default at $N{=}400$ lifts success from 50.4\% to 70.2\% while cutting collisions from 19.0\% to 8.4\%. Future work will close the planner loop and extend the framework to simulators such as CARLA.}

\section*{Limitations}
\phantomsection\label{sec:limitations}

{}

{}

{}

\textbf{Measured failure modes.} In the end-to-end chain, Trajectory and Population edits drop 13--17\% and 12--16\% of surviving queries on simulation round-trip, not on headline semantics (Table~\ref{tab:e2e-pipeline}, Appendix~\ref{app:mod-metrics}). Selection is the other leak. Tag over-prediction drives the 6\%/11\% commercial/open drop, and the best \texttt{sat\_all} is 0.880 (Appendix~\ref{app:sel-slot-extract}). The Analysis Module is not scored against ground truth.

\textbf{Scope and domain.} PlannerForge runs open-loop. Other agents follow recorded or SUMO-exported trajectories and do not react to the ego vehicle. Closed-loop falsification is left to future work. Quantitative planner, collision, and cost-tuning results use Frenetix. The MP-RBFN comparison is qualitative (Appendix~\ref{app:cross-planner-qualitative}). Evaluation uses CommonRoad, and the modification corpus is Germany-dominated (DEU is 80--92 queries per task among 15 country codes).

\section*{Ethical Considerations}
\phantomsection\label{sec:ethics}
PlannerForge is driven by off-the-shelf \gls{LLM} agents that can hallucinate structured outputs. In our evaluation this appears as invented scenario tags, invalid map or vehicle identifiers, misrouted module calls, and Analysis claims that are not supported by the run logs. Unchecked, such errors can produce invalid tests or misleading planner diagnostics. Every scored module therefore parses, schema-validates, and executes the model output before it is accepted. Residual risk remains where that check is incomplete, in particular the unscored Analysis module.

\section*{Acknowledgements}
{}
{The authors wrote the initial draft and used LLMs only to improve grammar, clarity, and readability. They reviewed every suggestion and take responsibility for the paper.}

\bibliography{literature}

@INPROCEEDINGS{Betz2024,
  author={Betz, Johannes and Lutwitzi, Melina and Peters, Steven},
  booktitle={2024 IEEE Intelligent Vehicles Symposium (IV)}, 
  title={A new Taxonomy for Automated Driving: Structuring Applications based on their Operational Design Domain, Level of Automation and Automation Readiness}, 
  year={2024},
  volume={},
  number={},
  pages={1-7},
  doi={10.1109/IV55156.2024.10588711}}

@misc{Waymo2018,
  author = {Waymo},
  title = {Waymo One: The Next Step on Our Self-Driving Journey},
  year = {2018},
  url = {https://waymo.com/blog/2018/12/waymo-one-next-step-on-our-self-driving},
}

@article{SAE2021,
  author = {SAE International},
  title = {Taxonomy and Definitions for Terms Related to Driving Automation Systems for On-Road Motor Vehicles},
  year = {2021},
  journal = {SAE J3016},
}

@article{Song2024, author = {Song, Qunying and Engstr\"{o}m, Emelie and Runeson, Per}, title = {Industry Practices for Challenging Autonomous Driving Systems with Critical Scenarios}, 
year = {2024}, issue_date = {May 2024}, 
publisher = {Association for Computing Machinery}, 
address = {New York, NY, USA}, volume = {33}, number = {4}, 
issn = {1049-331X}, doi = {10.1145/3640334}, journal = {ACM Trans. Softw. Eng. Methodol.}, 
month = apr, articleno = {99}, numpages = {35}
}

@ARTICLE{gao2026foundation,
  author={Gao, Yuan and Piccinini, Mattia and Zhang, Yuchen and Wang, Dingrui and Moller, Korbinian and Brusnicki, Roberto and Zarrouki, Baha and Gambi, Alessio and Totz, Jan Frederik and Storms, Kai and Peters, Steven and Stocco, Andrea and Alrifaee, Bassam and Pavone, Marco and Betz, Johannes},
  journal={IEEE Open Journal of Intelligent Transportation Systems}, 
  title={Foundation Models in Autonomous Driving: A Survey on Scenario Generation and Scenario Analysis}, 
  year={2026},
  volume={},
  number={},
  pages={1-1},
  doi={10.1109/OJITS.2026.3660686}}

@InProceedings{pegasus,
author="Winner, Hermann
and Lemmer, Karsten
and Form, Thomas
and Mazzega, Jens",
editor="Meyer, Gereon
and Beiker, Sven",
title="PEGASUS---First Steps for the Safe Introduction of Automated Driving",
booktitle="Road Vehicle Automation 5",
year="2019",
publisher="Springer International Publishing",
address="Cham",
pages="185--195",
isbn="978-3-319-94896-6"
}

@article{song2026generative,
  title={Generative AI for Testing of Autonomous Driving Systems: A Survey},
  author={Song, Qunying and Ye, He and Harman, Mark and Sarro, Federica},
  journal={ACM Transactions on Software Engineering and Methodology},
  year={2026},
  doi={10.1145/3806653},
  publisher={ACM New York, NY}
}

@article{riedmaier2020survey,
  title={Survey on scenario-based safety assessment of automated vehicles},
  author={Riedmaier, Stefan and Ponn, Thomas and Ludwig, Dieter and Schick, Bernhard and Diermeyer, Frank},
  journal={IEEE access},
  volume={8},
  pages={87456--87477},
  year={2020},
  publisher={IEEE}
}

@article{nakamura2022defining,
  title={Defining reasonably foreseeable parameter ranges using real-world traffic data for scenario-based safety assessment of automated vehicles},
  author={Nakamura, Hiroki and Muslim, Husam and Kato, Ryosuke and Pr{\'e}fontaine-Watanabe, Sandra and Nakamura, H and Kaneko, H and Imanaga, Hisashi and Antona-Makoshi, Jacobo and Kitajima, Sou and Uchida, Nobuyuki and others},
  journal={IEEE Access},
  volume={10},
  pages={37743--37760},
  year={2022},
  publisher={IEEE}
}

@inproceedings{nalic2020scenario,
  title={Scenario based testing of automated driving systems: A literature survey},
  author={Nalic, Demin and Mihalj, Tomislav and B{\"a}umler, Maximilian and Lehmann, Matthias and Eichberger, Arno and Bernsteiner, Stefan},
  booktitle={FISITA web Congress},
  volume={10},
  pages={1},
  year={2020}
}

@article{ding2023survey,
  title={A survey on safety-critical driving scenario generation—a methodological perspective},
  author={Ding, Wenhao and Xu, Chejian and Arief, Mansur and Lin, Haohong and Li, Bo and Zhao, Ding},
  journal={IEEE Transactions on Intelligent Transportation Systems},
  volume={24},
  number={7},
  pages={6971--6988},
  year={2023},
  publisher={IEEE}
}

@article{tang2024scenario,
  title={Scenario-Based Accelerated Testing for {SOTIF} in Autonomous Driving: A Review},
  author={Tang, Lei and Wang, Ruijie and Liu, Zhanwen and Liang, Yunji and Niu, Yuanyuan and Zhu, Wei and Duan, Zongtao},
  journal={IEEE Internet of Things Journal},
  volume={12},
  number={2},
  pages={1453--1470},
  year={2025},
  doi={10.1109/JIOT.2024.3490598},
  publisher={IEEE}
}

@article{sharath2021literature,
  title={A literature review of performance metrics of automated driving systems for on-road vehicles},
  author={Sharath, Mysore Narasimhamurthy and Mehran, Babak},
  journal={Frontiers in Future Transportation},
  volume={2},
  pages={759125},
  year={2021},
  publisher={Frontiers Media SA}
}

@inproceedings{SUMO2018,
          title = {Microscopic Traffic Simulation using SUMO},
         author = {Pablo Alvarez Lopez and Michael Behrisch and Laura Bieker-Walz and Jakob Erdmann and Yun-Pang Fl{\"o}tter{\"o}d and Robert Hilbrich and Leonhard L{\"u}cken and Johannes Rummel and Peter Wagner and Evamarie Wie{\ss}ner},
      publisher = {IEEE},
      booktitle = {The 21st IEEE International Conference on Intelligent Transportation Systems},
           year = {2018},
 }

@inproceedings{zhang2024chatscene,
  title={ChatScene: Knowledge-Enabled Safety-Critical Scenario Generation for Autonomous Vehicles},
  author={Zhang, Jiawei and Xu, Chejian and Li, Bo},
  booktitle={Proceedings of the IEEE/CVF Conference on Computer Vision and Pattern Recognition},
  pages={15459--15469},
  year={2024}
}

@article{ruan2024traffic,
  title={Traffic Scene Generation from Natural Language Description for Autonomous Vehicles with Large Language Model},
  author={Ruan, Bo-Kai and Tsui, Hao-Tang and Li, Yung-Hui and Shuai, Hong-Han},
  journal={arXiv preprint arXiv:2409.09575},
  year={2024}
}

@article{li2024chatsumo,
  title={ChatSUMO: Large Language Model for Automating Traffic Scenario Generation in Simulation of Urban {MObility}},
  author={Li, Shuyang and Azfar, Talha and Ke, Ruimin},
  journal={IEEE Transactions on Intelligent Vehicles},
  year={2025},
  doi={10.1109/TIV.2024.3508471},
  publisher={IEEE}
}

@inproceedings{tan2023language,
  title={Language Conditioned Traffic Generation},
  author={Tan, Shuhan and Ivanovic, Boris and Weng, Xinshuo and Pavone, Marco and Kraehenbuehl, Philipp},
  booktitle={Proceedings of the 7th Conference on Robot Learning},
  series={Proceedings of Machine Learning Research},
  volume={229},
  pages={2714--2738},
  year={2023},
  publisher={PMLR}
}

@inproceedings{zhao2024chat2scenario,
  title={Chat2Scenario: Scenario Extraction From Dataset Through Utilization of Large Language Model},
  author={Zhao, Yongqi and Xiao, Wenbo and Mihalj, Tomislav and Hu, Jia and Eichberger, Arno},
  booktitle={2024 IEEE Intelligent Vehicles Symposium (IV)},
  pages={559--566},
  year={2024},
  organization={IEEE}
}

@article{mei2025llm,
  title={{LLM-Attacker}: Enhancing Closed-Loop Adversarial Scenario Generation for Autonomous Driving with Large Language Models},
  author={Mei, Yuewen and Nie, Tong and Sun, Jian and Tian, Ye},
  journal={IEEE Transactions on Intelligent Transportation Systems},
  year={2025},
  doi={10.1109/TITS.2025.3578383}
}

@ARTICLE{deng2023target,
  author={Deng, Yao and Tu, Zhi and Yao, Jiaohong and Zhang, Mengshi and Zhang, Tianyi and Zheng, Xi},
  journal={IEEE Transactions on Software Engineering}, 
  title={TARGET: Traffic Rule-Based Test Generation for Autonomous Driving via Validated LLM-Guided Knowledge Extraction}, 
  year={2025},
  volume={51},
  number={7},
  pages={1950-1968},
  doi={10.1109/TSE.2025.3569086}}

@inproceedings{petrovic2024llm,
  title={LLM-Driven Testing for Autonomous Driving Scenarios},
  author={Petrovic, Nenad and Lebioda, Krzysztof and Zolfaghari, Vahid and Schamschurko, Andr{\'e} and Kirchner, Sven and Purschke, Nils and Pan, Fengjunjie and Knoll, Alois},
  booktitle={2024 2nd International Conference on Foundation and Large Language Models (FLLM)},
  pages={173--178},
  year={2024},
  organization={IEEE}
}

@inproceedings{guo2024sovar,
  title={Sovar: Build generalizable scenarios from accident reports for autonomous driving testing},
  author={Guo, An and Zhou, Yuan and Tian, Haoxiang and Fang, Chunrong and Sun, Yunjian and Sun, Weisong and Gao, Xinyu and Luu, Anh Tuan and Liu, Yang and Chen, Zhenyu},
  booktitle={Proceedings of the 39th IEEE/ACM International Conference on Automated Software Engineering},
  pages={268--280},
  year={2024}
}

@inproceedings{tang2024legend,
  title={Legend: A top-down approach to scenario generation of autonomous driving systems assisted by large language models},
  author={Tang, Shuncheng and Zhang, Zhenya and Zhou, Jixiang and Lei, Lei and Zhou, Yuan and Xue, Yinxing},
  booktitle={Proceedings of the 39th IEEE/ACM International Conference on Automated Software Engineering},
  pages={1497--1508},
  year={2024}
}

@article{chang2024llmscenario,
  title={Llmscenario: Large language model driven scenario generation},
  author={Chang, Cheng and Wang, Siqi and Zhang, Jiawei and Ge, Jingwei and Li, Li},
  journal={IEEE Transactions on Systems, Man, and Cybernetics: Systems},
  year={2024},
  publisher={IEEE}
}

@INPROCEEDINGS{gao2025words,
  author={Gao, Yuan and Piccinini, Mattia and Moller, Korbinian and Alanwar, Amr and Betz, Johannes},
  booktitle={2025 IEEE 28th International Conference on Intelligent Transportation Systems (ITSC)}, 
  title={From Words to Collisions: LLM-Guided Evaluation and Adversarial Generation of Safety-Critical Driving Scenarios}, 
  year={2025},
  volume={},
  number={},
  pages={2134-2141},
  doi={10.1109/ITSC60802.2025.11423486}}

@article{aasi2024generating,
  title={Generating Out-Of-Distribution Scenarios Using Language Models},
  author={Aasi, Erfan and Nguyen, Phat and Sreeram, Shiva and Rosman, Guy and Karaman, Sertac and Rus, Daniela},
  journal={arXiv preprint arXiv:2411.16554},
  year={2024}
}

@article{cai2026text2scenario,
  title={{Text2Scenario}: Text-Driven Scenario Generation for Autonomous Driving Test},
  author={Cai, Xuan and Bai, Xuesong and Cui, Zhiyong and Xie, Danmu and Fu, Daocheng and others},
  journal={Automotive Innovation},
  year={2026},
  doi={10.1007/s42154-025-00374-8},
  publisher={Springer}
}

@article{haklay2008openstreetmap,
  title={Openstreetmap: User-generated street maps},
  author={Haklay, Mordechai and Weber, Patrick},
  journal={IEEE Pervasive computing},
  volume={7},
  number={4},
  pages={12--18},
  year={2008},
  publisher={Ieee}
}

@inproceedings{Krajewski2018highd,
  title = {The highD Dataset: A Drone Dataset of Naturalistic Vehicle Trajectories on German Highways for Validation of Highly Automated Driving Systems},
  DOI = {10.1109/itsc.2018.8569552},
  booktitle = {2018 21st International Conference on Intelligent Transportation Systems (ITSC)},
  publisher = {IEEE},
  author = {Krajewski,  Robert and Bock,  Julian and Kloeker,  Laurent and Eckstein,  Lutz},
  year = {2018},
  month = nov,
  pages = {2118–2125}
}

@article{chang2022metascenario,
  title={MetaScenario: A framework for driving scenario data description, storage and indexing},
  author={Chang, Cheng and Cao, Dongpu and Chen, Long and Su, Kui and Su, Kuifeng and Su, Yuelong and Wang, Fei-Yue and Wang, Jue and Wang, Ping and Wei, Junqing and others},
  journal={IEEE Transactions on Intelligent Vehicles},
  volume={8},
  number={2},
  pages={1156--1175},
  year={2022},
  publisher={IEEE}
}

@inproceedings{althoff2017commonroad,
  title={CommonRoad: Composable benchmarks for motion planning on roads},
  author={Althoff, Matthias and Koschi, Markus and Manzinger, Stefanie},
  booktitle={2017 IEEE Intelligent Vehicles Symposium (IV)},
  pages={719--726},
  year={2017},
  organization={IEEE}
}

@inproceedings{baumann2025enhancing,
  title={Enhancing Autonomous Driving Systems with On-Board Deployed Large Language Models},
  author={Baumann, Nicolas and Hu, Cheng and Sivasothilingam, Paviththiren and Qin, Haotong and Xie, Lei and Magno, Michele and Benini, Luca},
  booktitle={Robotics: Science and Systems {XXI}},
  year={2025},
  doi={10.15607/rss.2025.xxi.140}
}

@article{wang2024dualad,
  title={Dualad: Dual-layer planning for reasoning in autonomous driving},
  author={Wang, Dingrui and Kaufeld, Marc and Betz, Johannes},
  journal={arXiv preprint arXiv:2409.18053},
  year={2024}
}

@article{zhang2025lead,
  title={Lead: The llm enhanced planning system converged with end-to-end autonomous driving},
  author={Zhang, Yuhang and Liu, Jiaqi and Xu, Chengkai and Hang, Peng and Sun, Jian},
  journal={arXiv preprint arXiv:2507.05754},
  year={2025}
}

@inproceedings{reimers2019sentence,
  title={Sentence-BERT: Sentence Embeddings using Siamese BERT-Networks},
  author={Reimers, Nils and Gurevych, Iryna},
  booktitle={Proceedings of the 2019 Conference on Empirical Methods in Natural Language Processing and the 9th International Joint Conference on Natural Language Processing (EMNLP-IJCNLP)},
  pages={3982--3992},
  year={2019},
  organization={Association for Computational Linguistics},
  doi={10.18653/v1/D19-1410}
}

@misc{chroma,
  title={Chroma: The AI-native open-source embedding database},
  author={{Chroma Team}},
  year={2023},
  howpublished={\url{https://www.trychroma.com/}},
  note={Accessed: 2026-01-18}
}

@inproceedings{sumocr,
  title={Coupling SUMO with a Motion Planning Framework for Automated Vehicles},
  author={Klischat, Moritz and Dragoi, Octav and Eissa, Mostafa and Althoff, Matthias},
  booktitle={SUMO User Conference},
  pages={1--9},
  year={2019}
}

@article{frenetix,
  title={FRENETIX: A high-performance and modular motion planning framework for autonomous driving},
  author={Trauth, Rainer and Moller, Korbinian and W{\"u}rsching, Gerald and Betz, Johannes},
  journal={IEEE Access},
  year={2024},
  publisher={IEEE}
}

@INPROCEEDINGS{rbfn,
  author={Kaufeld, Marc and Piccinini, Mattia and Betz, Johannes},
  booktitle={2025 IEEE 28th International Conference on Intelligent Transportation Systems (ITSC)}, 
  title={MP-RBFN: Learning-Based Vehicle Motion Primitives Using Radial Basis Function Networks}, 
  year={2025},
  volume={},
  number={},
  pages={1262-1269},
  doi={10.1109/ITSC60802.2025.11423753}}

@article{bauerfeind2025david,
  title={David vs. Goliath: A comparative study of different-sized LLMs for code generation in the domain of automotive scenario generation},
  author={Bauerfeind, Philipp and Salarpour, Amir and Fernandez, David and MohajerAnsari, Pedram and Reschke, Johannes and others},
  journal={arXiv preprint arXiv:2510.14115},
  year={2025}
}

@inproceedings{dosovitskiy2017carla,
  title={CARLA: An open urban driving simulator},
  author={Dosovitskiy, Alexey and Ros, German and Codevilla, Felipe and Lopez, Antonio and Koltun, Vladlen},
  booktitle={Conference on robot learning},
  year={2017},
  organization={PMLR}
}

@misc{qwen36plus,
    title = {{Qwen3.6-Plus}: Towards Real World Agents},
    url = {https://qwen.ai/blog?id=qwen3.6},
    author = {{Qwen Team}},
    month = {April},
    year = {2026}
}

@article{liu2025deepseek,
  title={Deepseek-v3. 2: Pushing the frontier of open large language models},
  author={Liu, Aixin and Mei, Aoxue and Lin, Bangcai and Xue, Bing and Wang, Bingxuan and Xu, Bingzheng and Wu, Bochao and Zhang, Bowei and Lin, Chaofan and Dong, Chen and others},
  journal={arXiv preprint arXiv:2512.02556},
  year={2025}
}

@article{zeng2026glm,
  title={Glm-5: from vibe coding to agentic engineering},
  author={Zeng, Aohan and Lv, Xin and Hou, Zhenyu and Du, Zhengxiao and Zheng, Qinkai and Chen, Bin and Yin, Da and Ge, Chendi and Huang, Chenghua and Xie, Chengxing and others},
  journal={arXiv preprint arXiv:2602.15763},
  year={2026}
}

@misc{qwen36_35b_a3b,
    title = {{Qwen3.6-35B-A3B}: Agentic Coding Power, Now Open to All},
    url = {https://qwen.ai/blog?id=qwen3.6-35b-a3b},
    author = {{Qwen Team}},
    month = {April},
    year = {2026}
}

@article{agarwal2025gpt,
  title={gpt-oss-120b \& gpt-oss-20b model card},
  author={Agarwal, Sandhini and Ahmad, Lama and Ai, Jason and Altman, Sam and Applebaum, Andy and Arbus, Edwin and Arora, Rahul K and Bai, Yu and Baker, Bowen and Bao, Haiming and others},
  journal={arXiv preprint arXiv:2508.10925},
  year={2025}
}

@article{mao2023language,
  title={A language agent for autonomous driving},
  author={Mao, Jiageng and Ye, Junjie and Qian, Yuxi and Pavone, Marco and Wang, Yue},
  journal={arXiv preprint arXiv:2311.10813},
  year={2023}
}

@article{finkeldei2025scenariofactory,
  title   = {Scenario Factory 2.0: Scenario-Based Testing of Automated Vehicles with {CommonRoad}},
  author  = {Finkeldei, Florian and Thees, Christoph and Weghorn, Jan-Niklas and Althoff, Matthias},
  journal = {Automotive Innovation},
  volume  = {8},
  number  = {2},
  pages   = {207--220},
  year    = {2025},
  doi     = {10.1007/s42154-025-00360-0}
}

@INPROCEEDINGS{gao2026chat2scenic,
  author    = {Gao, Yuan and Miao, Wenting and Piccinini, Mattia and Wang, Haoyu and Song, Qunying and Betz, Johannes},
  title     = {Chat2Scenic: An Iterative {RAG}-Based Framework for Scenario Generation in Autonomous Driving},
  booktitle = {2026 IEEE/RSJ International Conference on Intelligent Robots and Systems ({IROS})},
  year      = {2026},
  note      = {to appear},
  eprint    = {2607.14387},
  archivePrefix = {arXiv},
  primaryClass  = {cs.AI}
}

@article{robertson2009bm25,
  title   = {The Probabilistic Relevance Framework: {BM25} and Beyond},
  author  = {Robertson, Stephen and Zaragoza, Hugo},
  journal = {Foundations and Trends in Information Retrieval},
  volume  = {3},
  number  = {4},
  pages   = {333--389},
  year    = {2009},
  doi     = {10.1561/1500000019}
}

\clearpage
\appendix
\section{Appendix}
\label{sec:appendix}

\subsection{Implementation Details}
\label{app:impl}

\textbf{Hardware.} All experiments are conducted on a single workstation equipped with an NVIDIA GeForce RTX 5090 GPU (Blackwell architecture, 32\, GB GDDR7, 21,760 CUDA cores, 1,792 GB/s memory bandwidth, 575\, W TGP).
The host CPU is an Intel Core i9-14900K (24 cores / 32 threads), paired with 128\, GB DDR5 system memory and a 2\, TB NVMe SSD for scenario storage and SUMO simulation caches.

\textbf{Large Language Models.} The LLMs used in the experiments are listed in Table~\ref{tab:llm-models}. Models are grouped by provider, and the ``Think'' column indicates whether the model's internal reasoning mode is enabled at inference time (\texttt{ON}) or disabled (\texttt{OFF}). All cloud-hosted closed-weight models run with reasoning disabled to keep their structured-output behavior comparable. For local open-weight reasoning models (\texttt{gpt-oss}, \texttt{gemma-4}) reasoning is on by default, and for \texttt{qwen3.6:35b} and \texttt{gemma4:31b} we evaluate both modes.

\begin{table}[ht]
\centering
\footnotesize
\setlength{\tabcolsep}{4pt}
\caption{LLM backends used in the experiments, grouped by provider. ``Think'' = internal reasoning mode at inference time. {``VRAM'' = measured peak resident GPU memory of the local Ollama backend (\texttt{Q4\_K\_M} weights, 32k-token context window) on the RTX 5090; cloud models are served over a provider API and use no local VRAM.}}
\label{tab:llm-models}
\begin{tabular}{@{}llcc@{}}
\toprule
Model & Provider & Think & {VRAM} \\
\midrule
\rowcolor[HTML]{F0F0F0}
\multicolumn{4}{@{}l}{\emph{Cloud API}} \\
\quad \textcolor{blue}{\textbf{Qwen3.6-plus}}           & DashScope & \textcolor{red}{OFF} & {API} \\
\quad \textbf{Deepseek-v3.2}          & DashScope & \textcolor{red}{OFF} & {API} \\
\quad \textcolor{orange}{\textbf{Glm-5}}                  & DashScope & \textcolor{red}{OFF} & {API} \\
\quad \textcolor{teal}{\textbf{Gemini-3-flash}} & Google  & \textcolor{red}{OFF} & {API} \\
\quad \textcolor{violet}{\textbf{Gpt-5.4-mini}}           & OpenAI    & \textcolor{red}{OFF} & {API} \\
\midrule
\rowcolor[HTML]{F0F0F0}
\multicolumn{4}{@{}l}{\emph{Local Ollama (RTX 5090, Q4\_K\_M)}} \\
\quad \textcolor{darkgray}{\textbf{Qwen3.6:35b}}            & Ollama    & \textcolor{green}{ON}  & {27 GB} \\
\quad \textcolor{olive}{\textbf{Qwen3.6:35b}}            & Ollama    & \textcolor{red}{OFF} & {27 GB} \\
\quad \textcolor{brown}{\textbf{Gemma4:31b}}             & Ollama    & \textcolor{green}{ON}  & {27 GB} \\
\quad \textcolor{magenta}{\textbf{Gemma4:31b}}             & Ollama    & \textcolor{red}{OFF} & {27 GB} \\
\quad \textcolor{cyan}{\textbf{Gpt-oss:20b}}            & Ollama    & \textcolor{green}{ON}  & {14 GB} \\
\bottomrule
\end{tabular}
\end{table}

{\textbf{Pinned identifiers (reproducibility).} Commercial models are invoked by their exact provider model-id: \texttt{qwen3.6-plus}, \texttt{deepseek-v3.2}, \texttt{glm-5} (DashScope), \texttt{gemini-3-flash-preview} (Google), \texttt{gpt-5.4-mini} (OpenAI). Open backends are pinned Ollama image tags: \texttt{qwen3.6:35b} @ \texttt{07d3521}, \texttt{gemma4:31b} @ \texttt{6316f06}, \texttt{gpt-oss:20b} @ \texttt{17052f9}. A fully commercial-API-free run uses only the five open backends.}

\textbf{Framework components.} PlannerForge integrates various data sources and simulation tools into a unified pipeline:
\begin{itemize}
    \item \textbf{Frontend.} The framework is exposed as a web application built with \textbf{Gradio}, organized as two browser tabs: a \emph{Main} tab hosting the chatbot interface for query, modification, testing, and analysis, and a separate \emph{Generate} tab for the OSM-based scenario-generation pipeline. The chatbot widget streams \gls{LLM} responses and renders generated/simulated scenarios as inline animated GIFs.
    \item \textbf{Conversational memory.} To preserve chat history context across multi-turn interactions, the Python runtime leverages Langchain's\footnote{\url{https://www.langchain.com/}} \texttt{ConversationSummaryBufferMemory}, which automatically summarises older messages to fit within token limits.
    \item \textbf{Simulation backends.} PlannerForge operates on the CommonRoad scenario format~\cite{althoff2017commonroad}, with traffic dynamics provided by the SUMO microscopic traffic simulator~\cite{SUMO2018} via the CommonRoad-SUMO interface~\cite{sumocr}. We integrate two motion planners: \textbf{Frenetix}~\cite{frenetix} (sampling-based, weighted cost functions) and \textbf{MP-RBFN}~\cite{rbfn} (learning-based, radial basis function networks).
    \item \textbf{Scenario database.} We index 500+ curated CommonRoad scenarios in a persistent \textbf{ChromaDB}~\cite{chroma} vector store (\texttt{PersistentClient}), using SentenceTransformer~\cite{reimers2019sentence} embeddings for semantic retrieval. Each scenario includes a structured metadata dictionary covering location (country, road network type), roadside infrastructure (e.g., traffic light presence), participant characteristics (dynamic and static obstacle counts and types), and ego-vehicle properties (initial velocity), enabling exact-match filtering. The hybrid metadata-filtering and RAG strategy is described in the Methodology.
    \item \textbf{OSM data sources.} The Generation Module retrieves real-world road networks from the OpenStreetMap project~\cite{haklay2008openstreetmap} via two public APIs: (i) the \textbf{Overpass API} for raw \texttt{.osm} XML fetching. We use four mirror endpoints (\nolinkurl{overpass-api.de}, \nolinkurl{overpass.kumi.systems}, \nolinkurl{overpass.private.coffee}, \nolinkurl{maps.mail.ru}) as automatic fallbacks for resilience; and (ii) \textbf{Nominatim} (accessed through the \texttt{osmnx} library) for geocoding city names into bounding boxes. A polite delay between Overpass requests respects the public-endpoint rate limits.
\end{itemize}

\subsection{Scenario Generation Module}
\label{app:gen-corpus}

\subsubsection{LLM Scope and Hallucination Isolation}
\label{app:gen-llm-scope}

The Generation Module separates \emph{language-level} tasks (handled by the \gls{LLM}) from \emph{geometric- and dynamic-level} tasks (handled by deterministic tools). This split is a deliberate design choice: it preserves the flexibility of natural-language scenario authoring while preventing LLM hallucinations from propagating into map topology or vehicle dynamics.

\paragraph{LLM responsibilities.}
The \gls{LLM} is invoked at two well-typed boundaries:
\begin{itemize}\itemsep0pt\parsep0pt
  \item \textbf{Stage 1 --- intent parsing.} The free-form utterance is parsed into a JSON intent with fixed keys: location (city or explicit bounding box), drivable road classes (subset of OSM \texttt{highway} tags), traffic density (categorical: \texttt{low}/\texttt{medium}/\texttt{high}), vehicle mix (counts per \texttt{vClass}), and simulation duration. Each key has a schema-defined default that is applied when the LLM omits or emits an invalid value. The per-key defaults-compliance rates are reported in Table~\ref{tab:gen-all-results}.
  \item \textbf{Stage 2 --- ego selection.} From the populated traffic, the LLM picks one of three strategies (\emph{first car}, \emph{by type}, or \emph{by index}) to identify an ego vehicle. The goal region is then attached deterministically (\emph{chosen lanelet} or \emph{forward offset along the ego trajectory}).
\end{itemize}

\paragraph{Deterministic-tool responsibilities.}
Three downstream stages execute without LLM involvement:
\begin{itemize}\itemsep0pt\parsep0pt
  \item \textbf{OSM/Overpass fetch} takes the bounding box as input and returns the corresponding \texttt{.osm} data, with mirror-endpoint failover (\S\ref{app:impl}).
  \item \textbf{CR--SUMO conversion}~\cite{althoff2017commonroad,SUMO2018} transforms the OSM map into a CommonRoad scene (\texttt{.cr.xml}) and a SUMO road network (\texttt{.net.xml}).
  \item \textbf{Microscopic SUMO simulation} populates the network with background traffic according to the vehicle mix and density parsed by the LLM; the resulting trajectories satisfy car-following and lane-changing dynamics by construction.
\end{itemize}

\paragraph{Hallucination isolation.}
Because the LLM never emits map geometry, road-network topology, or vehicle trajectories directly, three classes of hallucination that would otherwise cause silent downstream failure are structurally ruled out:
\begin{itemize}\itemsep0pt\parsep0pt
  \item invalid lanelet IDs or topologies (only the converter produces these);
  \item kinematically infeasible trajectories (only SUMO produces these);
  \item references to non-existent map fragments (only the OSM fetch produces these).
\end{itemize}
The remaining LLM-side failure modes (mis-parsed location, wrong traffic-density level, missing ego specifier) are caught at schema-validation time and either default-applied or surfaced as an error. Per-key pass rates appear in the \emph{loadable} column of Table~\ref{tab:gen-all-results}.

\subsubsection{Query Corpus}
\label{app:gen-corpus-queries}

\begin{figure*}[p]
    \centering
    \makebox[\textwidth][c]{\includegraphics[width=1\textwidth]{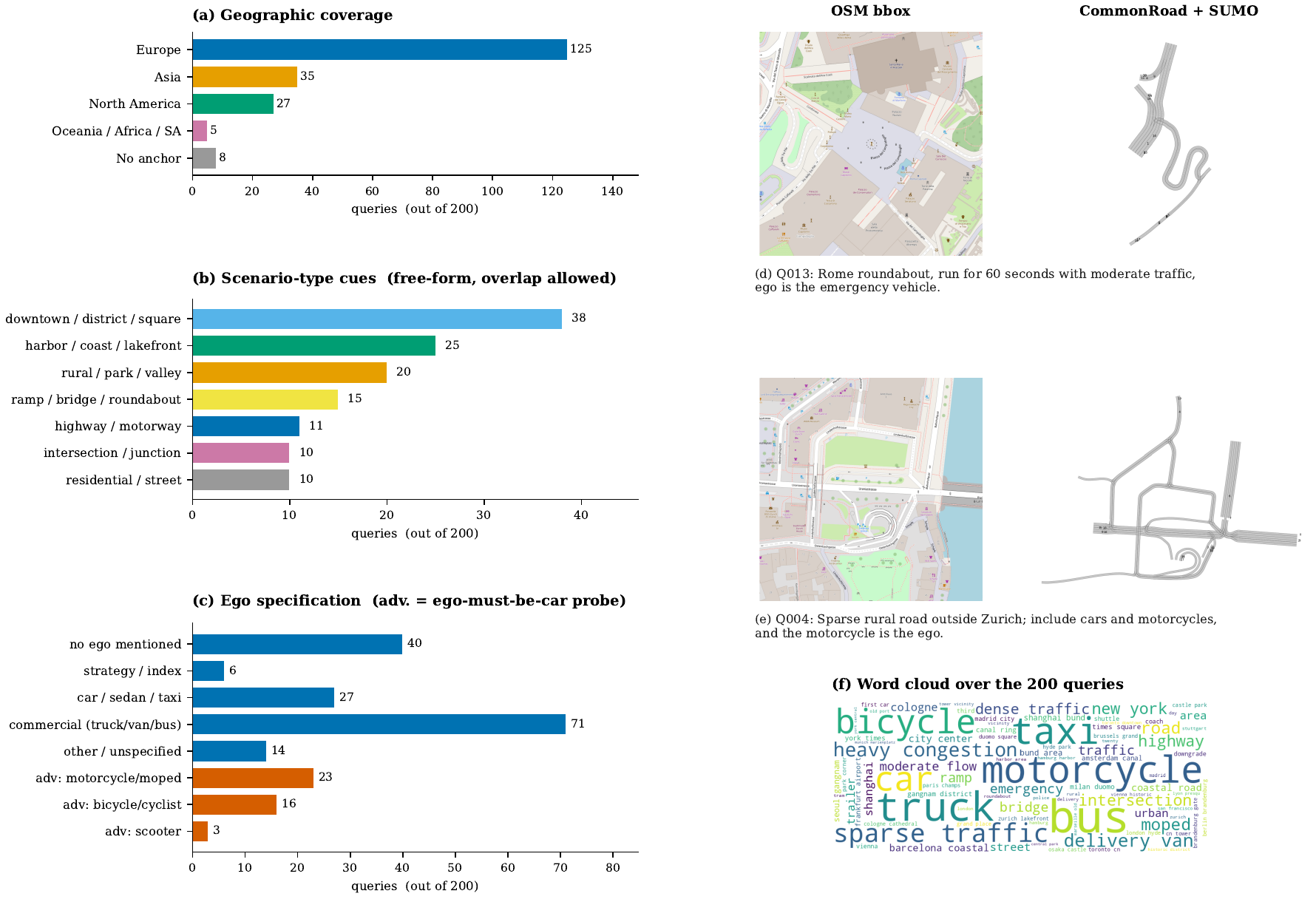}}
    \caption{Composition of the 200-query Scenario Generation corpus. (a) Geographic coverage. (b) Scenario-type cues (free-form, overlapping). (c) Ego specification, with non-car ego queries in vermillion. (d, e) Two example queries rendered through the pipeline: Q013 (clean commercial ego) and Q004 (adversarial non-car ego). (f) Word cloud over the queries.}
    \label{fig:gen-corpus-composition}
\end{figure*}

The Generation Module benchmark uses $N = 200$ natural-language queries with a median length of 8 words (p10 = 5, p90 = 11, max = 23). Frequently requested cities include Shanghai, Madrid, Cologne, Seoul, and London. The corpus splits into two reporting buckets:
\begin{itemize}\itemsep0pt\parsep0pt
    \item \textbf{CLEAN} ($n=158$) -- standard car-ego queries. These exercises the canonical pipeline and dominate the headline numbers.
    \item \textbf{ADVERSARIAL} ($n=42$) -- queries that explicitly request a non-car ego, split as motorcycle/moped (23), bicycle/cargo bike/cyclist (16), and scooter/e-scooter (3). This bucket probes the system rule that the Frenetix-planned ego \emph{must} be a car: the CP rules teach the LLM to keep the requested vehicle type in the surrounding traffic while downgrading the ego itself to the first car.
\end{itemize}
A separate cross-cutting count: 55 queries mention a traffic-density adjective (system-fixed default: \texttt{low}) and 21 request an explicit duration (system-fixed default: \texttt{20\,s}), probing the defaults-compliance metric defined in §\ref{app:gen-metrics}.

Figure~\ref{fig:gen-corpus-composition} summarises the geographic coverage, scenario-type cues, and ego specification. It also includes a word cloud of the queries.
\begin{table*}[htbp]
\centering
\scriptsize
\setlength{\tabcolsep}{3.2pt}
\renewcommand{\arraystretch}{1.02}
\caption{Performance of the Generation Module on the ALL bucket ($N{=}200$ per cell). }
\label{tab:gen-all-results}
\begin{tabular}{@{}llcccccccc@{}}
\toprule
\multirow{2}{*}{\textbf{Model}} & \multirow{2}{*}{\textbf{Prompt}} & \multicolumn{2}{c}{\textbf{Cost}} & \multicolumn{5}{c}{\textbf{Pipeline}} & \multirow{2}{*}{\textbf{Overall $\uparrow$}} \\
\cmidrule(lr){3-4} \cmidrule(lr){5-9}
 & & tokens $\downarrow$ & latency (s) $\downarrow$ & JSON \% $\uparrow$ & defaults $\uparrow$ & load \% $\uparrow$ & traffic \% $\uparrow$ & runnable \% $\uparrow$ & \\
\midrule
\rowcolor[HTML]{F0F0F0}
\multicolumn{10}{@{}l}{\emph{Cloud API}} \\
\multirow{5}{*}{\textcolor{blue}{\textbf{Qwen3.6-plus}}} & baseline & 749 & 25.8 & 100.0 & 0.497 & 90.5 & 90.5 & 90.5 & \cellcolor{blue!15}{0.780} \\
 & cp & 1970 & 27.0 & 100.0 & 1.000 & 96.0 & 96.0 & 96.0 & \cellcolor{blue!15}{0.955} \\
 & cp\_cot & 2827 & 33.7 & 100.0 & 1.000 & 96.5 & 96.5 & 96.5 & \cellcolor{blue!15}{\underline{0.956}} \\
 & cp\_icl & 3750 & 27.8 & 100.0 & 1.000 & 96.5 & 96.5 & 96.5 & \cellcolor{blue!15}{\underline{0.956}} \\
 & cp\_icl\_cot & 4480 & 32.1 & 100.0 & 1.000 & 96.5 & 96.5 & 96.5 & \cellcolor{blue!15}{\underline{0.956}} \\
\cmidrule(l){1-10}
\multirow{5}{*}{\textbf{Deepseek-v3.2}} & baseline & 724 & 29.1 & 100.0 & 0.497 & 94.0 & 94.0 & 94.0 & \cellcolor{blue!15}{0.784} \\
 & cp & 1912 & 28.3 & 100.0 & 1.000 & 96.0 & 96.0 & 96.0 & \cellcolor{blue!15}{0.955} \\
 & cp\_cot & 2684 & 32.9 & 100.0 & 1.000 & 96.5 & 96.5 & 96.5 & \cellcolor{blue!15}{\underline{0.956}} \\
 & cp\_icl & 3556 & 28.8 & 100.0 & 1.000 & 96.5 & 96.5 & 96.5 & \cellcolor{blue!15}{\underline{0.956}} \\
 & cp\_icl\_cot & 4318 & 34.3 & 100.0 & 1.000 & 96.5 & 96.5 & 96.5 & \cellcolor{blue!15}{\underline{0.956}} \\
\cmidrule(l){1-10}
\multirow{5}{*}{\textcolor{orange}{\textbf{Glm-5}}} & baseline & 671 & 28.6 & 100.0 & 0.500 & 92.0 & 91.5 & 92.0 & \cellcolor{blue!15}{0.783} \\
 & cp & 1850 & 30.0 & 100.0 & 1.000 & 95.0 & 95.0 & 95.0 & \cellcolor{blue!15}{0.954} \\
 & cp\_cot & 2601 & 36.5 & 100.0 & 1.000 & 97.0 & 97.0 & 97.0 & \cellcolor{blue!15}{\textbf{0.957}} \\
 & cp\_icl & 3436 & 36.6 & 100.0 & 1.000 & 96.0 & 96.0 & 95.5 & \cellcolor{blue!15}{0.955} \\
 & cp\_icl\_cot & 4204 & 41.1 & 100.0 & 1.000 & 96.5 & 96.5 & 96.5 & \cellcolor{blue!15}{\underline{0.956}} \\
\cmidrule(l){1-10}
\multirow{5}{*}{\textcolor{teal}{\textbf{Gemini-3-flash}}} & baseline & 763 & 29.7 & 100.0 & 0.497 & 95.0 & 95.0 & 95.0 & \cellcolor{blue!15}{0.787} \\
 & cp & 2026 & 26.7 & 100.0 & 1.000 & 95.5 & 95.5 & 95.5 & \cellcolor{blue!15}{0.954} \\
 & cp\_cot & 2822 & 25.1 & 100.0 & 1.000 & 97.0 & 97.0 & 97.0 & \cellcolor{blue!15}{\textbf{0.957}} \\
 & cp\_icl & 3890 & 23.7 & 100.0 & 1.000 & 96.5 & 96.5 & 96.5 & \cellcolor{blue!15}{\underline{0.956}} \\
 & cp\_icl\_cot & 4675 & 24.6 & 100.0 & 1.000 & 96.5 & 96.5 & 96.5 & \cellcolor{blue!15}{\underline{0.956}} \\
\cmidrule(l){1-10}
\multirow{5}{*}{\textcolor{violet}{\textbf{Gpt-5.4-mini}}} & baseline & 609 & 39.9 & 100.0 & 0.495 & 75.5 & 75.5 & 75.5 & \cellcolor{blue!15}{0.763} \\
 & cp & 1780 & 34.2 & 99.5 & 0.892 & 94.0 & 94.0 & 94.0 & \cellcolor{blue!15}{0.912} \\
 & cp\_cot & 2517 & 40.5 & 100.0 & 0.995 & 95.0 & 95.0 & 95.0 & \cellcolor{blue!15}{0.950} \\
 & cp\_icl & 3382 & 31.0 & 100.0 & 1.000 & 89.5 & 89.5 & 89.5 & \cellcolor{blue!15}{0.937} \\
 & cp\_icl\_cot & 4122 & 34.5 & 100.0 & 1.000 & 95.0 & 95.0 & 95.0 & \cellcolor{blue!15}{0.952} \\
\midrule
\rowcolor[HTML]{F0F0F0}
\multicolumn{10}{@{}l}{\emph{Local Ollama}} \\
\multirow{5}{*}{\shortstack[l]{\textcolor{darkgray}{\textbf{Qwen3.6:35b}} \\ \textcolor{darkgray}{\textbf{(Think)}}}} & baseline & 3322 & 53.4 & 100.0 & 0.497 & 49.5 & 49.5 & 49.5 & \cellcolor{blue!15}{0.735} \\
 & cp & 3352 & 46.5 & 97.0 & 0.995 & 94.0 & 94.0 & 94.0 & \cellcolor{blue!15}{0.927} \\
 & cp\_cot & 4159 & 52.0 & 94.0 & 1.000 & 90.5 & 90.5 & 90.5 & \cellcolor{blue!15}{0.899} \\
 & cp\_icl & 4528 & 36.8 & 99.5 & 1.000 & 96.0 & 96.0 & 96.0 & \cellcolor{blue!15}{0.951} \\
 & cp\_icl\_cot & 5469 & 43.3 & 99.0 & 1.000 & 95.5 & 95.5 & 95.5 & \cellcolor{blue!15}{0.947} \\
\cmidrule(l){1-10}
\multirow{5}{*}{\textcolor{olive}{\textbf{Qwen3.6:35b}}} & baseline & 753 & 20.1 & 100.0 & 0.495 & 44.0 & 44.0 & 44.0 & \cellcolor{blue!15}{0.728} \\
 & cp & 1972 & 25.5 & 100.0 & 1.000 & 96.5 & 96.5 & 95.5 & \cellcolor{blue!15}{\underline{0.956}} \\
 & cp\_cot & 2778 & 29.0 & 100.0 & 1.000 & 96.5 & 96.5 & 96.5 & \cellcolor{blue!15}{\underline{0.956}} \\
 & cp\_icl & 3741 & 26.8 & 100.0 & 1.000 & 96.0 & 96.0 & 96.0 & \cellcolor{blue!15}{\underline{0.956}} \\
 & cp\_icl\_cot & 4638 & 33.0 & 100.0 & 1.000 & 96.0 & 96.0 & 96.0 & \cellcolor{blue!15}{0.955} \\
\cmidrule(l){1-10}
\multirow{5}{*}{\shortstack[l]{\textcolor{brown}{\textbf{Gemma4:31b}} \\ \textcolor{brown}{\textbf{(Think)}}}} & baseline & 1585 & 40.3 & 100.0 & 0.497 & 93.0 & 93.0 & 93.0 & \cellcolor{blue!15}{0.784} \\
 & cp & 2813 & 38.9 & 100.0 & 1.000 & 94.5 & 94.5 & 94.5 & \cellcolor{blue!15}{0.950} \\
 & cp\_cot & 3572 & 43.1 & 100.0 & 1.000 & 96.0 & 96.0 & 96.0 & \cellcolor{blue!15}{0.955} \\
 & cp\_icl & 4408 & 35.7 & 100.0 & 1.000 & 95.0 & 95.0 & 95.0 & \cellcolor{blue!15}{0.953} \\
 & cp\_icl\_cot & 5004 & 37.2 & 100.0 & 1.000 & 96.5 & 96.5 & 96.5 & \cellcolor{blue!15}{\underline{0.956}} \\
\cmidrule(l){1-10}
\multirow{5}{*}{\textcolor{magenta}{\textbf{Gemma4:31b}}} & baseline & 785 & 28.8 & 100.0 & 0.497 & 94.0 & 94.0 & 94.0 & \cellcolor{blue!15}{0.785} \\
 & cp & 2044 & 27.8 & 100.0 & 1.000 & 92.0 & 92.0 & 92.0 & \cellcolor{blue!15}{0.946} \\
 & cp\_cot & 2790 & 31.2 & 100.0 & 1.000 & 96.0 & 96.0 & 96.0 & \cellcolor{blue!15}{0.955} \\
 & cp\_icl & 3914 & 28.0 & 100.0 & 1.000 & 91.0 & 91.0 & 91.0 & \cellcolor{blue!15}{0.942} \\
 & cp\_icl\_cot & 4649 & 31.7 & 100.0 & 1.000 & 96.5 & 96.5 & 96.5 & \cellcolor{blue!15}{\underline{0.956}} \\
\cmidrule(l){1-10}
\multirow{5}{*}{\shortstack[l]{\textcolor{cyan}{\textbf{Gpt-oss:20b}} \\ \textcolor{cyan}{\textbf{(Think)}}}} & baseline & 1894 & 33.6 & 100.0 & 0.485 & 86.0 & 86.0 & 86.0 & \cellcolor{blue!15}{0.771} \\
 & cp & 2545 & 32.5 & 100.0 & 1.000 & 96.5 & 96.5 & 96.5 & \cellcolor{blue!15}{\underline{0.956}} \\
 & cp\_cot & 3472 & 32.7 & 100.0 & 1.000 & 96.0 & 96.0 & 96.0 & \cellcolor{blue!15}{0.955} \\
 & cp\_icl & 4016 & 31.4 & 100.0 & 1.000 & 96.5 & 96.5 & 96.0 & \cellcolor{blue!15}{\underline{0.956}} \\
 & cp\_icl\_cot & 4816 & 31.1 & 100.0 & 1.000 & 96.5 & 96.5 & 96.5 & \cellcolor{blue!15}{\underline{0.956}} \\
\bottomrule
\end{tabular}
\vspace{1ex}
{\flushleft\footnotesize\textit{Note:} Metrics are split into Cost, Pipeline, and a standalone Overall summary column. Arrows mark preferred directions ($\uparrow$/$\downarrow$). In the Overall column, \textbf{bold} marks the
best score across the table and \underline{underline} marks the second-best. Models are grouped by provider and ordered within each provider by their peak \texttt{cp\_icl\_cot} overall score. Overall and default values are bounded in $[0,1]$; tokens and latency (s) are means per query.\par}
\end{table*}

\subsubsection{Evaluation Metrics}
\label{app:gen-metrics}

We evaluate the generation chain (\emph{query $\to$ JSON $\to$ Scenario $\to$ Planner}) across three buckets: \textbf{ALL} ($n{=}200$), \textbf{CLEAN} ($n{=}158$), and \textbf{ADVERSARIAL} ($n{=}42$; requiring ego-must-be-car rule). Table~\ref{tab:gen-all-results} tracks nine metrics clustered into three groups:

\begin{itemize}\itemsep0pt\parsep0pt
    \item \textbf{Cost:} Mean \textit{Tokens} (prompt+completion+reasoning) and \textit{Latency} (s) per query.
    \item \textbf{Pipeline:} Success rate at each stage: \textit{JSON valid \%} (schema-compliant output); \textit{Defaults compliance} (adherence to fixed settings like \texttt{sim.duration\_s=20}); \textit{Load \%} (CommonRoad scene parses successfully); \textit{Traffic \%} (loadable with SUMO $\ge 1$ background actor); \textit{Runnable \%} (Frenetix completes a trajectory, even if colliding); and \textit{Planner \%} (Frenetix reaches the goal safely).
    \item \textbf{Overall:} The composite \textit{Intent overall score} aggregating LLM and pipeline metrics, serving as the primary metric to rank (model, condition) pairs.
\end{itemize}

\subsubsection{Evaluation Results}
\label{app:gen-results}

Each cell aggregates 200 generation runs end-to-end (geocode $\to$ OSM fetch $\to$ CR convert $\to$ SUMO sim $\to$ Frenetix). We surface the full results matrix on the \textbf{ALL} bucket in Table~\ref{tab:gen-all-results}.

\paragraph{Ablation discussion.} The prompt-condition ladder reveals that intent parsing is gated almost entirely by the schema injection (\texttt{cp}): the \emph{defaults-compliance} column jumps from $\approx$0.50 to 1.00 for nine out of ten models the moment the JSON schema and the defaults are spelled out. Once the schema is in the prompt, adding CoT or ICL yields only marginal gains, and the four \textsc{ALL}-bucket overall scores converge to a tight 0.95--0.96 band across model families. The implication is that for structured-output tasks with a narrow grammar, contextual prompting alone is sufficient. The extra latency cost of CoT scaffolding (1.5--2$\times$ tokens per query) buys negligible additional reliability here, which is why small open-source models such as Qwen3.6:35b and Gemma4:31b match commercial APIs on this task.

\subsection{Scenario Selection Module}
\label{app:sel-corpus}
\subsubsection{Query Corpus}
\label{app:sel-corpus-queries}

The Selection Module benchmark uses 200 natural-language scenario-selection queries (median length: 16 words), each targeting a unique source scenario from the CommonRoad database. The structured \gls{GT} targets are notably narrow, reflecting realistic, specific user requests: 29.5\% resolve to exactly one matching scenario, and 32.5\% resolve to a tight set of 4--10 alternatives.

Each fixture's \gls{GT} JSON defines 5 extractor slots (as detailed in Table~\ref{tab:sel-slots}), corresponding to the fields the LLM pipeline attempts to extract. For each query, we count how many of these 5 slots are populated:
\begin{itemize}\itemsep0pt\parsep0pt
    \item \textbf{5/5 fields (14 queries):} The user specifies all five aspects (e.g., ``On a 2-lane road in Munich at $\sim$25 km/h, ego approaches a pedestrian crossing with 2 vehicles ahead'').
    \item \textbf{4/5 fields (111 queries):} Most common; one slot is omitted (typically obstacles or velocity).
    \item \textbf{3/5 fields (74 queries):} Three slots specified (e.g., location, tags, and road network).
    \item \textbf{2/5 fields (1 query):} An outlier with only two slots populated.
\end{itemize}

\begin{table}[ht]
\centering
\footnotesize
\renewcommand{\arraystretch}{1.1}
\caption{The 5 extractor slots in the Scenario Selection \gls{GT}. ``Pop.'' is the non-empty count across $N=200$ fixtures.}
\label{tab:sel-slots}
\begin{tabular}{@{}llc@{}}
\toprule
\textbf{Slot} & \textbf{Description (e.g.)} & \textbf{Pop.} \\
\midrule
\texttt{location}(req)  & Country/city (\texttt{\{"DEU"\}}) & 200 \\
\texttt{tags}(req)      & Scenario tags (\texttt{["urban"]}) & 200 \\
\texttt{road\_net}(opt) & Topology (\texttt{\{lanes: 2\}}) & 68 \\
\texttt{obstacles}(opt) & Actors (\texttt{\{car: [1, 2]\}}) & 190 \\
\texttt{velocity}(opt)  & Ego speed (\texttt{[10, 20]}) & 80 \\
\bottomrule
\end{tabular}
\end{table}

\begin{figure}[ht]
    \centering
    \begin{subfigure}[t]{\columnwidth}
        \centering
        \includegraphics[width=0.85\linewidth]{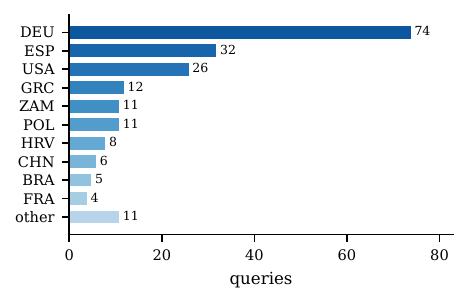}
        \caption{Country distribution (\texttt{DEU} dominates).}
        \label{fig:sel-corpus-a}
    \end{subfigure}
    
    \vspace{0.4em}
    \begin{subfigure}[t]{\columnwidth}
        \centering
        \includegraphics[width=0.95\linewidth]{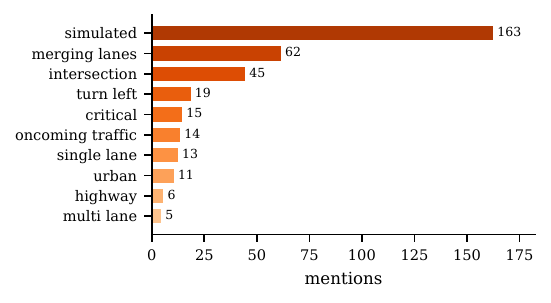}
        \caption{Top-10 \gls{GT} \texttt{scenarioTags}.}
        \label{fig:sel-corpus-b}
    \end{subfigure}
    
    \vspace{0.4em}
    \begin{subfigure}[t]{\columnwidth}
        \centering
        \includegraphics[width=0.85\linewidth]{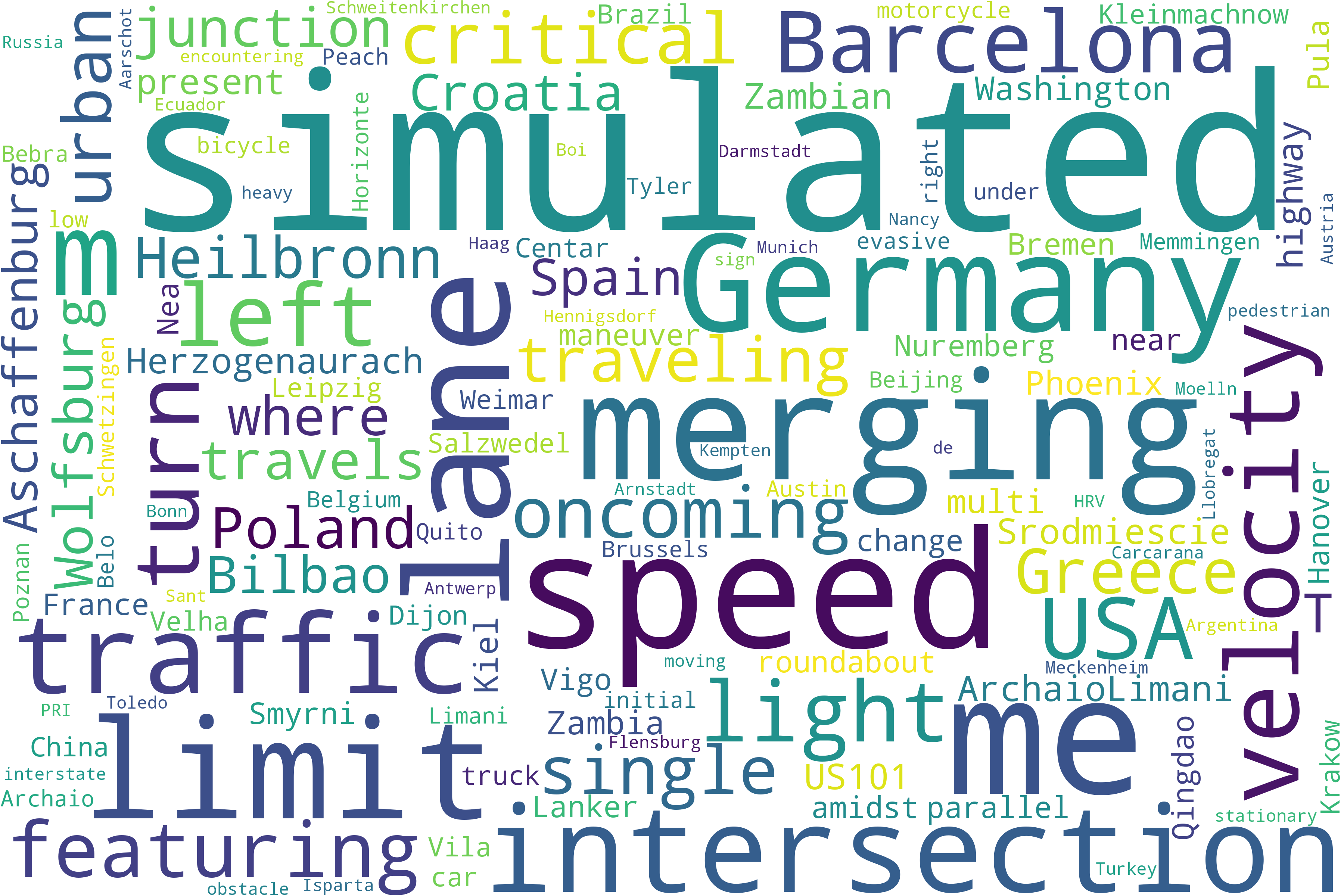}
        \caption{Word cloud (stopwords/function words removed).}
        \label{fig:sel-corpus-c}
    \end{subfigure}
    \caption{Composition of the 200-query Scenario Selection Module. Per-slot field coverage is reported separately in Table~\ref{tab:sel-slots}.}
    \label{fig:sel-corpus-composition}
\end{figure}

Per-slot field coverage is reported in Table~\ref{tab:sel-slots}. Figure~\ref{fig:sel-corpus-composition} visualizes the corpus along three axes: country distribution, \gls{GT} \texttt{scenarioTags} frequency, and a word cloud over the queries.

\paragraph{Retrieval methods.}
Given the structured extraction (the five \gls{GT} slots in Table~\ref{tab:sel-slots}) and the original natural-language (\gls{NL}) query, we evaluate four retrieval strategies that combine these signals against the ChromaDB scenario index in increasingly hybrid ways:
\begin{itemize}\itemsep0pt\parsep0pt
    \item \textbf{Funnel pipeline:} A strict 5-stage \emph{AND} filter over the extracted \gls{GT} slots (\texttt{location} $\to$ \texttt{tags} $\to$ \texttt{road\_net} $\to$ \texttt{obstacles} $\to$ \texttt{velocity}). It applies predicates as a left-to-right intersection over the scenario index, where each stage's output feeds the next, and empty intermediate results abort the pipeline. Velocity uses a tolerant range comparison (single-element bounds widen to $\pm 2$\,m/s); tags use case-insensitive matching against the controlled taxonomy; and locations fall back to country-only matching if the city specifier yields zero hits. This approach offers high precision but is brittle to noisy extraction.
    \item \textbf{Semantic search:} Computes ChromaDB cosine similarity over SentenceTransformer embeddings of the \gls{NL} query. It is language-tolerant but provides no structural guarantees.
    \item \textbf{Hybrid pre-filtering:} Semantic retrieval restricted to candidates whose metadata matches the extracted \texttt{country\_code}, ensuring grounded recall and country-safe results.
    \item \textbf{Reciprocal-rank fusion (RRF):} Fuses the top-$k$ results from both the \texttt{funnel} and \texttt{semantic} pipelines to balance strict metadata matching with semantic similarity.
\end{itemize}

\subsubsection{Evaluation Metrics}
\label{app:sel-metrics}

We evaluate the selection chain (\emph{query $\to$ structured extraction $\to$ retrieval $\to$ top-$k$ scenarios}) on the $N{=}200$ fixtures with cached \texttt{gt\_valid\_ids} (the programmatic answer key derived from the \gls{GT} structured conditions). Reported metrics fall into three groups: \textbf{Cost}, \textbf{Retrieval quality}, and \textbf{Extraction quality}.

\paragraph{Cost.} Mean \textit{Tokens} per query (prompt+completion+reasoning) and \textit{Latency} (s).

\paragraph{Retrieval quality.} The three retrieval-quality metrics share a common test (``\emph{does the returned scenario satisfy the \gls{GT} structured conditions?}'') but answer different operational questions:
\begin{itemize}\itemsep0pt\parsep0pt
    \item \texttt{sat\_any@5} $\to$ production-faithful (top-$k$ UI, one valid match = success).
    \item \texttt{sat\_top1} $\to$ strictest user-visible (single-result UIs).
    \item \textbf{\texttt{sat\_all@returned}} $\to$ precision-strict headline (deployed scoring contract; ranks variants in Table~\ref{tab:sel-all-results}).
\end{itemize}
By default, these are reported on the \texttt{funnel} pipeline. Both \texttt{sat\_any@5} and \texttt{sat\_all@returned} are additionally broken down across the four retrieval pipelines (\texttt{funnel}, \texttt{semantic}, \texttt{hybrid\_prefilter}, \texttt{rrf}) in Figure~\ref{fig:sel-retrieval-methods} to isolate structured filtering from semantic search.

\paragraph{Extraction quality.} \textit{Extract} is the mean of five per-slot extractor scores in $[0, 1]$, computed by comparing the LLM's structured output to the \gls{GT} slot-by-slot. It is \emph{independent of retrieval} and isolates pure LLM extraction quality. The per-slot scoring functions are:
\begin{itemize}\itemsep0pt\parsep0pt
    \item \texttt{location} $\to$ exact \texttt{country\_code} + fuzzy \texttt{specifier}.
    \item \texttt{tags} $\to$ Jaccard set overlap.
    \item \texttt{road\_net} $\to$ field-wise topology agreement.
    \item \texttt{obstacles} $\to$ range/count overlap.
    \item \texttt{velocity} $\to$ numeric range overlap ($\pm 2$\,m/s for single-element bounds).
\end{itemize}

\paragraph{Diagnosing failures with \textit{Extract} vs.\ \textit{sat\_all}.} The two columns decouple two distinct failure modes (see Table~\ref{tab:sel-all-results}): high \textit{Extract} with low \textit{sat\_all} indicates correct extraction followed by a low-precision retrieval stage that admits spurious items into the returned list (a candidate for retrieval-side refinement). Low \textit{Extract} with high \textit{sat\_all} indicates a noisy LLM whose extraction errors are masked by a permissive downstream filter. We therefore report both columns. We rank \textbf{(model, condition)} pairs by funnel \textit{sat\_all@returned}, the precision-strict headline metric: every scenario the system returns must satisfy the user's structured request.

\subsubsection{Evaluation Results}
\label{app:sel-results}

\begin{figure*}[htbp]
    \centering
    \includegraphics[width=\linewidth]{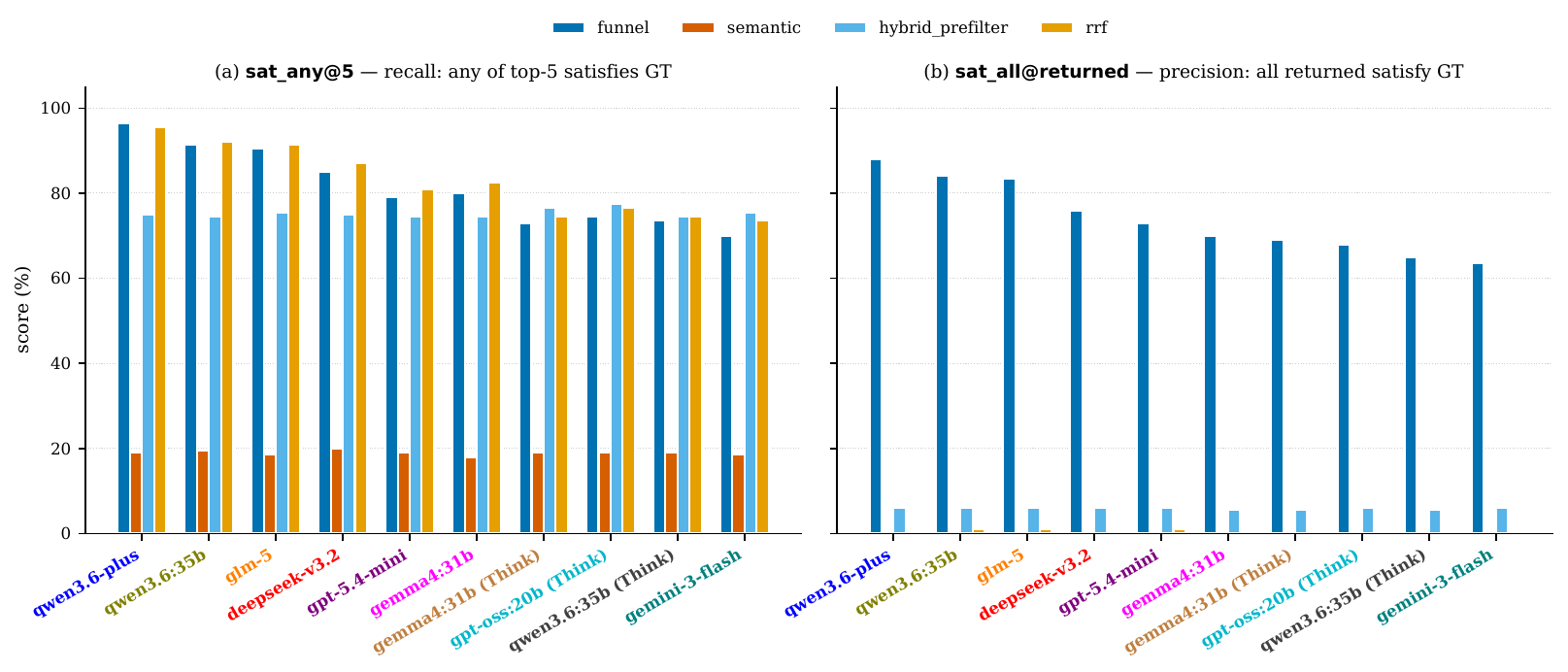}
    \caption{Cross-pipeline retrieval comparison under two scoring contracts: \textbf{(a)} recall-oriented \texttt{sat\_any@5}; \textbf{(b)} precision-strict \texttt{sat\_all@returned}. Each model shown at its best funnel condition; x-axis order shared across panels; colours match Table~\ref{tab:sel-all-results}.}
    \label{fig:sel-retrieval-methods}
\end{figure*}

\paragraph{Experimental design.} We evaluate a grid of ten model variants, five prompt conditions, and four retrieval methods over the same $N{=}200$ fixtures, with \texttt{gt\_valid\_ids} cached as the programmatic answer key. Since the four retrieval methods consume the same LLM-extracted slots per (model, condition) call, the experiment requires 10,000 LLM extractions but yields 40,000 (cell, fixture) retrieval outcomes. The two reporting artifacts surface complementary slices: Table~\ref{tab:sel-all-results} gives every (model, condition) pair on the production \texttt{funnel} pipeline (50 cells), while Figure~\ref{fig:sel-retrieval-methods} reports \texttt{sat\_any@5} across all four pipelines at each model's best prompt condition (40 cells). Each cell aggregates 200 end-to-end runs: \gls{NL} query $\to$ slot extraction $\to$ retrieval $\to$ top-$k$ scoring against the cached answer key.

\paragraph{Reading Table~\ref{tab:sel-all-results}.} The metrics are split into two clusters: \textbf{Cost} (tokens, latency) and \textbf{Retrieval (funnel)}, where the latter also embeds the LLM-only \textit{Extract} column so that the LLM-vs.-retrieval failure-mode comparison is visible at a glance (cf.\ \S\ref{app:sel-metrics}). The shaded column \textit{sat\_all} (\texttt{sat\_all@returned}) is the precision-strict headline we use to rank variants: every returned scenario must satisfy the \gls{GT} structured conditions. Models are grouped by provider and listed in the same order as Table~\ref{tab:llm-models}. Within each model family, the \texttt{(Think)} (reasoning-on) variant precedes the default. The cross-pipeline comparison across all four retrieval methods is shown in Figure~\ref{fig:sel-retrieval-methods}.

\begin{table*}[htbp]
\centering
\scriptsize
\setlength{\tabcolsep}{3.2pt}
\renewcommand{\arraystretch}{1.02}
\caption{Performance of the Selection Module on the \texttt{funnel} pipeline ($N{=}200$ per cell). \textit{sat\_all}~\% denotes the joint satisfaction rate across all five GT slots (location $\to$ tags $\to$ road network $\to$ obstacles $\to$ velocity) for the returned top-$k$ scenario list; we report it as the precision-strict headline metric.}
\label{tab:sel-all-results}
\begin{tabular}{@{}llccccccc@{}}
\toprule
\multirow{2}{*}{\textbf{Model}} & \multirow{2}{*}{\textbf{Prompt}} & \multicolumn{2}{c}{\textbf{Cost}} & \multirow{2}{*}{{\textbf{Fail\% $\downarrow$}}} & \multicolumn{3}{c}{\textbf{Retrieval (funnel)}} & \multirow{2}{*}{\textbf{Overall $\uparrow$}} \\
\cmidrule(lr){3-4} \cmidrule(lr){6-8}
 & & tokens $\downarrow$ & latency (s) $\downarrow$ & & sat\_any@5 \% $\uparrow$ & sat\_top1 \% $\uparrow$ & extract $\uparrow$ & \textbf{sat\_all \% $\uparrow$} \\
\midrule
\rowcolor[HTML]{F0F0F0}
\multicolumn{9}{@{}l}{\emph{Cloud API}} \\
\multirow{5}{*}{\textcolor{blue}{\textbf{Qwen3.6-plus}}} & baseline & 900 & 6.5 & {72.5} & 27.5 & 22.5 & 0.748 & \cellcolor{blue!15}{18.0} \\
 & cp & 4845 & 7.1 & {26.5} & 73.5 & 64.0 & 0.870 & \cellcolor{blue!15}{57.0} \\
 & cp\_cot & 7029 & 21.9 & {3.5} & 96.5 & 93.0 & 0.903 & \cellcolor{blue!15}{88.0} \\
 & cp\_icl & 6621 & 6.4 & {27.5} & 72.5 & 63.5 & 0.858 & \cellcolor{blue!15}{56.0} \\
 & cp\_icl\_cot & 8707 & 21.6 & {3.5} & 96.5 & 90.5 & 0.927 & \cellcolor{blue!15}{83.5} \\
\cmidrule(l){1-9}
\multirow{5}{*}{\textbf{Deepseek-v3.2}} & baseline & 827 & 17.5 & {51.0} & 49.0 & 34.5 & 0.694 & \cellcolor{blue!15}{26.0} \\
 & cp & 4615 & 18.2 & {41.0} & 59.0 & 53.5 & 0.831 & \cellcolor{blue!15}{48.0} \\
 & cp\_cot & 6602 & 32.8 & {19.5} & 80.5 & 80.0 & 0.902 & \cellcolor{blue!15}{76.0} \\
 & cp\_icl & 6242 & 16.6 & {25.0} & 75.0 & 66.0 & 0.838 & \cellcolor{blue!15}{56.0} \\
 & cp\_icl\_cot & 8158 & 36.4 & {15.0} & 85.0 & 80.5 & 0.903 & \cellcolor{blue!15}{74.5} \\
\cmidrule(l){1-9}
\multirow{5}{*}{\textcolor{orange}{\textbf{Glm-5}}} & baseline & 812 & 14.7 & {85.0} & 15.0 & 12.5 & 0.674 & \cellcolor{blue!15}{9.0} \\
 & cp & 4559 & 13.3 & {29.5} & 70.5 & 62.5 & 0.840 & \cellcolor{blue!15}{56.0} \\
 & cp\_cot & 6482 & 29.1 & {13.0} & 87.0 & 83.0 & 0.865 & \cellcolor{blue!15}{77.5} \\
 & cp\_icl & 6209 & 16.3 & {47.5} & 52.5 & 51.0 & 0.758 & \cellcolor{blue!15}{47.0} \\
 & cp\_icl\_cot & 7974 & 29.1 & {9.5} & 90.5 & 88.5 & 0.886 & \cellcolor{blue!15}{83.5} \\
\cmidrule(l){1-9}
\multirow{5}{*}{\textcolor{teal}{\textbf{Gemini-3-flash}}} & baseline & 861 & 5.7 & {76.5} & 23.5 & 18.0 & 0.685 & \cellcolor{blue!15}{14.0} \\
 & cp & 4835 & 6.4 & {32.5} & 67.5 & 58.0 & 0.824 & \cellcolor{blue!15}{51.0} \\
 & cp\_cot & 6821 & 8.1 & {32.5} & 67.5 & 61.0 & 0.861 & \cellcolor{blue!15}{56.0} \\
 & cp\_icl & 6630 & 5.2 & {35.5} & 64.5 & 58.0 & 0.813 & \cellcolor{blue!15}{53.5} \\
 & cp\_icl\_cot & 8548 & 7.9 & {30.0} & 70.0 & 67.0 & 0.858 & \cellcolor{blue!15}{63.5} \\
\cmidrule(l){1-9}
\multirow{5}{*}{\textcolor{violet}{\textbf{Gpt-5.4-mini}}} & baseline & 834 & 5.8 & {76.5} & 23.5 & 22.5 & 0.612 & \cellcolor{blue!15}{21.5} \\
 & cp & 4569 & 5.6 & {49.5} & 50.5 & 48.0 & 0.802 & \cellcolor{blue!15}{46.0} \\
 & cp\_cot & 6139 & 8.7 & {25.0} & 75.0 & 73.0 & 0.892 & \cellcolor{blue!15}{72.0} \\
 & cp\_icl & 6185 & 6.1 & {51.5} & 48.5 & 43.5 & 0.789 & \cellcolor{blue!15}{38.0} \\
 & cp\_icl\_cot & 7772 & 8.1 & {21.0} & 79.0 & 76.5 & 0.889 & \cellcolor{blue!15}{73.0} \\
\midrule
\rowcolor[HTML]{F0F0F0}
\multicolumn{9}{@{}l}{\emph{Local Ollama}} \\
\multirow{5}{*}{\shortstack[l]{\textcolor{darkgray}{\textbf{Qwen3.6:35b}} \\ \textcolor{darkgray}{\textbf{(Think)}}}} & baseline & 5642 & 34.5 & {59.0} & 41.0 & 35.0 & 0.666 & \cellcolor{blue!15}{26.5} \\
 & cp & 9703 & 36.2 & {26.5} & 73.5 & 68.0 & 0.857 & \cellcolor{blue!15}{65.0} \\
 & cp\_cot & 13356 & 82.2 & {37.0} & 63.0 & 52.5 & 0.777 & \cellcolor{blue!15}{40.0} \\
 & cp\_icl & 11843 & 65.6 & {27.0} & 73.0 & 66.0 & 0.826 & \cellcolor{blue!15}{60.0} \\
 & cp\_icl\_cot & 16857 & 77.7 & {30.5} & 69.5 & 50.0 & 0.763 & \cellcolor{blue!15}{33.5} \\
\cmidrule(l){1-9}
\multirow{5}{*}{\textcolor{olive}{\textbf{Qwen3.6:35b}}} & baseline & 913 & 1.3 & {47.5} & 52.5 & 48.5 & 0.679 & \cellcolor{blue!15}{46.5} \\
 & cp & 4857 & 2.2 & {33.5} & 66.5 & 64.0 & 0.792 & \cellcolor{blue!15}{63.5} \\
 & cp\_cot & 6949 & 7.5 & {9.0} & 91.0 & 87.5 & 0.884 & \cellcolor{blue!15}{83.0} \\
 & cp\_icl & 6822 & 3.9 & {44.0} & 56.0 & 54.5 & 0.779 & \cellcolor{blue!15}{54.0} \\
 & cp\_icl\_cot & 8574 & 6.8 & {8.5} & 91.5 & 88.5 & 0.895 & \cellcolor{blue!15}{84.0} \\
\cmidrule(l){1-9}
\multirow{5}{*}{\shortstack[l]{\textcolor{brown}{\textbf{Gemma4:31b}} \\ \textcolor{brown}{\textbf{(Think)}}}} & baseline & 2915 & 59.6 & {77.0} & 23.0 & 16.0 & 0.645 & \cellcolor{blue!15}{11.0} \\
 & cp & 6487 & 29.1 & {27.0} & 73.0 & 70.5 & 0.830 & \cellcolor{blue!15}{67.0} \\
 & cp\_cot & 8318 & 36.8 & {60.5} & 39.5 & 34.0 & 0.884 & \cellcolor{blue!15}{30.0} \\
 & cp\_icl & 8275 & 29.6 & {27.5} & 72.5 & 72.0 & 0.829 & \cellcolor{blue!15}{69.0} \\
 & cp\_icl\_cot & 10055 & 36.5 & {62.0} & 38.0 & 35.5 & 0.883 & \cellcolor{blue!15}{33.5} \\
\cmidrule(l){1-9}
\multirow{5}{*}{\textcolor{magenta}{\textbf{Gemma4:31b}}} & baseline & 958 & 2.7 & {97.0} & 3.0 & 2.0 & 0.537 & \cellcolor{blue!15}{1.5} \\
 & cp & 4928 & 4.2 & {22.5} & 77.5 & 70.5 & 0.809 & \cellcolor{blue!15}{61.0} \\
 & cp\_cot & 6763 & 12.3 & {20.0} & 80.0 & 72.0 & 0.877 & \cellcolor{blue!15}{65.0} \\
 & cp\_icl & 6728 & 4.0 & {28.5} & 71.5 & 70.5 & 0.830 & \cellcolor{blue!15}{68.0} \\
 & cp\_icl\_cot & 8558 & 12.1 & {27.5} & 72.5 & 71.5 & 0.880 & \cellcolor{blue!15}{70.0} \\
\cmidrule(l){1-9}
\multirow{5}{*}{\shortstack[l]{\textcolor{cyan}{\textbf{Gpt-oss:20b}} \\ \textcolor{cyan}{\textbf{(Think)}}}} & baseline & 3121 & 16.3 & {91.0} & 9.0 & 8.0 & 0.703 & \cellcolor{blue!15}{7.0} \\
 & cp & 5891 & 7.1 & {27.5} & 72.5 & 64.5 & 0.852 & \cellcolor{blue!15}{58.0} \\
 & cp\_cot & 7234 & 9.5 & {49.0} & 51.0 & 37.5 & 0.790 & \cellcolor{blue!15}{32.0} \\
 & cp\_icl & 8123 & 9.6 & {25.5} & 74.5 & 72.0 & 0.862 & \cellcolor{blue!15}{68.0} \\
 & cp\_icl\_cot & 10044 & 14.4 & {48.5} & 51.5 & 46.5 & 0.857 & \cellcolor{blue!15}{39.5} \\
\bottomrule
\end{tabular}
\vspace{1ex}
{\flushleft\footnotesize\textit{Note:} Arrows mark preferred directions ($\uparrow$/$\downarrow$). Tokens and latency (s) are means per query.\par}
\end{table*}

\paragraph{Slot-level extract.}
\label{app:sel-slot-extract}
Table~\ref{tab:sel-all-results} reports a single \textit{extract} mean that hides which of the five slots fails. Table~\ref{tab:sel-slot-extract} decomposes that mean. The bottleneck is \texttt{tags} (e.g.\ \texttt{urban}): it is the lowest slot in all five prompt conditions. Once the schema is injected it sits at 0.64--0.67, while the other four slots range 0.75--0.97. The cause is over-prediction, not misses (recall $99.7\%$): the model adds plausible unsupported descriptors such as \texttt{traffic\_jam} (invented 139 times although no scenario in the database carries it). \gls{CP} alone recovers most of the slot ($0.205{\to}0.665$); no further technique moves it. Under \gls{ICL} without \gls{CoT}, \texttt{velocity} drops $0.957{\to}0.745$ and \texttt{road\_net} drops $0.937{\to}0.864$, likely from copying exemplar values; adding \gls{CoT} restores both. On Qwen3.6-plus, \texttt{cp\_cot} has the lower \textit{extract} ($0.903$) yet the paper's best \texttt{sat\_all} ($88.0$), while \texttt{cp\_icl\_cot} raises \textit{extract} to $0.927$ and lowers \texttt{sat\_all} to $83.5$, because the funnel is a strict AND over slots.
\begin{table}[ht]
{\ifshownew\color{red}\fi
\centering\scriptsize
\setlength{\tabcolsep}{2.8pt}
\caption{Slot-level \textit{extract} score by prompt condition (mean over the ten models, $N{=}200$). The bottom row equals the \textit{extract} column of Table~\ref{tab:sel-all-results} averaged across models.}
\label{tab:sel-slot-extract}
\begin{tabular}{@{}lccccc@{}}
\toprule
\textbf{Extractor slot} & \textbf{\texttt{baseline}} & \textbf{\texttt{cp}} & \textbf{\texttt{cp\_cot}} & \textbf{\texttt{cp\_icl}} & \textbf{\texttt{cp\_icl\_cot}} \\
\midrule
location   & 0.881 & 0.914 & 0.857 & 0.911 & 0.879 \\
\textbf{tags} & \textbf{0.205} & \textbf{0.665} & \textbf{0.648} & \textbf{0.642} & \textbf{0.658} \\
road\_net  & 0.825 & 0.878 & 0.937 & 0.864 & 0.951 \\
obstacles  & 0.745 & 0.947 & 0.918 & 0.928 & 0.910 \\
velocity   & 0.665 & 0.749 & 0.957 & 0.745 & 0.974 \\
\midrule
Mean $=$ \textit{extract} & 0.664 & 0.831 & 0.864 & 0.818 & 0.874 \\
\bottomrule
\end{tabular}}
\end{table}

\paragraph{Cross-pipeline analysis.} Figure~\ref{fig:sel-retrieval-methods} reports both \texttt{sat\_any@5} (recall) and \texttt{sat\_all@returned} (precision) across all four retrieval pipelines at each model's best prompt. Under recall (panel a), the ranking is essentially universal (\texttt{rrf} $\approx$ \texttt{funnel} $\gg$ \texttt{hybrid\_prefilter} $\gg$ \texttt{semantic}), with \texttt{funnel} at 70.0--96.5\% and \texttt{hybrid\_prefilter} flat near 75\%. Any LLM-consulting pipeline can usually surface one valid match in five. Under precision (panel b), only the strict-AND \texttt{funnel} survives at 63.5--88.0\%. The semantic-based pipelines collapse: \texttt{semantic} returns 0\% (no structured constraints), \texttt{hybrid\_prefilter} reaches only 5.5--6.0\% (country pre-filter too weak), and \texttt{rrf} stays below 1.0\% (semantic candidates dilute the funnel list). The Selection Module ships the precision view, so \texttt{funnel} is the production pipeline. \texttt{Rrf}'s parity with \texttt{funnel} under recall is an artefact of the top-5 hit-rate metric.

\subsection{Scenario Modification Module}
\label{app:mod-corpus}
\subsubsection{Query Corpus}
\label{app:mod-corpus-queries}

The Modification Module benchmark uses $N = 200$ natural-language queries per task across four task types: \textbf{T} (trajectory redirection), \textbf{B} (behavior preset), \textbf{P} (population edit), and \textbf{G} (goal extraction), for a total of $800$ queries. Task P is a synthetic union of the population-insertion ($P_{\text{add}}$) and population-deletion ($P_{\text{remove}}$) sub-tasks: 100 queries from each side, yielding $N{=}200$ for P and matching the per-cell sample size used elsewhere. The 800 queries are anchored in 345 unique CommonRoad scenarios spanning 15 country codes (DEU dominates: 80--92 queries per task). The three scenario counts in the paper denote distinct sets: the Selection Module's persistent ChromaDB store indexes 500+ curated scenarios (Appendix \S\ref{app:impl}). The Modification corpus draws from a smaller 245-scenario SUMO-cached pool (with the 5/245 round-trip exclusion below, leaving 240 clean seeds), and the 345 unique anchor scenarios above are the union across all five Modification tasks, larger than the 240 simulable seeds because Task G only edits the planning problem and can use scenarios outside the SUMO-cached pool. Five of 245 cached CommonRoad scenarios (2.0\%) were excluded from the source pool because their \emph{unmodified} form fails to round-trip through the CR$\leftrightarrow$SUMO interface: four due to a missing \texttt{ObstacleType.MOTORCYCLE} mapping in the simulator-to-CommonRoad converter, and one for an unrelated SUMO route-validity issue. These are infrastructure limits of the bridge, not failures of LLM-generated modifications. The LLM never touches them in the sweep. Replacement queries are drawn from the same clean 240-scenario pool to keep the per-task count at exactly $N{=}200$.

\begin{figure*}[p]
    \centering
    \begin{subfigure}[t]{0.25\textwidth}
        \centering
        \includegraphics[width=\linewidth]{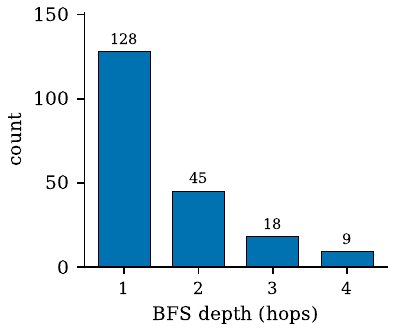}
        \caption{T: BFS depth.}\label{fig:mod-corpus-a}
    \end{subfigure}\hfill
    \begin{subfigure}[t]{0.25\textwidth}
        \centering
        \includegraphics[width=\linewidth]{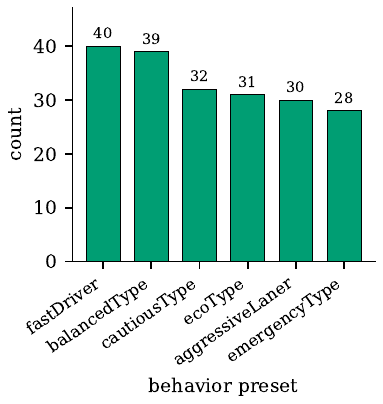}
        \caption{B: behavior preset.}\label{fig:mod-corpus-b}
    \end{subfigure}\hfill
    \begin{subfigure}[t]{0.25\textwidth}
        \centering
        \includegraphics[width=\linewidth]{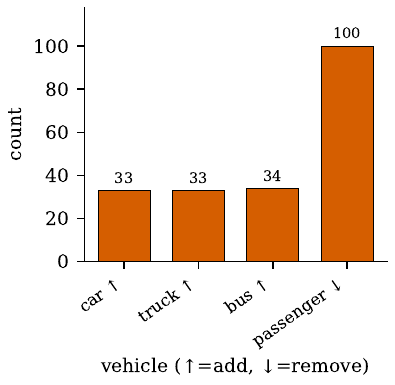}
        \caption{P: \texttt{add}/\texttt{remove} vehicle mix.}\label{fig:mod-corpus-c}
    \end{subfigure}\hfill
    \begin{subfigure}[t]{0.25\textwidth}
        \centering
        \includegraphics[width=\linewidth]{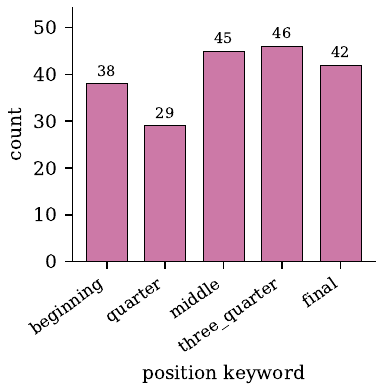}
        \caption{G: position keyword.}\label{fig:mod-corpus-d}
    \end{subfigure}

    \vspace{0.5em}
    \begin{subfigure}[t]{0.25\textwidth}
        \centering
        \includegraphics[width=\linewidth]{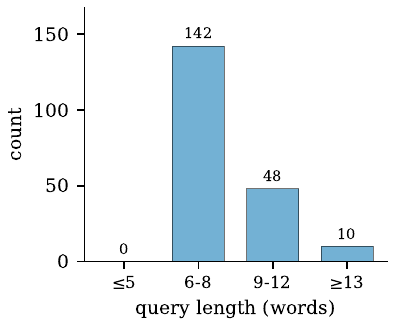}
        \caption{T: query length.}\label{fig:mod-corpus-e}
    \end{subfigure}\hfill
    \begin{subfigure}[t]{0.25\textwidth}
        \centering
        \includegraphics[width=\linewidth]{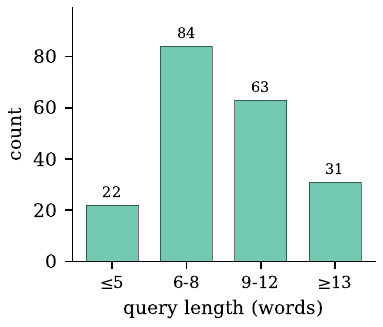}
        \caption{B: query length.}\label{fig:mod-corpus-f}
    \end{subfigure}\hfill
    \begin{subfigure}[t]{0.25\textwidth}
        \centering
        \includegraphics[width=\linewidth]{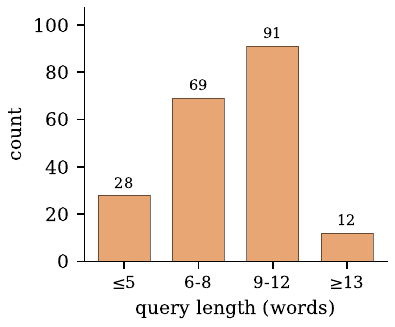}
        \caption{P: query length.}\label{fig:mod-corpus-g}
    \end{subfigure}\hfill
    \begin{subfigure}[t]{0.25\textwidth}
        \centering
        \includegraphics[width=\linewidth]{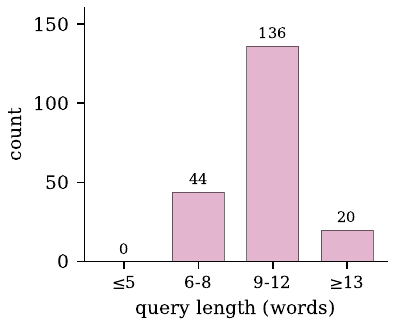}
        \caption{G: query length.}\label{fig:mod-corpus-h}
    \end{subfigure}

    \vspace{0.5em}
    \begin{subfigure}[t]{0.25\textwidth}
        \centering
        \includegraphics[width=\linewidth]{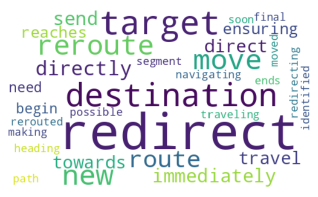}
        \caption{T: word cloud.}\label{fig:mod-corpus-i}
    \end{subfigure}\hfill
    \begin{subfigure}[t]{0.25\textwidth}
        \centering
        \includegraphics[width=\linewidth]{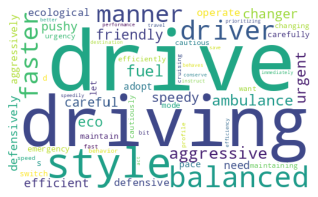}
        \caption{B: word cloud.}\label{fig:mod-corpus-j}
    \end{subfigure}\hfill
    \begin{subfigure}[t]{0.25\textwidth}
        \centering
        \includegraphics[width=\linewidth]{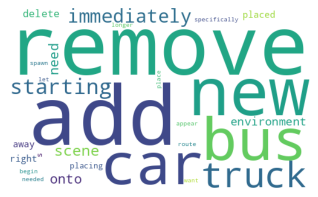}
        \caption{P: word cloud.}\label{fig:mod-corpus-k}
    \end{subfigure}\hfill
    \begin{subfigure}[t]{0.25\textwidth}
        \centering
        \includegraphics[width=\linewidth]{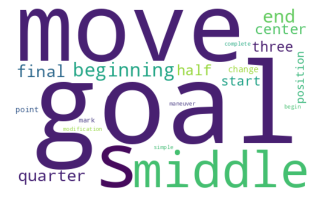}
        \caption{G: word cloud.}\label{fig:mod-corpus-l}
    \end{subfigure}

    \caption{Per-task corpus composition for the Scenario Modification benchmark ($N{=}200$ queries each). \textbf{Row 1 (a--d):} principal stratification axis per task. \textbf{Row 2 (e--h):} query-length distribution. \textbf{Row 3 (i--l):} per-task word cloud.}
    \label{fig:mod-corpus-composition}
\end{figure*}

The \gls{GT} schema is task-specific because each modification target requires different anchor fields: \textbf{T} carries the source vehicle, source edge, target edge, and BFS depth. \textbf{B} carries the target-vehicle list and the expected behavior preset. \textbf{P} carries the $\pm 1$ vehicle-count delta together with the target edge/vehicle and vehicle-type fields. \textbf{G} carries the target edge and the position keyword (start/quarter/middle/three-quarter/end). The canonical \gls{GT} shapes feed the per-task scorers (\S\ref{app:mod-metrics}). The principal corpus-stratification dimensions per task are visualized in Figure~\ref{fig:mod-corpus-composition}.

To illustrate the diversity of query phrasings across tasks, one representative \gls{GT} row per task is shown below (the first query from each task's query set. For P, we list both sub-task examples since the synthetic task interleaves them:
\begin{itemize}\itemsep0pt\parsep0pt
    \item \textbf{T} (\emph{POL\_Krakow-22\_1\_T-4}): ``Redirect vehicle 14 to edge 595.''
        \quad \gls{GT}: \texttt{\{vid:14, source\_edge:332, target\_edge:595, bfs\_depth:1\}}
    \item \textbf{B} (\emph{GRC\_NeaSmyrni-102\_1\_T-6}): ``Set vehicles 20026 and 20019 to urgent driving.''
        \quad \gls{GT}: \texttt{\{target\_vids:[20026,20019], expected\_preset:emergencyType, descriptor\_kind:multi\}}
    \item \textbf{P (add)} (\emph{DEU\_Hanover-44\_29\_T-1}): ``Please add a new truck, starting on edge 874, to the simulation.''
        \quad \gls{GT}: \texttt{\{expected\_delta:+1, target\_edge:874, target\_vtype:truck\}}
    \item \textbf{P (remove)} (\emph{DEU\_Bremen-5\_5\_T-1}): ``Could you please remove vehicle with ID 30243 from the simulation?''
        \quad \gls{GT}: \texttt{\{expected\_delta:-1, target\_vid:30243, target\_vClass:passenger\}}
    \item \textbf{G} (\emph{RUS\_Bicycle-3\_2\_T-1}): ``Set ego goal to 3/4 of lanelet 7.''
        \quad \gls{GT}: \texttt{\{target\_edge:7, position:three\_quarter\}}
\end{itemize}

Figure~\ref{fig:mod-corpus-composition} visualizes the corpus along three axes per task: principal stratification (row 1), query-length distribution (row 2), and a word cloud over the 200 \gls{NL} queries per task (row 3). The behavior-preset distribution (panel b) and the goal-position distribution (panel d) are intentionally near-uniform to ensure the prompt techniques cover every preset and every position keyword. Task T concentrates on short-hop reroutes (1--2 BFS hops cover 86.5\% of queries), reflecting the typical local-edit semantics of trajectory modification requests. Task P's combined panel (c) makes the asymmetry between the add and remove sub-tasks visible: the add side is balanced across car/truck/bus while the remove side is dominated by passenger targets (the first 100 $P_{\text{remove}}$ queries are all passenger), because passenger vehicles are by far the most common non-ego actors in CommonRoad scenarios. The word-cloud row (i--l) highlights the verbs and adjectives that characterize each task: ``redirect/route/target'' for T, ``preset/aggressive/comfort'' for B, ``add/remove/truck/passenger'' for P, ``goal/lanelet/quarter/middle'' for G.

\subsubsection{Evaluation Metrics}
\label{app:mod-metrics}

We evaluate each (model, condition) pair on the $N{=}200$ queries per task with a two-stage scoring funnel: per-task \textbf{semantic correctness} (does the modified scenario reflect the requested change?), followed by \textbf{downstream simulability} (does the modified scenario still produce a valid SUMO trace that round-trips back into CommonRoad?). All checks are boolean. The headline metric \textit{overall} is the unweighted mean of these boolean checks. Per-query and per-cell telemetry (latency and tokens) is auto-recorded alongside.

\paragraph{Shared funnel stages (every task).} The Cost columns (mean tokens, mean latency) and the syntactic/simulability checks below apply uniformly to T, B, P, and G. Task G omits SUMO \% and CR \% because it only edits the planning problem, so there is no traffic round-trip.

\begin{itemize}
    \item \textit{tokens}: Mean total tokens per query (Cost).
    \item \textit{latency (s)}: Mean LLM wall-clock per query (Cost).
    \item \textit{XML \%} / \textit{JSON \%}: Output parses as well-formed SUMO route XML (or JSON for G).
    \item \textit{SUMO \%}: Modified route file simulates end-to-end without runtime errors or stuck vehicles.
    \item \textit{CR \%}: SUMO output round-trips back into a valid CommonRoad scenario via the CR$\to$SUMO bridge.
\end{itemize}

The optional Frenetix-runnability gate is disabled by default in this sweep (the cross-planner Frenetix evaluation is instead reported as the separate batch comparison summarised in Figure~\ref{fig:best_per_task}), so the column does not appear in Tables~\ref{tab:mod-T-all-results}--\ref{tab:mod-P-all-results}.

\paragraph{Per-task semantic checks.} Each task carries its own semantic checks against the per-task \gls{GT} schema described above. For each task, we identify the strictest semantic gate as the \textbf{HEADLINE} metric (the check whose failure most directly indicates that the LLM has not performed the requested edit). The headline column is rendered in red in the corresponding result table.

\noindent\textbf{Task T --- Trajectory Redirection} (Table~\ref{tab:mod-T-all-results}).

\begin{itemize}
    \item \textit{tgt \%}: The named vehicle survives the edit.
    \item \textit{ends \%} --- \textbf{HEADLINE}: The redirected route terminates at the requested target edge.
    \item \textit{preserve \%}: Non-target vehicles' routes remain byte-equivalent to baseline.
\end{itemize}

\noindent\textbf{Task B --- Behaviour Preset} (Table~\ref{tab:mod-B-all-results}).

\begin{itemize}
    \item \textit{tgt \%}: The named vehicle survives the edit.
    \item \textit{preset \%} --- \textbf{HEADLINE}: Modified behaviour-parameter vector matches the requested preset within relative tolerance $10^{-3}$.
    \item \textit{vClass \%}: Vehicle class is preserved (e.g.\ a bus stays a bus).
\end{itemize}

\noindent\textbf{Task P --- Population Edit} ($P_{\text{add}} \cup P_{\text{remove}}$; Table~\ref{tab:mod-P-all-results}). The add and remove sub-task scorers share an indistinguishable arithmetic semantics (both grade the model on producing the requested $\pm 1$ delta while leaving the rest of the route file unchanged) so the combined P overall is the unweighted mean of the boolean checks across all 200 queries.

\begin{itemize}
    \item \textit{count\_match \%} --- \textbf{HEADLINE}: Vehicle count changes by exactly $+1$ (add) or $-1$ (remove). Without this, the requested edit has not happened.
    \item \textit{tgt-chg \%}: New vehicle starts on the requested edge (add) or the named vehicle is absent from the modified file (remove).
    \item \textit{others \%}: Non-target route entries remain byte-equivalent to baseline.
\end{itemize}

\noindent\textbf{Task G --- Goal Extraction} (Table~\ref{tab:mod-G-all-results}).

\begin{itemize}
    \item \textit{edge \%}: A target edge ID is emitted.
    \item \textit{pos-enum \%}: Position keyword belongs to the allowed enumeration.
    \item \textit{edge-\gls{GT} \%} --- \textbf{HEADLINE}: Extracted target lanelet matches the \gls{GT} --- the structural goal decision.
    \item \textit{pos-\gls{GT} \%}: Extracted position keyword matches the \gls{GT}.
    \item \textit{lanelet \%}: Lanelet ID resolves in the CommonRoad scenario.
\end{itemize}

\noindent Two further G checks (the planning-problem-accept check and the shared CR-reload check) are computed but not displayed in Table~\ref{tab:mod-G-all-results} because they are perfectly correlated with \textit{lanelet \%} in this sweep. Both still contribute to \textit{Overall}.

\paragraph{Auto-recorded telemetry.} For each LLM invocation, we record prompt, completion, and (where the provider exposes it) reasoning token counts, the total token count, the per-invocation API latency, and the full per-query wall-clock (which also includes parsing and simulation overhead). These columns appear unaltered in the per-task KPI summaries and let us decouple ``the LLM did its job correctly'' (semantic checks) from ``the LLM was fast enough to be deployable'' (latency/token cost).

\paragraph{Diagnosing failures with the funnel.} The two stages decouple distinct failure modes: high semantic correctness with low SUMO simulability indicates a semantically correct edit that produces a structurally fragile route (e.g.\ a target edge that is connected but disallows the vehicle's class), while uniformly low semantic checks indicate a model that either fails to identify the right anchor field or emits malformed XML/JSON. The two paths, therefore, call for different fixes (network-aware prompting vs.\ stricter output-format prompting), which the per-task per-check tables in \S\ref{app:mod-results} surface independently.

\subsubsection{Evaluation Results}
\label{app:mod-results}

\begin{figure*}[!t]
    \centering
    \includegraphics[width=0.65\textwidth]{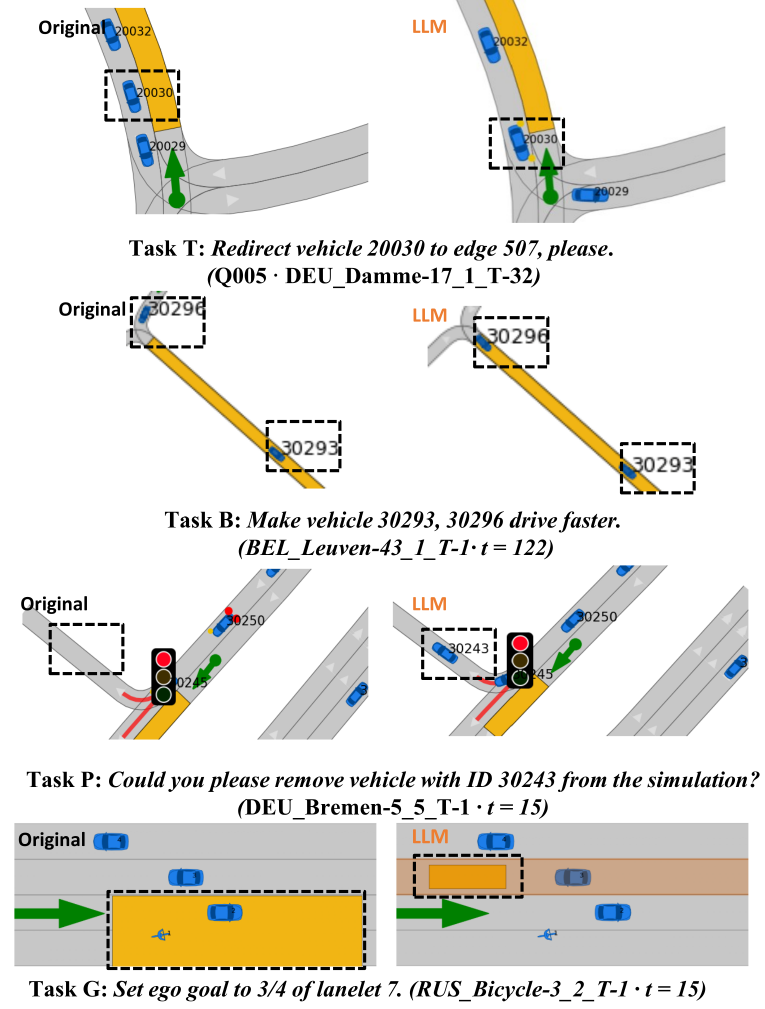}
    \caption{Per-task qualitative diff grid (four rows = four tasks). Each row compares baseline (left) and LLM-modified (right) scenarios at the same mid-trajectory timestep (annotated in the row header). All examples from \texttt{gemini-3-flash-preview} under \texttt{cp\_icl\_cot}.}
    \label{fig:mod-qualitative-grid}
\end{figure*}

\paragraph{Qualitative examples.} Before reporting quantitative outcomes, Figure~\ref{fig:mod-qualitative-grid} shows one representative successful modification per task, all from \texttt{gemini-3-flash-preview} under the \texttt{cp\_icl\_cot} condition. Each row compares the baseline scenario (left) and the LLM-modified scenario (right) rendered at the same mid-trajectory timestep so the modification's effect on the dynamic traffic is directly visible. The four examples surface the visible footprint of each task: in T, the redirected vehicle has joined a different downstream lanelet by $t{=}32$. In B, vehicles 30293 and 30296 have already pulled ahead under the faster behavior preset by $t{=}122$. The P (remove) sub-task shows that vehicle 30242 is removed. G renders the ego's modified goal region on the requested lanelet at $t{=}15$ (G's modification only edits the planning problem, so the surrounding traffic is unchanged by design).

\paragraph{Quantitative results.} Tables~\ref{tab:mod-T-all-results}, \ref{tab:mod-B-all-results}, \ref{tab:mod-P-all-results}, and \ref{tab:mod-G-all-results} report the per-task funnel KPIs for all ten models across the five prompt conditions. Each table covers $N{=}200$ queries per (model, condition) cell (Task P combines the first-100 of $P_{\text{add}}$ with the first-100 of $P_{\text{remove}}$). Cells marked ``N/A'' correspond to open-source models still in the data-collection phase as of submission. The table-generation pipeline will replace them with values for the camera-ready version.

\begin{table*}[htbp]
\centering
\scriptsize
\setlength{\tabcolsep}{3.2pt}
\renewcommand{\arraystretch}{1.02}
\caption{Task T (trajectory redirection) per-cell funnel KPIs ($N{=}200$ queries per cell). The rightmost column \textit{ends \%} is the headline semantic check (the LLM-edited route terminates on the requested target edge); \textit{preserve \%} verifies non-target vehicles' routes stay byte-equivalent to baseline.}
\label{tab:mod-T-all-results}
\begin{tabular}{@{}llccccccccc@{}}
\toprule
\multirow{2}{*}{\textbf{Model}} & \multirow{2}{*}{\textbf{Prompt}} & \multicolumn{2}{c}{\textbf{Cost}} & \multirow{2}{*}{{\textbf{Fail\% $\downarrow$}}} & \multicolumn{6}{c}{\textbf{Pipeline stage pass rates}} \\
\cmidrule(lr){3-4} \cmidrule(lr){6-11}
 & & tokens $\downarrow$ & latency (s) $\downarrow$ & & XML \% $\uparrow$ & tgt \% $\uparrow$ & preserve \% $\uparrow$ & SUMO \% $\uparrow$ & CR \% $\uparrow$ & \cellcolor{blue!15}\textbf{ends \% $\uparrow$} \\
\midrule
\rowcolor[HTML]{F0F0F0}
\multicolumn{11}{@{}l}{\emph{Cloud API}} \\
\multirow{5}{*}{\textcolor{blue}{\textbf{Qwen3.6-plus}}} & baseline & 18324 & 52.6 & {6.5} & 99.5 & 99.5 & 99.0 & 93.5 & 93.5 & \cellcolor{blue!15}{99.0} \\
 & cp & 19336 & 55.2 & {3.5} & 99.5 & 99.5 & 99.0 & 96.5 & 96.5 & \cellcolor{blue!15}{98.5} \\
 & cp\_cot & 21011 & 61.9 & {2.0} & 100.0 & 100.0 & 100.0 & 98.0 & 98.0 & \cellcolor{blue!15}{99.5} \\
 & cp\_icl & 21611 & 50.5 & {4.0} & 100.0 & 100.0 & 100.0 & 96.0 & 96.0 & \cellcolor{blue!15}{98.5} \\
 & cp\_icl\_cot & 24336 & 61.1 & {2.5} & 100.0 & 100.0 & 100.0 & 97.5 & 97.5 & \cellcolor{blue!15}{98.0} \\
\cmidrule(l){1-11}
\multirow{5}{*}{\textbf{Deepseek-v3.2}} & baseline & 14722 & 69.5 & {12.5} & 99.5 & 99.5 & 99.0 & 87.5 & 87.5 & \cellcolor{blue!15}{99.0} \\
 & cp & 15274 & 67.5 & {4.5} & 99.5 & 99.5 & 99.5 & 95.5 & 95.5 & \cellcolor{blue!15}{94.0} \\
 & cp\_cot & 16962 & 83.8 & {0.5} & 99.5 & 99.5 & 99.0 & 99.5 & 99.5 & \cellcolor{blue!15}{97.0} \\
 & cp\_icl & 17067 & 58.5 & {5.0} & 100.0 & 100.0 & 99.5 & 95.0 & 95.0 & \cellcolor{blue!15}{99.0} \\
 & cp\_icl\_cot & 18350 & 68.7 & {0.5} & 100.0 & 100.0 & 100.0 & 99.5 & 99.5 & \cellcolor{blue!15}{99.0} \\
\cmidrule(l){1-11}
\multirow{5}{*}{\textcolor{orange}{\textbf{Glm-5}}} & baseline & 14223 & 36.3 & {23.5} & 99.5 & 99.5 & 99.5 & 76.5 & 76.5 & \cellcolor{blue!15}{98.0} \\
 & cp & 15559 & 43.8 & {3.0} & 99.5 & 99.5 & 99.0 & 97.0 & 97.0 & \cellcolor{blue!15}{97.0} \\
 & cp\_cot & 16568 & 48.8 & {2.5} & 99.5 & 99.5 & 99.0 & 97.5 & 97.5 & \cellcolor{blue!15}{99.5} \\
 & cp\_icl & 17844 & 42.4 & {1.5} & 99.5 & 99.5 & 99.0 & 98.5 & 98.5 & \cellcolor{blue!15}{98.0} \\
 & cp\_icl\_cot & 19004 & 51.5 & {2.0} & 99.5 & 99.5 & 99.5 & 98.0 & 98.0 & \cellcolor{blue!15}{99.5} \\
\cmidrule(l){1-11}
\multirow{5}{*}{\textcolor{teal}{\textbf{Gemini-3-flash}}} & baseline & 18658 & 10.0 & {2.5} & 99.5 & 99.5 & 99.5 & 97.5 & 97.5 & \cellcolor{blue!15}{99.0} \\
 & cp & 19757 & 10.4 & {1.5} & 99.5 & 99.5 & 99.5 & 98.5 & 98.5 & \cellcolor{blue!15}{99.5} \\
 & cp\_cot & 20831 & 12.4 & {2.5} & 98.5 & 98.5 & 98.5 & 97.5 & 97.5 & \cellcolor{blue!15}{98.5} \\
 & cp\_icl & 22225 & 10.4 & {2.0} & 99.5 & 99.5 & 98.5 & 98.0 & 98.0 & \cellcolor{blue!15}{99.5} \\
 & cp\_icl\_cot & 23398 & 12.1 & {2.0} & 99.5 & 99.5 & 99.5 & 98.0 & 98.0 & \cellcolor{blue!15}{99.5} \\
\cmidrule(l){1-11}
\multirow{5}{*}{\textcolor{violet}{\textbf{Gpt-5.4-mini}}} & baseline & 14727 & 11.3 & {13.0} & 99.5 & 99.5 & 99.0 & 87.0 & 87.0 & \cellcolor{blue!15}{98.5} \\
 & cp & 15839 & 11.7 & {9.5} & 98.0 & 98.0 & 98.0 & 90.5 & 90.5 & \cellcolor{blue!15}{94.5} \\
 & cp\_cot & 16866 & 13.3 & {3.5} & 99.5 & 99.5 & 99.5 & 96.5 & 96.5 & \cellcolor{blue!15}{98.5} \\
 & cp\_icl & 16348 & 10.8 & {6.5} & 99.5 & 99.5 & 99.5 & 93.5 & 93.5 & \cellcolor{blue!15}{95.5} \\
 & cp\_icl\_cot & 17446 & 11.5 & {2.0} & 99.5 & 99.5 & 99.0 & 98.0 & 98.0 & \cellcolor{blue!15}{97.0} \\
\midrule
\rowcolor[HTML]{F0F0F0}
\multicolumn{11}{@{}l}{\emph{Local Ollama}} \\
\multirow{5}{*}{\shortstack[l]{\textcolor{darkgray}{\textbf{Qwen3.6:35b}} \\ \textcolor{darkgray}{\textbf{(Think)}}}} & baseline & 22700 & 55.9 & {11.5} & 98.0 & 98.0 & 97.5 & 88.5 & 88.5 & \cellcolor{blue!15}{97.0} \\
 & cp & 25990 & 68.4 & {5.0} & 100.0 & 100.0 & 100.0 & 95.0 & 95.0 & \cellcolor{blue!15}{100.0} \\
 & cp\_cot & 26216 & 66.3 & {4.5} & 99.5 & 99.5 & 99.5 & 95.5 & 95.5 & \cellcolor{blue!15}{99.5} \\
 & cp\_icl & 27796 & 66.7 & {7.5} & 98.0 & 98.0 & 98.0 & 92.5 & 92.5 & \cellcolor{blue!15}{98.0} \\
 & cp\_icl\_cot & 28681 & 68.8 & {8.0} & 98.5 & 98.5 & 98.5 & 92.0 & 92.0 & \cellcolor{blue!15}{98.5} \\
\cmidrule(l){1-11}
\multirow{5}{*}{\textcolor{olive}{\textbf{Qwen3.6:35b}}} & baseline & 19169 & 24.0 & {10.0} & 99.5 & 99.5 & 99.5 & 90.0 & 90.0 & \cellcolor{blue!15}{98.0} \\
 & cp & 20492 & 26.2 & {4.5} & 100.0 & 100.0 & 100.0 & 95.5 & 95.5 & \cellcolor{blue!15}{99.5} \\
 & cp\_cot & 21664 & 30.4 & {7.0} & 99.5 & 99.5 & 99.5 & 93.0 & 93.0 & \cellcolor{blue!15}{98.5} \\
 & cp\_icl & 23758 & 32.5 & {7.5} & 100.0 & 100.0 & 100.0 & 92.5 & 92.5 & \cellcolor{blue!15}{99.0} \\
 & cp\_icl\_cot & 23676 & 27.8 & {3.5} & 100.0 & 100.0 & 99.5 & 96.5 & 96.5 & \cellcolor{blue!15}{100.0} \\
\cmidrule(l){1-11}
\multirow{5}{*}{\shortstack[l]{\textcolor{brown}{\textbf{Gemma4:31b}} \\ \textcolor{brown}{\textbf{(Think)}}}} & baseline & 19840 & 88.7 & {6.5} & 98.5 & 98.5 & 98.5 & 93.5 & 93.5 & \cellcolor{blue!15}{98.0} \\
 & cp & 21374 & 97.3 & {4.5} & 99.5 & 99.5 & 99.5 & 95.5 & 95.5 & \cellcolor{blue!15}{99.5} \\
 & cp\_cot & 22479 & 104.1 & {5.5} & 99.0 & 99.0 & 99.0 & 94.5 & 94.5 & \cellcolor{blue!15}{98.5} \\
 & cp\_icl & 23446 & 104.8 & {7.5} & 96.1 & 96.1 & 96.1 & 89.6 & 89.6 & \cellcolor{blue!15}{96.1} \\
 & cp\_icl\_cot & {22346} & {118.9} & {6.0} & {98.0} & {98.0} & {98.0} & {94.0} & {94.0} & \cellcolor{blue!15}{{97.0}} \\
\cmidrule(l){1-11}
\multirow{5}{*}{\textcolor{magenta}{\textbf{Gemma4:31b}}} & baseline & 17864 & 41.2 & {8.0} & 100.0 & 100.0 & 100.0 & 92.0 & 92.0 & \cellcolor{blue!15}{99.5} \\
 & cp & 18868 & 48.6 & {7.5} & 99.5 & 99.5 & 99.5 & 92.5 & 92.5 & \cellcolor{blue!15}{99.5} \\
 & cp\_cot & 20225 & 60.1 & {5.0} & 99.0 & 99.0 & 99.0 & 95.0 & 95.0 & \cellcolor{blue!15}{98.5} \\
 & cp\_icl & 21264 & 50.0 & {7.5} & 99.0 & 99.0 & 99.0 & 92.5 & 92.5 & \cellcolor{blue!15}{99.0} \\
 & cp\_icl\_cot & 22420 & 57.7 & {5.5} & 99.0 & 99.0 & 99.0 & 94.5 & 94.5 & \cellcolor{blue!15}{99.0} \\
\cmidrule(l){1-11}
\multirow{5}{*}{\shortstack[l]{\textcolor{cyan}{\textbf{Gpt-oss:20b}} \\ \textcolor{cyan}{\textbf{(Think)}}}} & baseline & 18348 & 36.6 & {31.5} & 89.0 & 89.0 & 88.0 & 68.5 & 68.5 & \cellcolor{blue!15}{88.0} \\
 & cp & 18362 & 30.2 & {14.0} & 90.0 & 90.0 & 90.0 & 86.0 & 86.0 & \cellcolor{blue!15}{89.0} \\
 & cp\_cot & 19784 & 35.3 & {14.5} & 85.5 & 85.5 & 85.0 & 85.5 & 85.5 & \cellcolor{blue!15}{82.0} \\
 & cp\_icl & 23048 & 46.3 & {31.5} & 71.5 & 71.5 & 71.5 & 68.5 & 68.5 & \cellcolor{blue!15}{70.0} \\
 & cp\_icl\_cot & 31955 & 96.6 & {24.0} & 80.0 & 80.0 & 79.0 & 76.0 & 76.0 & \cellcolor{blue!15}{73.0} \\
\bottomrule
\end{tabular}
\vspace{1ex}
{\flushleft\footnotesize\textit{Note:} Arrows mark preferred directions ($\uparrow$/$\downarrow$). Tokens and latency (s) are means per query (errored / rate-limited zero-token rows excluded).\par}
\end{table*}

\begin{table*}[htbp]
\centering
\scriptsize
\setlength{\tabcolsep}{3.2pt}
\renewcommand{\arraystretch}{1.02}
\caption{Task B (behaviour preset) per-cell funnel KPIs ($N{=}200$ per cell). The rightmost column \textit{preset \%} is the headline check (the modified \texttt{vType} feature vector matches the expected preset within $rtol{=}10^{-3}$); \textit{vClass \%} confirms the LLM did not silently switch the vehicle class.}
\label{tab:mod-B-all-results}
\begin{tabular}{@{}llccccccccc@{}}
\toprule
\multirow{2}{*}{\textbf{Model}} & \multirow{2}{*}{\textbf{Prompt}} & \multicolumn{2}{c}{\textbf{Cost}} & \multirow{2}{*}{{\textbf{Fail\% $\downarrow$}}} & \multicolumn{6}{c}{\textbf{Pipeline stage pass rates}} \\
\cmidrule(lr){3-4} \cmidrule(lr){6-11}
 & & tokens $\downarrow$ & latency (s) $\downarrow$ & & XML \% $\uparrow$ & tgt \% $\uparrow$ & vClass \% $\uparrow$ & SUMO \% $\uparrow$ & CR \% $\uparrow$ & \cellcolor{blue!15}\textbf{preset \% $\uparrow$} \\
\midrule
\rowcolor[HTML]{F0F0F0}
\multicolumn{11}{@{}l}{\emph{Cloud API}} \\
\multirow{5}{*}{\textcolor{blue}{\textbf{Qwen3.6-plus}}} & baseline & 19149 & 53.8 & {8.5} & 100.0 & 100.0 & 92.5 & 91.5 & 91.5 & \cellcolor{blue!15}{0.0} \\
 & cp & 20667 & 56.3 & {0.0} & 100.0 & 100.0 & 100.0 & 100.0 & 100.0 & \cellcolor{blue!15}{100.0} \\
 & cp\_cot & 20784 & 61.5 & {0.0} & 100.0 & 100.0 & 100.0 & 100.0 & 100.0 & \cellcolor{blue!15}{100.0} \\
 & cp\_icl & 21882 & 46.8 & {0.0} & 100.0 & 100.0 & 100.0 & 100.0 & 100.0 & \cellcolor{blue!15}{100.0} \\
 & cp\_icl\_cot & 24666 & 55.5 & {0.5} & 100.0 & 100.0 & 100.0 & 99.5 & 99.5 & \cellcolor{blue!15}{100.0} \\
\cmidrule(l){1-11}
\multirow{5}{*}{\textbf{Deepseek-v3.2}} & baseline & 14152 & 73.8 & {9.5} & 98.0 & 98.0 & 95.5 & 90.5 & 90.5 & \cellcolor{blue!15}{0.0} \\
 & cp & 15801 & 73.2 & {1.0} & 99.5 & 99.5 & 99.5 & 99.0 & 99.0 & \cellcolor{blue!15}{99.5} \\
 & cp\_cot & 17074 & 84.9 & {1.0} & 99.0 & 99.0 & 99.0 & 99.0 & 99.0 & \cellcolor{blue!15}{99.0} \\
 & cp\_icl & 17027 & 59.2 & {0.5} & 99.5 & 99.5 & 99.5 & 99.5 & 99.5 & \cellcolor{blue!15}{99.5} \\
 & cp\_icl\_cot & 19543 & 74.0 & {2.5} & 97.5 & 97.5 & 97.5 & 97.5 & 97.5 & \cellcolor{blue!15}{97.5} \\
\cmidrule(l){1-11}
\multirow{5}{*}{\textcolor{orange}{\textbf{Glm-5}}} & baseline & 14233 & 40.2 & {9.5} & 99.0 & 99.0 & 88.5 & 90.5 & 90.5 & \cellcolor{blue!15}{0.0} \\
 & cp & 15531 & 43.9 & {1.0} & 99.0 & 99.0 & 99.0 & 99.0 & 99.0 & \cellcolor{blue!15}{96.5} \\
 & cp\_cot & 16723 & 51.9 & {1.0} & 99.0 & 99.0 & 99.0 & 99.0 & 99.0 & \cellcolor{blue!15}{99.0} \\
 & cp\_icl & 18382 & 45.4 & {3.0} & 99.0 & 99.0 & 97.0 & 97.0 & 97.0 & \cellcolor{blue!15}{99.0} \\
 & cp\_icl\_cot & 19756 & 53.1 & {2.0} & 99.0 & 99.0 & 98.0 & 98.0 & 98.0 & \cellcolor{blue!15}{99.0} \\
\cmidrule(l){1-11}
\multirow{5}{*}{\textcolor{teal}{\textbf{Gemini-3-flash}}} & baseline & 18266 & 11.4 & {17.0} & 99.0 & 99.0 & 87.0 & 83.0 & 83.0 & \cellcolor{blue!15}{0.5} \\
 & cp & 19732 & 11.5 & {1.0} & 99.0 & 99.0 & 99.0 & 99.0 & 99.0 & \cellcolor{blue!15}{99.0} \\
 & cp\_cot & 21028 & 13.1 & {1.0} & 99.0 & 99.0 & 99.0 & 99.0 & 99.0 & \cellcolor{blue!15}{98.5} \\
 & cp\_icl & 22326 & 10.6 & {1.0} & 99.0 & 99.0 & 99.0 & 99.0 & 99.0 & \cellcolor{blue!15}{99.0} \\
 & cp\_icl\_cot & 23915 & 12.7 & {2.0} & 99.0 & 99.0 & 99.0 & 98.0 & 98.0 & \cellcolor{blue!15}{99.0} \\
\cmidrule(l){1-11}
\multirow{5}{*}{\textcolor{violet}{\textbf{Gpt-5.4-mini}}} & baseline & 14756 & 13.3 & {12.5} & 99.0 & 99.0 & 88.5 & 87.5 & 87.5 & \cellcolor{blue!15}{0.0} \\
 & cp & 16121 & 12.4 & {2.0} & 98.5 & 98.5 & 98.0 & 98.0 & 98.0 & \cellcolor{blue!15}{98.0} \\
 & cp\_cot & 17223 & 14.2 & {1.0} & 99.0 & 99.0 & 99.0 & 99.0 & 99.0 & \cellcolor{blue!15}{99.0} \\
 & cp\_icl & 18447 & 12.0 & {1.5} & 99.0 & 99.0 & 99.0 & 98.5 & 98.5 & \cellcolor{blue!15}{99.0} \\
 & cp\_icl\_cot & 19521 & 13.3 & {1.0} & 99.0 & 99.0 & 99.0 & 99.0 & 99.0 & \cellcolor{blue!15}{99.0} \\
\midrule
\rowcolor[HTML]{F0F0F0}
\multicolumn{11}{@{}l}{\emph{Local Ollama}} \\
\multirow{5}{*}{\shortstack[l]{\textcolor{darkgray}{\textbf{Qwen3.6:35b}} \\ \textcolor{darkgray}{\textbf{(Think)}}}} & baseline & 23582 & 64.9 & {14.0} & 98.0 & 98.0 & 85.0 & 86.0 & 86.0 & \cellcolor{blue!15}{0.0} \\
 & cp & 28174 & 84.4 & {0.5} & 100.0 & 100.0 & 100.0 & 99.5 & 99.5 & \cellcolor{blue!15}{99.5} \\
 & cp\_cot & 28190 & 79.4 & {0.0} & 100.0 & 100.0 & 100.0 & 100.0 & 100.0 & \cellcolor{blue!15}{99.5} \\
 & cp\_icl & 29818 & 80.0 & {0.0} & 100.0 & 100.0 & 100.0 & 100.0 & 100.0 & \cellcolor{blue!15}{100.0} \\
 & cp\_icl\_cot & 31863 & 89.0 & {0.5} & 99.5 & 99.5 & 99.5 & 99.5 & 99.5 & \cellcolor{blue!15}{99.5} \\
\cmidrule(l){1-11}
\multirow{5}{*}{\textcolor{olive}{\textbf{Qwen3.6:35b}}} & baseline & 18980 & 28.3 & {22.5} & 98.0 & 98.0 & 84.5 & 77.5 & 77.5 & \cellcolor{blue!15}{0.5} \\
 & cp & 20650 & 28.7 & {3.0} & 97.5 & 97.5 & 97.0 & 97.0 & 97.0 & \cellcolor{blue!15}{97.5} \\
 & cp\_cot & 21437 & 30.9 & {2.0} & 98.0 & 98.0 & 98.0 & 98.0 & 98.0 & \cellcolor{blue!15}{97.5} \\
 & cp\_icl & 23511 & 33.1 & {4.0} & 98.5 & 98.5 & 96.0 & 96.0 & 96.0 & \cellcolor{blue!15}{98.5} \\
 & cp\_icl\_cot & 23546 & 28.3 & {1.0} & 99.0 & 99.0 & 99.0 & 99.0 & 99.0 & \cellcolor{blue!15}{99.0} \\
\cmidrule(l){1-11}
\multirow{5}{*}{\shortstack[l]{\textcolor{brown}{\textbf{Gemma4:31b}} \\ \textcolor{brown}{\textbf{(Think)}}}} & baseline & 19462 & 94.2 & {3.5} & 99.0 & 99.0 & 94.0 & 96.5 & 96.5 & \cellcolor{blue!15}{1.0} \\
 & cp & 21564 & 100.7 & {0.0} & 100.0 & 100.0 & 100.0 & 100.0 & 100.0 & \cellcolor{blue!15}{99.5} \\
 & cp\_cot & 22031 & 104.2 & {0.0} & 100.0 & 100.0 & 100.0 & 100.0 & 100.0 & \cellcolor{blue!15}{100.0} \\
 & cp\_icl & 23799 & 92.8 & {1.5} & 98.5 & 98.5 & 98.5 & 98.5 & 98.5 & \cellcolor{blue!15}{98.5} \\
 & cp\_icl\_cot & 24478 & 106.0 & {1.0} & 99.0 & 99.0 & 99.0 & 99.0 & 99.0 & \cellcolor{blue!15}{99.0} \\
\cmidrule(l){1-11}
\multirow{5}{*}{\textcolor{magenta}{\textbf{Gemma4:31b}}} & baseline & 17312 & 44.8 & {9.0} & 99.0 & 99.0 & 90.5 & 91.0 & 91.0 & \cellcolor{blue!15}{0.5} \\
 & cp & 18902 & 49.5 & {1.5} & 98.5 & 98.5 & 98.5 & 98.5 & 98.5 & \cellcolor{blue!15}{98.5} \\
 & cp\_cot & 20397 & 63.7 & {1.0} & 99.0 & 99.0 & 99.0 & 99.0 & 99.0 & \cellcolor{blue!15}{99.0} \\
 & cp\_icl & 21448 & 53.1 & {1.0} & 99.0 & 99.0 & 99.0 & 99.0 & 99.0 & \cellcolor{blue!15}{99.0} \\
 & cp\_icl\_cot & 22730 & 61.9 & {1.0} & 99.0 & 99.0 & 99.0 & 99.0 & 99.0 & \cellcolor{blue!15}{99.0} \\
\cmidrule(l){1-11}
\multirow{5}{*}{\shortstack[l]{\textcolor{cyan}{\textbf{Gpt-oss:20b}} \\ \textcolor{cyan}{\textbf{(Think)}}}} & baseline & 19056 & 39.1 & {12.5} & 93.5 & 93.5 & 87.0 & 87.5 & 87.5 & \cellcolor{blue!15}{0.0} \\
 & cp & 19023 & 32.1 & {8.5} & 91.5 & 91.5 & 91.0 & 91.5 & 91.5 & \cellcolor{blue!15}{87.5} \\
 & cp\_cot & 22174 & 46.8 & {7.0} & 93.0 & 93.0 & 93.0 & 93.0 & 93.0 & \cellcolor{blue!15}{91.0} \\
 & cp\_icl & 28777 & 76.7 & {18.5} & 82.0 & 82.0 & 82.0 & 81.5 & 81.5 & \cellcolor{blue!15}{81.5} \\
 & cp\_icl\_cot & 30613 & 79.2 & {15.0} & 85.0 & 85.0 & 85.0 & 85.0 & 85.0 & \cellcolor{blue!15}{83.0} \\
\bottomrule
\end{tabular}
\vspace{1ex}
{\flushleft\footnotesize\textit{Note:} Arrows mark preferred directions ($\uparrow$/$\downarrow$). Tokens and latency (s) are means per query (errored / rate-limited zero-token rows excluded).\par}
\end{table*}

\begin{table*}[htbp]
\centering
\scriptsize
\setlength{\tabcolsep}{3.2pt}
\renewcommand{\arraystretch}{1.02}
\caption{Task P (population edit, combined add+remove) per-cell funnel KPIs (first-100 of P\textsubscript{add} + first-100 of P\textsubscript{remove} = $N{=}200$ per cell). The rightmost column \textit{count\_match \%} is the headline check (the requested $+1/-1$ vehicle-count delta is present); \textit{tgt-chg \%} fires when the new vehicle starts on the requested edge (add) or the named vehicle is gone (remove); \textit{others \%} confirms non-target route entries are byte-equivalent.}
\label{tab:mod-P-all-results}
\begin{tabular}{@{}llccccccccc@{}}
\toprule
\multirow{2}{*}{\textbf{Model}} & \multirow{2}{*}{\textbf{Prompt}} & \multicolumn{2}{c}{\textbf{Cost}} & \multirow{2}{*}{{\textbf{Fail\% $\downarrow$}}} & \multicolumn{6}{c}{\textbf{Pipeline stage pass rates}} \\
\cmidrule(lr){3-4} \cmidrule(lr){6-11}
 & & tokens $\downarrow$ & latency (s) $\downarrow$ & & XML \% $\uparrow$ & tgt-chg \% $\uparrow$ & others \% $\uparrow$ & SUMO \% $\uparrow$ & CR \% $\uparrow$ & \cellcolor{blue!15}\textbf{count\_match \% $\uparrow$} \\
\midrule
\rowcolor[HTML]{F0F0F0}
\multicolumn{11}{@{}l}{\emph{Cloud API}} \\
\multirow{5}{*}{\textcolor{blue}{\textbf{Qwen3.6-plus}}} & baseline & 17948 & 50.9 & {33.0} & 99.0 & 90.5 & 98.5 & 67.0 & 67.0 & \cellcolor{blue!15}{99.0} \\
 & cp & 19066 & 53.2 & {4.0} & 99.0 & 99.0 & 99.0 & 96.0 & 96.0 & \cellcolor{blue!15}{99.0} \\
 & cp\_cot & 19958 & 58.5 & {2.0} & 99.0 & 99.0 & 99.0 & 98.0 & 98.0 & \cellcolor{blue!15}{99.0} \\
 & cp\_icl & 21371 & 47.3 & {3.5} & 99.0 & 98.5 & 98.5 & 96.5 & 96.5 & \cellcolor{blue!15}{98.5} \\
 & cp\_icl\_cot & 23369 & 55.6 & {0.0} & 100.0 & 100.0 & 99.5 & 100.0 & 100.0 & \cellcolor{blue!15}{100.0} \\
\cmidrule(l){1-11}
\multirow{5}{*}{\textbf{Deepseek-v3.2}} & baseline & 14410 & 70.0 & {34.5} & 99.5 & 98.5 & 99.5 & 65.5 & 65.5 & \cellcolor{blue!15}{99.0} \\
 & cp & 14993 & 67.3 & {1.0} & 99.5 & 99.5 & 99.5 & 99.0 & 99.0 & \cellcolor{blue!15}{99.5} \\
 & cp\_cot & 17068 & 77.7 & {0.5} & 100.0 & 100.0 & 99.5 & 99.5 & 99.5 & \cellcolor{blue!15}{100.0} \\
 & cp\_icl & 16418 & 56.6 & {2.0} & 99.5 & 99.5 & 98.5 & 98.0 & 98.0 & \cellcolor{blue!15}{99.5} \\
 & cp\_icl\_cot & 17683 & 69.4 & {1.0} & 99.5 & 99.5 & 99.0 & 99.0 & 99.0 & \cellcolor{blue!15}{99.5} \\
\cmidrule(l){1-11}
\multirow{5}{*}{\textcolor{orange}{\textbf{Glm-5}}} & baseline & 14635 & 36.4 & {39.0} & 99.0 & 98.0 & 99.0 & 61.0 & 61.0 & \cellcolor{blue!15}{99.0} \\
 & cp & 15763 & 41.6 & {2.0} & 99.0 & 99.0 & 98.5 & 98.0 & 98.0 & \cellcolor{blue!15}{99.0} \\
 & cp\_cot & 16505 & 49.6 & {1.0} & 99.0 & 99.0 & 99.0 & 99.0 & 99.0 & \cellcolor{blue!15}{99.0} \\
 & cp\_icl & 17122 & 41.5 & {3.0} & 99.0 & 99.0 & 98.5 & 97.0 & 97.0 & \cellcolor{blue!15}{99.0} \\
 & cp\_icl\_cot & 18640 & 51.2 & {1.0} & 99.0 & 99.0 & 99.0 & 99.0 & 99.0 & \cellcolor{blue!15}{99.0} \\
\cmidrule(l){1-11}
\multirow{5}{*}{\textcolor{teal}{\textbf{Gemini-3-flash}}} & baseline & 18992 & 10.7 & {51.0} & 98.5 & 96.0 & 98.5 & 49.0 & 49.0 & \cellcolor{blue!15}{98.5} \\
 & cp & 19996 & 11.0 & {1.5} & 99.0 & 99.0 & 99.0 & 98.5 & 98.5 & \cellcolor{blue!15}{99.0} \\
 & cp\_cot & 21198 & 12.4 & {1.0} & 99.0 & 99.0 & 99.0 & 99.0 & 99.0 & \cellcolor{blue!15}{99.0} \\
 & cp\_icl & 21946 & 10.5 & {1.5} & 99.0 & 99.0 & 99.0 & 98.5 & 98.5 & \cellcolor{blue!15}{99.0} \\
 & cp\_icl\_cot & 23271 & 12.0 & {1.0} & 99.0 & 99.0 & 99.0 & 99.0 & 99.0 & \cellcolor{blue!15}{99.0} \\
\cmidrule(l){1-11}
\multirow{5}{*}{\textcolor{violet}{\textbf{Gpt-5.4-mini}}} & baseline & 15085 & 11.9 & {12.0} & 97.5 & 96.5 & 97.5 & 88.0 & 88.0 & \cellcolor{blue!15}{97.5} \\
 & cp & 16124 & 12.5 & {3.0} & 99.0 & 98.5 & 98.0 & 97.0 & 97.0 & \cellcolor{blue!15}{98.5} \\
 & cp\_cot & 17215 & 13.7 & {3.0} & 99.0 & 99.0 & 99.0 & 97.0 & 97.0 & \cellcolor{blue!15}{99.0} \\
 & cp\_icl & 17642 & 11.2 & {1.5} & 99.0 & 98.0 & 98.0 & 98.5 & 98.5 & \cellcolor{blue!15}{98.5} \\
 & cp\_icl\_cot & 18676 & 13.4 & {2.5} & 99.0 & 99.0 & 99.0 & 97.5 & 97.5 & \cellcolor{blue!15}{99.0} \\
\midrule
\rowcolor[HTML]{F0F0F0}
\multicolumn{11}{@{}l}{\emph{Local Ollama}} \\
\multirow{5}{*}{\shortstack[l]{\textcolor{darkgray}{\textbf{Qwen3.6:35b}} \\ \textcolor{darkgray}{\textbf{(Think)}}}} & baseline & 23502 & 56.2 & {12.5} & 100.0 & 95.5 & 98.0 & 87.5 & 87.5 & \cellcolor{blue!15}{98.0} \\
 & cp & 26284 & 69.4 & {2.0} & 99.5 & 99.0 & 97.0 & 98.0 & 98.0 & \cellcolor{blue!15}{97.5} \\
 & cp\_cot & 27253 & 71.5 & {3.0} & 98.5 & 98.5 & 95.0 & 97.0 & 97.0 & \cellcolor{blue!15}{95.0} \\
 & cp\_icl & 26972 & 62.8 & {3.0} & 99.5 & 99.5 & 97.0 & 97.0 & 97.0 & \cellcolor{blue!15}{96.5} \\
 & cp\_icl\_cot & 29308 & 74.2 & {0.5} & 100.0 & 100.0 & 96.5 & 99.5 & 99.5 & \cellcolor{blue!15}{96.5} \\
\cmidrule(l){1-11}
\multirow{5}{*}{\textcolor{olive}{\textbf{Qwen3.6:35b}}} & baseline & 19200 & 24.0 & {6.0} & 100.0 & 95.5 & 99.0 & 94.0 & 94.0 & \cellcolor{blue!15}{99.0} \\
 & cp & 20452 & 25.8 & {3.0} & 100.0 & 98.0 & 100.0 & 97.0 & 97.0 & \cellcolor{blue!15}{100.0} \\
 & cp\_cot & 21402 & 28.6 & {9.0} & 100.0 & 100.0 & 96.5 & 91.0 & 91.0 & \cellcolor{blue!15}{96.5} \\
 & cp\_icl & 23033 & 32.6 & {1.5} & 100.0 & 98.0 & 99.0 & 98.5 & 98.5 & \cellcolor{blue!15}{99.0} \\
 & cp\_icl\_cot & 23296 & 30.3 & {2.5} & 100.0 & 99.5 & 95.5 & 97.5 & 97.5 & \cellcolor{blue!15}{95.5} \\
\cmidrule(l){1-11}
\multirow{5}{*}{\shortstack[l]{\textcolor{brown}{\textbf{Gemma4:31b}} \\ \textcolor{brown}{\textbf{(Think)}}}} & baseline & 20408 & 95.3 & {6.0} & 99.5 & 98.5 & 99.5 & 94.0 & 94.0 & \cellcolor{blue!15}{99.5} \\
 & cp & 21321 & 89.7 & {0.5} & 100.0 & 100.0 & 100.0 & 99.5 & 99.5 & \cellcolor{blue!15}{100.0} \\
 & cp\_cot & 22424 & 98.1 & {0.5} & 100.0 & 100.0 & 100.0 & 99.5 & 99.5 & \cellcolor{blue!15}{100.0} \\
 & cp\_icl & 23181 & 90.8 & {1.5} & 98.7 & 98.7 & 98.7 & 98.7 & 98.7 & \cellcolor{blue!15}{98.7} \\
 & cp\_icl\_cot & 24394 & 97.7 & {1.0} & 99.3 & 99.3 & 99.3 & 98.7 & 98.7 & \cellcolor{blue!15}{99.3} \\
\cmidrule(l){1-11}
\multirow{5}{*}{\textcolor{magenta}{\textbf{Gemma4:31b}}} & baseline & 17710 & 45.7 & {38.0} & 99.0 & 96.0 & 99.0 & 62.0 & 62.0 & \cellcolor{blue!15}{99.0} \\
 & cp & 19090 & 52.2 & {2.0} & 99.0 & 99.0 & 99.0 & 98.0 & 98.0 & \cellcolor{blue!15}{99.0} \\
 & cp\_cot & 20452 & 58.9 & {2.0} & 99.0 & 99.0 & 99.0 & 98.0 & 98.0 & \cellcolor{blue!15}{99.0} \\
 & cp\_icl & 20726 & 46.0 & {1.5} & 99.0 & 99.0 & 99.0 & 98.5 & 98.5 & \cellcolor{blue!15}{99.0} \\
 & cp\_icl\_cot & 22078 & 54.5 & {1.5} & 99.0 & 99.0 & 99.0 & 98.5 & 98.5 & \cellcolor{blue!15}{99.0} \\
\cmidrule(l){1-11}
\multirow{5}{*}{\shortstack[l]{\textcolor{cyan}{\textbf{Gpt-oss:20b}} \\ \textcolor{cyan}{\textbf{(Think)}}}} & baseline & 21892 & 53.2 & {12.0} & 90.0 & 90.0 & 90.0 & 85.0 & 85.0 & \cellcolor{blue!15}{90.0} \\
 & cp & 22965 & 49.1 & {10.0} & 90.0 & 90.0 & 90.0 & 88.0 & 88.0 & \cellcolor{blue!15}{90.0} \\
 & cp\_cot & 25822 & 74.1 & {10.5} & 92.9 & 92.9 & 92.9 & 90.5 & 90.5 & \cellcolor{blue!15}{92.9} \\
 & cp\_icl & {28929} & {102.9} & {18.5} & {82.0} & {82.0} & {81.5} & {81.5} & {81.5} & \cellcolor{blue!15}{{81.5}} \\
 & cp\_icl\_cot & {30534} & {86.3} & {20.5} & {80.0} & {80.0} & {79.5} & {79.5} & {79.5} & \cellcolor{blue!15}{{80.0}} \\
\bottomrule
\end{tabular}
\vspace{1ex}
{\flushleft\footnotesize\textit{Note:} Arrows mark preferred directions ($\uparrow$/$\downarrow$). Tokens and latency (s) are means per query (errored / rate-limited zero-token rows excluded).\par}
\end{table*}

\begin{table*}[htbp]
\centering
\scriptsize
\setlength{\tabcolsep}{3.2pt}
\renewcommand{\arraystretch}{1.02}
\caption{Task G (goal extraction) per-cell funnel KPIs ($N{=}200$ per cell). The rightmost column \textit{edge-GT \%} is the headline semantic check (the extracted target lanelet matches the GT); \textit{pos-GT \%} is the secondary semantic check (the position keyword inside that lanelet matches the GT). G has no Frenetix or SUMO column because the edit modifies only the planning problem, not traffic.}
\label{tab:mod-G-all-results}
\begin{tabular}{@{}llccccccccc@{}}
\toprule
\multirow{2}{*}{\textbf{Model}} & \multirow{2}{*}{\textbf{Prompt}} & \multicolumn{2}{c}{\textbf{Cost}} & \multirow{2}{*}{{\textbf{Fail\% $\downarrow$}}} & \multicolumn{6}{c}{\textbf{Pipeline stage pass rates}} \\
\cmidrule(lr){3-4} \cmidrule(lr){6-11}
 & & tokens $\downarrow$ & latency (s) $\downarrow$ & & JSON \% $\uparrow$ & edge \% $\uparrow$ & pos-enum \% $\uparrow$ & pos-GT \% $\uparrow$ & lanelet \% $\uparrow$ & \cellcolor{blue!15}\textbf{edge-GT \% $\uparrow$} \\
\midrule
\rowcolor[HTML]{F0F0F0}
\multicolumn{11}{@{}l}{\emph{Cloud API}} \\
\multirow{5}{*}{\textcolor{blue}{\textbf{Qwen3.6-plus}}} & baseline & 6089 & 2.4 & {11.5} & 100.0 & 88.5 & 100.0 & 94.5 & 88.5 & \cellcolor{blue!15}{88.5} \\
 & cp & 6345 & 2.2 & {1.0} & 100.0 & 99.0 & 100.0 & 100.0 & 99.0 & \cellcolor{blue!15}{99.0} \\
 & cp\_cot & 6695 & 3.6 & {0.0} & 100.0 & 100.0 & 100.0 & 100.0 & 100.0 & \cellcolor{blue!15}{100.0} \\
 & cp\_icl & 6746 & 2.2 & {0.0} & 100.0 & 100.0 & 100.0 & 100.0 & 100.0 & \cellcolor{blue!15}{100.0} \\
 & cp\_icl\_cot & 7097 & 3.3 & {0.0} & 100.0 & 100.0 & 100.0 & 100.0 & 100.0 & \cellcolor{blue!15}{100.0} \\
\cmidrule(l){1-11}
\multirow{5}{*}{\textbf{Deepseek-v3.2}} & baseline & 4732 & 9.0 & {11.0} & 100.0 & 89.0 & 100.0 & 98.5 & 89.0 & \cellcolor{blue!15}{89.0} \\
 & cp & 4966 & 8.2 & {8.0} & 100.0 & 92.0 & 100.0 & 100.0 & 92.0 & \cellcolor{blue!15}{92.0} \\
 & cp\_cot & 5280 & 8.9 & {0.0} & 100.0 & 100.0 & 100.0 & 100.0 & 100.0 & \cellcolor{blue!15}{100.0} \\
 & cp\_icl & 5326 & 10.8 & {0.0} & 100.0 & 100.0 & 100.0 & 100.0 & 100.0 & \cellcolor{blue!15}{100.0} \\
 & cp\_icl\_cot & 5631 & 9.8 & {0.0} & 100.0 & 100.0 & 100.0 & 100.0 & 100.0 & \cellcolor{blue!15}{100.0} \\
\cmidrule(l){1-11}
\multirow{5}{*}{\textcolor{orange}{\textbf{Glm-5}}} & baseline & 4958 & 3.6 & {11.0} & 100.0 & 89.0 & 98.5 & 98.5 & 89.0 & \cellcolor{blue!15}{89.0} \\
 & cp & 5194 & 3.3 & {3.0} & 100.0 & 97.0 & 99.5 & 99.5 & 97.0 & \cellcolor{blue!15}{97.0} \\
 & cp\_cot & 5521 & 4.3 & {0.0} & 100.0 & 100.0 & 100.0 & 100.0 & 100.0 & \cellcolor{blue!15}{100.0} \\
 & cp\_icl & 5559 & 3.0 & {7.5} & 100.0 & 92.5 & 100.0 & 100.0 & 92.5 & \cellcolor{blue!15}{92.5} \\
 & cp\_icl\_cot & 5883 & 3.9 & {0.0} & 100.0 & 100.0 & 100.0 & 100.0 & 100.0 & \cellcolor{blue!15}{100.0} \\
\cmidrule(l){1-11}
\multirow{5}{*}{\textcolor{teal}{\textbf{Gemini-3-flash}}} & baseline & 6037 & 1.0 & {0.0} & 100.0 & 100.0 & 100.0 & 100.0 & 100.0 & \cellcolor{blue!15}{100.0} \\
 & cp & 6301 & 0.9 & {0.0} & 100.0 & 100.0 & 100.0 & 100.0 & 100.0 & \cellcolor{blue!15}{100.0} \\
 & cp\_cot & 6646 & 1.3 & {0.0} & 100.0 & 100.0 & 100.0 & 100.0 & 100.0 & \cellcolor{blue!15}{100.0} \\
 & cp\_icl & 6710 & 0.9 & {0.0} & 100.0 & 100.0 & 100.0 & 100.0 & 100.0 & \cellcolor{blue!15}{100.0} \\
 & cp\_icl\_cot & 7047 & 1.3 & {0.0} & 100.0 & 100.0 & 100.0 & 100.0 & 100.0 & \cellcolor{blue!15}{100.0} \\
\cmidrule(l){1-11}
\multirow{5}{*}{\textcolor{violet}{\textbf{Gpt-5.4-mini}}} & baseline & 4687 & 0.7 & {0.0} & 100.0 & 100.0 & 100.0 & 91.5 & 100.0 & \cellcolor{blue!15}{100.0} \\
 & cp & 4927 & 0.7 & {0.0} & 100.0 & 100.0 & 100.0 & 100.0 & 100.0 & \cellcolor{blue!15}{100.0} \\
 & cp\_cot & 5262 & 1.1 & {0.0} & 100.0 & 100.0 & 100.0 & 100.0 & 100.0 & \cellcolor{blue!15}{100.0} \\
 & cp\_icl & 5281 & 0.7 & {0.0} & 100.0 & 100.0 & 100.0 & 100.0 & 100.0 & \cellcolor{blue!15}{100.0} \\
 & cp\_icl\_cot & 5575 & 0.9 & {0.0} & 100.0 & 100.0 & 100.0 & 100.0 & 100.0 & \cellcolor{blue!15}{100.0} \\
\midrule
\rowcolor[HTML]{F0F0F0}
\multicolumn{11}{@{}l}{\emph{Local Ollama}} \\
\multirow{5}{*}{\shortstack[l]{\textcolor{darkgray}{\textbf{Qwen3.6:35b}} \\ \textcolor{darkgray}{\textbf{(Think)}}}} & baseline & 6745 & 5.9 & {0.5} & 99.5 & 99.5 & 99.5 & 99.5 & 99.5 & \cellcolor{blue!15}{99.5} \\
 & cp & 6955 & 5.6 & {1.0} & 99.0 & 99.0 & 99.0 & 99.0 & 99.0 & \cellcolor{blue!15}{99.0} \\
 & cp\_cot & 8074 & 11.5 & {5.0} & 95.0 & 95.0 & 95.0 & 95.0 & 95.0 & \cellcolor{blue!15}{95.0} \\
 & cp\_icl & 7161 & 4.5 & {0.5} & 99.5 & 99.5 & 99.5 & 99.5 & 99.5 & \cellcolor{blue!15}{99.5} \\
 & cp\_icl\_cot & 8012 & 8.3 & {9.0} & 91.0 & 91.0 & 91.0 & 91.0 & 91.0 & \cellcolor{blue!15}{91.0} \\
\cmidrule(l){1-11}
\multirow{5}{*}{\textcolor{olive}{\textbf{Qwen3.6:35b}}} & baseline & 6085 & 1.7 & {0.5} & 100.0 & 99.5 & 100.0 & 94.0 & 99.5 & \cellcolor{blue!15}{99.5} \\
 & cp & 6346 & 1.7 & {0.0} & 100.0 & 100.0 & 100.0 & 100.0 & 100.0 & \cellcolor{blue!15}{100.0} \\
 & cp\_cot & 6697 & 2.2 & {0.0} & 100.0 & 100.0 & 100.0 & 98.0 & 100.0 & \cellcolor{blue!15}{100.0} \\
 & cp\_icl & 6747 & 1.7 & {0.0} & 100.0 & 100.0 & 100.0 & 100.0 & 100.0 & \cellcolor{blue!15}{100.0} \\
 & cp\_icl\_cot & 7105 & 2.3 & {0.0} & 100.0 & 100.0 & 100.0 & 100.0 & 100.0 & \cellcolor{blue!15}{100.0} \\
\cmidrule(l){1-11}
\multirow{5}{*}{\shortstack[l]{\textcolor{brown}{\textbf{Gemma4:31b}} \\ \textcolor{brown}{\textbf{(Think)}}}} & baseline & 6325 & 7.4 & {1.5} & 100.0 & 98.5 & 100.0 & 100.0 & 98.5 & \cellcolor{blue!15}{98.5} \\
 & cp & 6483 & 5.1 & {0.0} & 100.0 & 100.0 & 100.0 & 100.0 & 100.0 & \cellcolor{blue!15}{100.0} \\
 & cp\_cot & 6734 & 4.8 & {0.0} & 100.0 & 100.0 & 100.0 & 100.0 & 100.0 & \cellcolor{blue!15}{100.0} \\
 & cp\_icl & 6840 & 4.6 & {0.0} & 100.0 & 100.0 & 100.0 & 100.0 & 100.0 & \cellcolor{blue!15}{100.0} \\
 & cp\_icl\_cot & 7140 & 5.6 & {0.0} & 100.0 & 100.0 & 100.0 & 100.0 & 100.0 & \cellcolor{blue!15}{100.0} \\
\cmidrule(l){1-11}
\multirow{5}{*}{\textcolor{magenta}{\textbf{Gemma4:31b}}} & baseline & 6055 & 2.5 & {2.0} & 100.0 & 98.0 & 100.0 & 100.0 & 98.0 & \cellcolor{blue!15}{98.0} \\
 & cp & 6320 & 2.5 & {0.0} & 100.0 & 100.0 & 100.0 & 100.0 & 100.0 & \cellcolor{blue!15}{100.0} \\
 & cp\_cot & 6657 & 3.7 & {0.0} & 100.0 & 100.0 & 100.0 & 100.0 & 100.0 & \cellcolor{blue!15}{100.0} \\
 & cp\_icl & 6730 & 2.6 & {0.0} & 100.0 & 100.0 & 100.0 & 100.0 & 100.0 & \cellcolor{blue!15}{100.0} \\
 & cp\_icl\_cot & 7067 & 3.7 & {0.0} & 100.0 & 100.0 & 100.0 & 100.0 & 100.0 & \cellcolor{blue!15}{100.0} \\
\cmidrule(l){1-11}
\multirow{5}{*}{\shortstack[l]{\textcolor{cyan}{\textbf{Gpt-oss:20b}} \\ \textcolor{cyan}{\textbf{(Think)}}}} & baseline & 4851 & 1.2 & {1.0} & 100.0 & 99.0 & 100.0 & 99.0 & 99.0 & \cellcolor{blue!15}{99.0} \\
 & cp & 5078 & 1.0 & {0.0} & 100.0 & 100.0 & 100.0 & 100.0 & 100.0 & \cellcolor{blue!15}{100.0} \\
 & cp\_cot & 5415 & 1.6 & {6.0} & 94.0 & 94.0 & 94.0 & 94.0 & 94.0 & \cellcolor{blue!15}{94.0} \\
 & cp\_icl & 5432 & 1.2 & {0.0} & 100.0 & 100.0 & 100.0 & 100.0 & 100.0 & \cellcolor{blue!15}{100.0} \\
 & cp\_icl\_cot & 5825 & 1.6 & {0.0} & 100.0 & 100.0 & 100.0 & 100.0 & 100.0 & \cellcolor{blue!15}{100.0} \\
\bottomrule
\end{tabular}
\vspace{1ex}
{\flushleft\footnotesize\textit{Note:} Arrows mark preferred directions ($\uparrow$/$\downarrow$). Tokens and latency (s) are means per query (errored / rate-limited zero-token rows excluded).\par}
\end{table*}

\paragraph{Application: safety-criticality.}\label{app:mod-safety}
As a downstream application of the B-task modification, we verify that it can systematically shift the \emph{safety criticality} of a scenario. On $N{=}100$ paired (baseline, modified) scenarios run through Frenetix under identical default cost weights, where the modification (\texttt{gemini-3-flash-preview}/\texttt{cp\_icl\_cot}) sets the three NPCs closest along ego's executed baseline trajectory to the \texttt{aggressiveLaner} preset, we measure per-scenario \texttt{min\_risk\_score} (integer 0--5; 0~=~collision risk, 5~=~safe) using the criticality toolkit \emph{From-Words-to-Collisions} \citep{gao2025words}, computed directly on each side's Frenetix-executed trajectory (no re-planning required). The score distribution shifts measurably toward more critical under modification{}: the 0-bucket (collision risk) doubles from 10 to 20 and $23/84$ paired scenarios move to a strictly more critical bucket vs.\ $18$ less critical (paired Wilcoxon $count\_match{=}-0.095$, $p{=}0.44$). The B-task modification thereby provides a natural-language interface for shifting downstream scenario safety profiles, a building block for safety-critical scenario generation.

\subsection{Module Router}
\label{app:router}

\subsubsection{Dispatch Pseudocode}
\label{app:router-dispatch}

Algorithm~\ref{alg:router} formalises the per-turn dispatch. The router observes $(u_t, h_t)$, computes the \textit{argmax} over the action set $\mathcal{A}$ defined in \S\ref{sec:problem}, and invokes the action-conditional operator. The five branches cover the five-action $\mathcal{A}$. The actual implementation expands $\textsc{modify}$ into the four modification sub-tasks (T/B/P/G), $\textsc{test}$ into single- and batch-execution variants, and $\textsc{qa}$ into general/parameter/batch Q\&A specialisations.

\begin{algorithm}[h]
\caption{Module Router Dispatch}
\label{alg:router}
\begin{algorithmic}[1]
\Require database $\mathcal{D}$, planner config $\theta$, action set $\mathcal{A}$
\State $s \gets f_{\mathrm{gen}}(u_0)\;\textbf{or}\;f_{\mathrm{sel}}(u_0, \mathcal{D})$ \Comment{initial population}
\State $h \gets \emptyset$;\quad $o \gets \bot$ \Comment{empty history, no outcome yet}
\While{dialogue is active}
    \State observe utterance $u_t$
    \State \textbf{Stage 1 (LLM):} $(\hat{a}_t,\,\texttt{args}) \gets$ JSON output with $\hat{a}_t = \arg\max_{a\in\mathcal{A}} f_{\mathrm{router}}(a \mid u_t, h)$
    \State \textbf{Stage 2 (Process Engine):} dispatch on $\hat{a}_t$ with operand $u_t$
    \If{$\hat{a}_t = \textsc{modify}$}
        \State $s \gets f_{\mathrm{mod}}(s, u_t)$;\;\;\; $r \gets s$
    \ElsIf{$\hat{a}_t = \textsc{tune}$}
        \State $\theta \gets f_{\mathrm{tune}}(\theta, u_t)$;\;\;\; $r \gets \theta$
    \ElsIf{$\hat{a}_t = \textsc{test}$}
        \State $o \gets f_{\mathrm{test}}(s, \theta)$;\;\;\; $r \gets o$ \Comment{$o = (\tau, c, m)$}
    \ElsIf{$\hat{a}_t = \textsc{analyse}$}
        \State $r \gets f_{\mathrm{eval}}(o, u_t)$ \Comment{requires prior \textsc{test}}
    \ElsIf{$\hat{a}_t = \textsc{qa}$}
        \State $r \gets$ LLM Q\&A on $(u_t, h)$
    \EndIf
    \State $h \gets h \cup \{(u_t, \hat{a}_t, r)\}$
\EndWhile
\end{algorithmic}
\end{algorithm}

\subsubsection{Query Corpus}
\label{app:router-corpus}

The Module Router benchmark uses 200 queries spanning 9 action classes (the eight supported actions plus the \texttt{fail} class that the router emits when a request is unsupported). Table~\ref{tab:router-corpus} summarises the action-class distribution and, for the \texttt{vehicle\_mod} class, the sub-tasks each query exercises. Composite \texttt{vehicle\_mod} codes (e.g.\ \texttt{T+B+P}) test that the router enumerates multiple letters and respects the canonical $T{\to}B{\to}P{\to}G$ ordering rule.

\begin{table}[ht]
\centering
\footnotesize
\renewcommand{\arraystretch}{1.1}
\caption{Composition of the Module Router corpus.}
\label{tab:router-corpus}
\begin{tabular}{@{}lr | lr@{}}
\toprule
\textbf{Action Class} & \textbf{\#} & \textbf{Mod Sub-tasks} & \textbf{\#} \\
\midrule
vehicle\_mod & 40 & Goal (G) & 9 \\
qa & 23 & Behavior (B) & 9 \\
param\_qa & 21 & Add/Rem (P) & 7 \\
batch\_qa & 20 & T+B+P & 3 \\
batch\_sim & 20 & Trajectory (T) & 3 \\
param\_mod & 20 & T+B & 2 \\
analysis & 19 & T+P & 2 \\
batch\_analysis & 19 & P+G & 2 \\
fail & 18 & B+P & 2 \\
\bottomrule
\end{tabular}
\end{table}

\subsubsection{Evaluation Metrics}
\label{app:router-metrics}

The Module Router is graded by a five-stage funnel computed against the \gls{GT} JSON action emitted by each prompt:
\begin{itemize}\itemsep0pt\parsep0pt
    \item \textbf{JSON \%}: the model output parses as a JSON object with exactly one top-level key.
    \item \textbf{key \%}: the top-level key matches the \gls{GT} class (e.g.\ the composite vehicle mod or the refusal class fail).
    \item \textbf{value \%} (\textbf{headline}): the value is a strict string match against \gls{GT}. For vehicle mod composites, this metric is sensitive to letter ordering.
    \item \textbf{set \%}: the letter set matches \gls{GT}, relaxing the canonical $T{\to}B{\to}P{\to}G$ ordering rule. Reported only for vehicle mod composites; isolates the ordering signal from the underlying classification accuracy.
    \item \textbf{order \%}: the emitted letter sequence honours the canonical order. Reported on the composite-only subset.
\end{itemize}
The composite \textit{Overall} score is the mean of the per-query boolean checks above. We report mean tokens and per-query latency as the cost cluster.

\subsubsection{Evaluation Results}
\label{app:router-results}

\begin{table*}[htbp]
\centering
\scriptsize
\setlength{\tabcolsep}{3.2pt}
\renewcommand{\arraystretch}{1.02}
\caption{Performance of the Module Router on the action-classification benchmark ($N{=}200$ per cell). \textit{value~\%} is a strict string match against the GT JSON action (ordering-sensitive for \texttt{vehicle\_mod} composites); \textit{set~\%} relaxes the canonical $T{\to}B{\to}P{\to}G$ ordering rule and is reported only on the \texttt{vehicle\_mod} composite subset.}
\label{tab:router-all-results}
\begin{tabular}{@{}llccccccccc@{}}
\toprule
\multirow{2}{*}{\textbf{Model}} & \multirow{2}{*}{\textbf{Prompt}} & \multicolumn{2}{c}{\textbf{Cost}} & \multirow{2}{*}{{\textbf{Fail\% $\downarrow$}}} & \multicolumn{5}{c}{\textbf{Pipeline stage pass rates}} & \multirow{2}{*}{\textbf{Overall $\uparrow$}} \\
\cmidrule(lr){3-4} \cmidrule(lr){6-10}
 & & tokens $\downarrow$ & latency (s) $\downarrow$ & & JSON \% $\uparrow$ & key \% $\uparrow$ & value \% $\uparrow$ & set \% $\uparrow$ & order \% $\uparrow$ & \\
\midrule
\rowcolor[HTML]{F0F0F0}
\multicolumn{11}{@{}l}{\emph{Cloud API}} \\
\multirow{5}{*}{\textcolor{blue}{\textbf{Qwen3.6-plus}}} & baseline & 274 & 1.7 & {0.0} & 100.0 & 49.5 & 47.0 & 0.0 & --- & \cellcolor{blue!15}{0.651} \\
 & cp & 1013 & 1.6 & {0.0} & 100.0 & 95.0 & 94.5 & 100.0 & 91.7 & \cellcolor{blue!15}{0.965} \\
 & cp\_cot & 1446 & 3.4 & {0.0} & 100.0 & 97.0 & 96.5 & 97.5 & 100.0 & \cellcolor{blue!15}{0.978} \\
 & cp\_icl & 2026 & 1.6 & {0.0} & 100.0 & 99.0 & 99.0 & 100.0 & 100.0 & \cellcolor{blue!15}{0.993} \\
 & cp\_icl\_cot & 2459 & 3.5 & {0.0} & 100.0 & 97.5 & 97.0 & 97.5 & 100.0 & \cellcolor{blue!15}{0.981} \\
\cmidrule(l){1-11}
\multirow{5}{*}{\textbf{Deepseek-v3.2}} & baseline & 261 & 7.1 & {0.0} & 100.0 & 62.5 & 48.5 & 22.2 & 100.0 & \cellcolor{blue!15}{0.681} \\
 & cp & 983 & 2.3 & {0.0} & 100.0 & 92.5 & 91.5 & 95.0 & 100.0 & \cellcolor{blue!15}{0.946} \\
 & cp\_cot & 1397 & 4.4 & {0.0} & 100.0 & 91.0 & 90.5 & 97.5 & 100.0 & \cellcolor{blue!15}{0.938} \\
 & cp\_icl & 1943 & 6.4 & {0.0} & 100.0 & 97.0 & 96.5 & 97.4 & 100.0 & \cellcolor{blue!15}{0.978} \\
 & cp\_icl\_cot & 2353 & 4.3 & {0.0} & 100.0 & 95.5 & 95.5 & 100.0 & 100.0 & \cellcolor{blue!15}{0.970} \\
\cmidrule(l){1-11}
\multirow{5}{*}{\textcolor{orange}{\textbf{Glm-5}}} & baseline & 261 & 3.1 & {0.5} & 99.5 & 64.5 & 51.5 & 17.2 & 100.0 & \cellcolor{blue!15}{0.699} \\
 & cp & 979 & 2.7 & {0.0} & 100.0 & 94.5 & 94.5 & 100.0 & 100.0 & \cellcolor{blue!15}{0.963} \\
 & cp\_cot & 1386 & 5.1 & {0.0} & 100.0 & 96.5 & 96.0 & 97.5 & 100.0 & \cellcolor{blue!15}{0.974} \\
 & cp\_icl & 1919 & 2.4 & {0.0} & 100.0 & 97.0 & 96.5 & 97.5 & 100.0 & \cellcolor{blue!15}{0.978} \\
 & cp\_icl\_cot & 2320 & 3.8 & {0.0} & 100.0 & 98.0 & 98.0 & 100.0 & 100.0 & \cellcolor{blue!15}{0.987} \\
\cmidrule(l){1-11}
\multirow{5}{*}{\textcolor{teal}{\textbf{Gemini-3-flash}}} & baseline & 256 & 0.9 & {0.0} & 100.0 & 73.0 & 62.0 & 29.0 & 100.0 & \cellcolor{blue!15}{0.767} \\
 & cp & 1019 & 0.9 & {0.0} & 100.0 & 97.5 & 97.5 & 100.0 & 100.0 & \cellcolor{blue!15}{0.983} \\
 & cp\_cot & 1429 & 1.3 & {0.0} & 100.0 & 97.0 & 96.5 & 97.5 & 100.0 & \cellcolor{blue!15}{0.978} \\
 & cp\_icl & 2064 & 1.0 & {0.0} & 100.0 & 97.5 & 97.0 & 97.5 & 100.0 & \cellcolor{blue!15}{0.981} \\
 & cp\_icl\_cot & 2479 & 1.3 & {0.0} & 100.0 & 99.0 & 99.0 & 100.0 & 100.0 & \cellcolor{blue!15}{0.993} \\
\cmidrule(l){1-11}
\multirow{5}{*}{\textcolor{violet}{\textbf{Gpt-5.4-mini}}} & baseline & 272 & 1.2 & {0.0} & 100.0 & 57.0 & 43.5 & 22.9 & 100.0 & \cellcolor{blue!15}{0.649} \\
 & cp & 982 & 1.0 & {0.0} & 100.0 & 91.0 & 91.0 & 100.0 & 100.0 & \cellcolor{blue!15}{0.940} \\
 & cp\_cot & 1309 & 1.1 & {0.0} & 100.0 & 90.5 & 90.5 & 100.0 & 100.0 & \cellcolor{blue!15}{0.937} \\
 & cp\_icl & 1913 & 1.0 & {0.0} & 100.0 & 98.0 & 98.0 & 100.0 & 100.0 & \cellcolor{blue!15}{0.987} \\
 & cp\_icl\_cot & 2243 & 1.1 & {0.0} & 100.0 & 98.5 & 98.5 & 100.0 & 100.0 & \cellcolor{blue!15}{0.990} \\
\midrule
\rowcolor[HTML]{F0F0F0}
\multicolumn{11}{@{}l}{\emph{Local Ollama}} \\
\multirow{5}{*}{\shortstack[l]{\textcolor{darkgray}{\textbf{Qwen3.6:35b}} \\ \textcolor{darkgray}{\textbf{(Think)}}}} & baseline & 5110 & 35.9 & {26.5} & 73.5 & 48.0 & 45.5 & 37.5 & 100.0 & \cellcolor{blue!15}{0.552} \\
 & cp & 1291 & 2.1 & {0.0} & 100.0 & 93.0 & 92.0 & 94.9 & 100.0 & \cellcolor{blue!15}{0.949} \\
 & cp\_cot & 2743 & 10.2 & {4.5} & 95.5 & 91.0 & 91.0 & 100.0 & 100.0 & \cellcolor{blue!15}{0.925} \\
 & cp\_icl & 2271 & 2.2 & {0.0} & 100.0 & 96.0 & 96.0 & 100.0 & 100.0 & \cellcolor{blue!15}{0.973} \\
 & cp\_icl\_cot & 2874 & 4.0 & {0.0} & 100.0 & 96.5 & 96.5 & 100.0 & 100.0 & \cellcolor{blue!15}{0.977} \\
\cmidrule(l){1-11}
\multirow{5}{*}{\textcolor{olive}{\textbf{Qwen3.6:35b}}} & baseline & 289 & 0.5 & {21.0} & 79.0 & 46.0 & 30.5 & 18.4 & 100.0 & \cellcolor{blue!15}{0.494} \\
 & cp & 1018 & 0.4 & {0.0} & 100.0 & 94.5 & 93.0 & 97.5 & 83.3 & \cellcolor{blue!15}{0.957} \\
 & cp\_cot & 1465 & 1.2 & {0.0} & 100.0 & 95.5 & 95.5 & 100.0 & 100.0 & \cellcolor{blue!15}{0.970} \\
 & cp\_icl & 2028 & 0.6 & {0.0} & 100.0 & 96.5 & 95.5 & 95.0 & 100.0 & \cellcolor{blue!15}{0.973} \\
 & cp\_icl\_cot & 2448 & 1.2 & {0.0} & 100.0 & 97.0 & 97.0 & 100.0 & 100.0 & \cellcolor{blue!15}{0.980} \\
\cmidrule(l){1-11}
\multirow{5}{*}{\shortstack[l]{\textcolor{brown}{\textbf{Gemma4:31b}} \\ \textcolor{brown}{\textbf{(Think)}}}} & baseline & 1035 & 12.6 & {0.0} & 100.0 & 69.5 & 65.5 & 33.3 & 75.0 & \cellcolor{blue!15}{0.778} \\
 & cp & 1319 & 4.9 & {0.0} & 100.0 & 96.5 & 95.0 & 92.5 & 100.0 & \cellcolor{blue!15}{0.971} \\
 & cp\_cot & 1704 & 5.6 & {0.0} & 100.0 & 97.0 & 96.0 & 95.0 & 100.0 & \cellcolor{blue!15}{0.976} \\
 & cp\_icl & 2347 & 4.8 & {0.0} & 100.0 & 97.0 & 97.0 & 100.0 & 100.0 & \cellcolor{blue!15}{0.980} \\
 & cp\_icl\_cot & 2721 & 5.5 & {0.0} & 100.0 & 98.0 & 98.0 & 100.0 & 100.0 & \cellcolor{blue!15}{0.987} \\
\cmidrule(l){1-11}
\multirow{5}{*}{\textcolor{magenta}{\textbf{Gemma4:31b}}} & baseline & 278 & 0.5 & {0.0} & 100.0 & 60.0 & 57.5 & 0.0 & --- & \cellcolor{blue!15}{0.722} \\
 & cp & 1038 & 0.4 & {0.0} & 100.0 & 97.0 & 96.5 & 97.5 & 100.0 & \cellcolor{blue!15}{0.978} \\
 & cp\_cot & 1440 & 1.4 & {0.0} & 100.0 & 98.0 & 98.0 & 100.0 & 100.0 & \cellcolor{blue!15}{0.987} \\
 & cp\_icl & 2080 & 0.4 & {0.0} & 100.0 & 98.5 & 97.0 & 100.0 & 75.0 & \cellcolor{blue!15}{0.984} \\
 & cp\_icl\_cot & 2488 & 1.5 & {0.0} & 100.0 & 99.5 & 99.5 & 100.0 & 100.0 & \cellcolor{blue!15}{0.997} \\
\cmidrule(l){1-11}
\multirow{5}{*}{\shortstack[l]{\textcolor{cyan}{\textbf{Gpt-oss:20b}} \\ \textcolor{cyan}{\textbf{(Think)}}}} & baseline & 696 & 2.1 & {0.5} & 99.5 & 53.0 & 42.5 & 32.3 & 62.5 & \cellcolor{blue!15}{0.633} \\
 & cp & 1135 & 0.6 & {18.5} & 81.5 & 78.0 & 78.0 & 100.0 & 100.0 & \cellcolor{blue!15}{0.792} \\
 & cp\_cot & 1656 & 1.6 & {0.0} & 100.0 & 95.5 & 95.5 & 100.0 & 100.0 & \cellcolor{blue!15}{0.970} \\
 & cp\_icl & 2110 & 0.7 & {6.0} & 94.0 & 91.0 & 91.0 & 100.0 & 100.0 & \cellcolor{blue!15}{0.920} \\
 & cp\_icl\_cot & 2581 & 1.3 & {0.5} & 99.5 & 96.0 & 96.0 & 100.0 & 100.0 & \cellcolor{blue!15}{0.972} \\
\bottomrule
\end{tabular}
\vspace{1ex}
{\flushleft\footnotesize\textit{Note:} Arrows mark preferred directions ($\uparrow$/$\downarrow$). Tokens and latency (s) are means per query.\par}
\end{table*}

Table~\ref{tab:router-all-results} reports the full results matrix ($N=200$ per cell, all 50 (model, condition) cells fully populated). The prompt ladder lifts strict-match accuracy substantially for every model: \texttt{gpt-5.4-mini} climbs from 43.5\% (baseline) to 98.5\% under \texttt{cp\_icl\_cot}. \texttt{Gemini-3-flash-preview} climbs from 62.0\% to 99.0\%. The \texttt{cp} condition delivers the biggest single jump (the categorical action schema and the ordering rule both fit cleanly into a curated-prompt block). Adding \gls{ICL} exemplars (\texttt{cp\_icl}) and \gls{CoT} scaffolding (\texttt{cp\_icl\_cot}) yield smaller but consistent gains. The relaxed \textit{set \%} metric is at or near 100\% for almost every (model, condition) cell once any prompting is applied, confirming that the residual error under the headline metric is dominated by letter-ordering mistakes rather than misclassification of the underlying sub-tasks.

\subsection{Planner Testing and Enhancement Module}
\label{app:planner}

\subsubsection{Query Corpus}
\label{app:planner-corpus}

The Planner Testing and Enhancement benchmark uses 200 queries that test the LLM's ability to emit a complete updated Frenetix \texttt{cost.yaml}. Table~\ref{tab:planner-corpus} summarises the request-type distribution and per-parameter coverage. Queries are partitioned into four request types (\textit{pure preset} = name a preset only; \textit{pure explicit} = name one or more parameters with target values; \textit{mixed} = preset + explicit overrides; \textit{out of vocab} = name a parameter that does not exist in the schema, which the LLM must refuse).

\begin{table}[ht]
\centering
\footnotesize
\renewcommand{\arraystretch}{1.1}
\caption{Composition of the Planner Testing and Enhancement configuration corpus.}
\label{tab:planner-corpus}
\begin{tabular}{@{}lr | lr@{}}
\toprule
\textbf{Request Type} & \textbf{\#} & \textbf{Top Parameters} & \textbf{\#} \\
\midrule
Pure preset & 71 & dist to ref path & 21 \\
Pure explicit & 69 & dist to obstacles & 18 \\
Mixed & 51 & orientation offset & 16 \\
Out of vocab & 9 & accel / jerk & 14 \\
& & respons. / path len & 13 \\
\bottomrule
\end{tabular}
\end{table}

\subsubsection{Evaluation Metrics}
\label{app:planner-metrics}

The Planner Testing and Enhancement task is graded by a six-stage funnel against the \gls{GT} YAML, plus a downstream simulator-runnability check:
\begin{itemize}\itemsep0pt\parsep0pt
    \item \textbf{YAML \%}: the model output parses as a YAML mapping.
    \item \textbf{struct \%}: the 13 cost weights keys and 3 external cost weights keys are present, with identical names and ordering to the input \texttt{cost.yaml}.
    \item \textbf{preset \%}: on queries whose \gls{GT} carries a named preset, every cost weight matches the preset's tabulated value within relative tolerance $10^{-3}$.
    \item \textbf{explicit \%}: on queries whose \gls{GT} carries explicit (key, value) pairs, each named parameter matches \gls{GT} within relative tolerance $10^{-3}$.
    \item \textbf{full-YAML \%} (\textbf{headline}): the strictest stage: every value in the emitted YAML matches \gls{GT} within relative tolerance $10^{-3}$ (combines the previous two stages plus untouched defaults).
    \item \textbf{refused \%}: on the 9-query out-of-vocab slice, the model declines to emit a YAML and instead surfaces a refusal token (e.g.\ ``do not'', ``not a valid parameter'').
\end{itemize}
The composite \textit{Overall} score is the mean of the per-query boolean checks. As an additional downstream check we run Frenetix on every emitted YAML against a paired baseline-simulable scenario; an outcome of either reaching the goal or hitting the simulation horizon without a crash counts as runnable.

\subsubsection{Evaluation Results}
\label{app:planner-results}

\begin{table*}[htbp]
\centering
\scriptsize
\setlength{\tabcolsep}{2.8pt}
\renewcommand{\arraystretch}{1.02}
\caption{Performance of the Planner Testing and Enhancement module ($N{=}200$ per cell). \textit{full-YAML~\%} is a strict element-wise match against the GT \texttt{cost.yaml} (rtol $10^{-3}$); \textit{preset~\%} and \textit{explicit~\%} are reported on the query subsets carrying a preset or explicit assignment respectively; \textit{refused~\%} is the 9-query out-of-vocab refusal slice.}
\label{tab:planner-all-results}
\begin{tabular}{@{}llcccccccccc@{}}
\toprule
\multirow{2}{*}{\textbf{Model}} & \multirow{2}{*}{\textbf{Prompt}} & \multicolumn{2}{c}{\textbf{Cost}} & \multirow{2}{*}{{\textbf{Fail\% $\downarrow$}}} & \multicolumn{6}{c}{\textbf{Pipeline stage pass rates}} & \multirow{2}{*}{\textbf{Overall $\uparrow$}} \\
\cmidrule(lr){3-4} \cmidrule(lr){6-11}
 & & tokens $\downarrow$ & latency (s) $\downarrow$ & & YAML \% $\uparrow$ & struct \% $\uparrow$ & preset \% $\uparrow$ & explicit \% $\uparrow$ & full-YAML \% $\uparrow$ & refused \% $\uparrow$ & \\
\midrule
\rowcolor[HTML]{F0F0F0}
\multicolumn{12}{@{}l}{\emph{Cloud API}} \\
\multirow{5}{*}{\textcolor{blue}{\textbf{Qwen3.6-plus}}} & baseline & 532 & 4.3 & {0.0} & 100.0 & 100.0 & 0.0 & 98.3 & 35.6 & 0.0 & \cellcolor{blue!15}{0.672} \\
 & cp & 1479 & 4.4 & {0.0} & 100.0 & 100.0 & 98.4 & 99.2 & 99.0 & 88.9 & \cellcolor{blue!15}{0.990} \\
 & cp\_cot & 1966 & 7.1 & {0.0} & 100.0 & 100.0 & 100.0 & 100.0 & 100.0 & 88.9 & \cellcolor{blue!15}{0.995} \\
 & cp\_icl & 2964 & 4.3 & {0.0} & 100.0 & 100.0 & 99.2 & 100.0 & 99.5 & 100.0 & \cellcolor{blue!15}{0.998} \\
 & cp\_icl\_cot & 3444 & 7.1 & {0.0} & 100.0 & 100.0 & 100.0 & 100.0 & 100.0 & 100.0 & \cellcolor{blue!15}{1.000} \\
\cmidrule(l){1-12}
\multirow{5}{*}{\textbf{Deepseek-v3.2}} & baseline & 532 & 9.8 & {0.0} & 100.0 & 100.0 & 0.0 & 99.2 & 36.1 & 0.0 & \cellcolor{blue!15}{0.674} \\
 & cp & 1498 & 8.3 & {0.5} & 99.5 & 99.5 & 100.0 & 99.2 & 99.0 & 88.9 & \cellcolor{blue!15}{0.989} \\
 & cp\_cot & 1965 & 11.2 & {0.5} & 99.5 & 99.5 & 99.2 & 100.0 & 99.0 & 100.0 & \cellcolor{blue!15}{0.994} \\
 & cp\_icl & 3003 & 6.9 & {0.5} & 99.5 & 99.5 & 99.2 & 99.2 & 99.0 & 100.0 & \cellcolor{blue!15}{0.993} \\
 & cp\_icl\_cot & 3465 & 10.1 & {0.0} & 100.0 & 100.0 & 97.5 & 100.0 & 98.4 & 100.0 & \cellcolor{blue!15}{0.993} \\
\cmidrule(l){1-12}
\multirow{5}{*}{\textcolor{orange}{\textbf{Glm-5}}} & baseline & 511 & 5.6 & {0.0} & 100.0 & 100.0 & 0.0 & 97.5 & 35.6 & 0.0 & \cellcolor{blue!15}{0.671} \\
 & cp & 1437 & 5.0 & {0.5} & 99.5 & 99.5 & 98.4 & 97.5 & 98.4 & 100.0 & \cellcolor{blue!15}{0.989} \\
 & cp\_cot & 1910 & 8.0 & {0.5} & 99.5 & 99.5 & 99.2 & 99.2 & 99.0 & 100.0 & \cellcolor{blue!15}{0.993} \\
 & cp\_icl & 2883 & 4.9 & {1.0} & 99.0 & 99.0 & 97.5 & 99.2 & 98.4 & 100.0 & \cellcolor{blue!15}{0.987} \\
 & cp\_icl\_cot & 3353 & 7.7 & {1.0} & 99.0 & 99.0 & 96.7 & 97.5 & 96.3 & 100.0 & \cellcolor{blue!15}{0.978} \\
\cmidrule(l){1-12}
\multirow{5}{*}{\textcolor{teal}{\textbf{Gemini-3-flash}}} & baseline & 561 & 1.5 & {0.0} & 100.0 & 100.0 & 0.0 & 99.2 & 36.1 & 0.0 & \cellcolor{blue!15}{0.674} \\
 & cp & 1581 & 2.3 & {0.0} & 100.0 & 100.0 & 100.0 & 99.2 & 99.0 & 100.0 & \cellcolor{blue!15}{0.996} \\
 & cp\_cot & 2099 & 6.2 & {0.0} & 100.0 & 100.0 & 100.0 & 100.0 & 99.0 & 100.0 & \cellcolor{blue!15}{0.998} \\
 & cp\_icl & 3202 & 1.6 & {0.0} & 100.0 & 100.0 & 100.0 & 100.0 & 99.0 & 100.0 & \cellcolor{blue!15}{0.998} \\
 & cp\_icl\_cot & 3697 & 2.3 & {0.0} & 100.0 & 100.0 & 98.4 & 99.2 & 97.9 & 100.0 & \cellcolor{blue!15}{0.992} \\
\cmidrule(l){1-12}
\multirow{5}{*}{\textcolor{violet}{\textbf{Gpt-5.4-mini}}} & baseline & 521 & 1.7 & {0.0} & 100.0 & 100.0 & 0.0 & 100.0 & 36.1 & 0.0 & \cellcolor{blue!15}{0.675} \\
 & cp & 1456 & 1.7 & {0.0} & 100.0 & 100.0 & 100.0 & 100.0 & 100.0 & 44.4 & \cellcolor{blue!15}{0.975} \\
 & cp\_cot & 1884 & 2.4 & {0.0} & 100.0 & 100.0 & 100.0 & 100.0 & 100.0 & 77.8 & \cellcolor{blue!15}{0.990} \\
 & cp\_icl & 2924 & 1.6 & {0.0} & 100.0 & 100.0 & 100.0 & 100.0 & 100.0 & 100.0 & \cellcolor{blue!15}{1.000} \\
 & cp\_icl\_cot & 3246 & 1.9 & {0.0} & 100.0 & 100.0 & 100.0 & 100.0 & 100.0 & 100.0 & \cellcolor{blue!15}{1.000} \\
\midrule
\rowcolor[HTML]{F0F0F0}
\multicolumn{12}{@{}l}{\emph{Local Ollama}} \\
\multirow{5}{*}{\shortstack[l]{\textcolor{darkgray}{\textbf{Qwen3.6:35b}} \\ \textcolor{darkgray}{\textbf{(Think)}}}} & baseline & 2459 & 13.9 & {0.0} & 100.0 & 99.0 & 0.0 & 100.0 & 36.1 & 0.0 & \cellcolor{blue!15}{0.673} \\
 & cp & 3252 & 13.1 & {0.5} & 99.5 & 99.0 & 95.9 & 99.2 & 96.9 & 88.9 & \cellcolor{blue!15}{0.979} \\
 & cp\_cot & 4140 & 16.9 & {0.0} & 100.0 & 100.0 & 95.9 & 100.0 & 97.4 & 100.0 & \cellcolor{blue!15}{0.990} \\
 & cp\_icl & 4129 & 9.0 & {0.0} & 100.0 & 100.0 & 96.7 & 100.0 & 97.9 & 100.0 & \cellcolor{blue!15}{0.991} \\
 & cp\_icl\_cot & 5206 & 14.1 & {0.0} & 100.0 & 100.0 & 100.0 & 100.0 & 100.0 & 100.0 & \cellcolor{blue!15}{1.000} \\
\cmidrule(l){1-12}
\multirow{5}{*}{\textcolor{olive}{\textbf{Qwen3.6:35b}}} & baseline & 533 & 1.3 & {0.0} & 100.0 & 100.0 & 0.0 & 100.0 & 36.1 & 0.0 & \cellcolor{blue!15}{0.675} \\
 & cp & 1480 & 1.4 & {0.0} & 100.0 & 100.0 & 99.2 & 100.0 & 99.5 & 88.9 & \cellcolor{blue!15}{0.993} \\
 & cp\_cot & 1961 & 2.5 & {0.0} & 100.0 & 99.5 & 97.5 & 99.2 & 97.9 & 100.0 & \cellcolor{blue!15}{0.990} \\
 & cp\_icl & 2964 & 1.3 & {0.0} & 100.0 & 100.0 & 99.2 & 100.0 & 99.5 & 100.0 & \cellcolor{blue!15}{0.998} \\
 & cp\_icl\_cot & 3410 & 2.2 & {0.5} & 99.5 & 99.5 & 98.4 & 100.0 & 99.0 & 100.0 & \cellcolor{blue!15}{0.993} \\
\cmidrule(l){1-12}
\multirow{5}{*}{\shortstack[l]{\textcolor{brown}{\textbf{Gemma4:31b}} \\ \textcolor{brown}{\textbf{(Think)}}}} & baseline & 2918 & 40.9 & {0.0} & 100.0 & 100.0 & 0.0 & 99.2 & 36.1 & 0.0 & \cellcolor{blue!15}{0.674} \\
 & cp & 2541 & 18.2 & {0.5} & 99.5 & 99.5 & 98.4 & 99.2 & 98.4 & 100.0 & \cellcolor{blue!15}{0.991} \\
 & cp\_cot & 3019 & 20.9 & {0.0} & 100.0 & 100.0 & 99.2 & 100.0 & 99.0 & 100.0 & \cellcolor{blue!15}{0.996} \\
 & cp\_icl & 3967 & 16.0 & {0.5} & 99.5 & 99.5 & 97.5 & 99.2 & 97.9 & 100.0 & \cellcolor{blue!15}{0.989} \\
 & cp\_icl\_cot & 4416 & 17.8 & {0.0} & 100.0 & 100.0 & 99.2 & 100.0 & 99.0 & 100.0 & \cellcolor{blue!15}{0.996} \\
\cmidrule(l){1-12}
\multirow{5}{*}{\textcolor{magenta}{\textbf{Gemma4:31b}}} & baseline & 578 & 3.1 & {0.0} & 100.0 & 100.0 & 0.0 & 99.2 & 36.1 & 0.0 & \cellcolor{blue!15}{0.674} \\
 & cp & 1598 & 3.1 & {0.0} & 100.0 & 100.0 & 100.0 & 100.0 & 100.0 & 100.0 & \cellcolor{blue!15}{1.000} \\
 & cp\_cot & 2075 & 5.4 & {0.0} & 100.0 & 100.0 & 99.2 & 100.0 & 99.5 & 100.0 & \cellcolor{blue!15}{0.998} \\
 & cp\_icl & 3219 & 3.1 & {0.0} & 100.0 & 100.0 & 99.2 & 100.0 & 99.5 & 100.0 & \cellcolor{blue!15}{0.998} \\
 & cp\_icl\_cot & 3725 & 5.9 & {0.0} & 100.0 & 100.0 & 100.0 & 100.0 & 100.0 & 100.0 & \cellcolor{blue!15}{1.000} \\
\cmidrule(l){1-12}
\multirow{5}{*}{\shortstack[l]{\textcolor{cyan}{\textbf{Gpt-oss:20b}} \\ \textcolor{cyan}{\textbf{(Think)}}}} & baseline & 1632 & 5.4 & {0.5} & 99.5 & 99.0 & 0.0 & 99.2 & 36.1 & 0.0 & \cellcolor{blue!15}{0.671} \\
 & cp & 2210 & 4.2 & {1.0} & 99.0 & 99.0 & 96.7 & 99.2 & 97.9 & 100.0 & \cellcolor{blue!15}{0.985} \\
 & cp\_cot & 2877 & 6.0 & {1.0} & 99.0 & 99.0 & 96.7 & 99.2 & 97.9 & 100.0 & \cellcolor{blue!15}{0.985} \\
 & cp\_icl & 3567 & 3.8 & {3.1} & 96.9 & 96.9 & 94.3 & 100.0 & 95.8 & 100.0 & \cellcolor{blue!15}{0.966} \\
 & cp\_icl\_cot & 4198 & 5.3 & {1.6} & 98.4 & 98.4 & 95.1 & 100.0 & 96.9 & 100.0 & \cellcolor{blue!15}{0.978} \\
\midrule
\multicolumn{12}{@{}l}{\textit{Frenetix validation (\texttt{gpt-5.4-mini}, \texttt{cp\_icl\_cot}, $N{=}191$): \textbf{155/191 (81.2\%)} runnable (goal\_reached + max\_steps)}} \\
\bottomrule
\end{tabular}
\vspace{1ex}
{\flushleft\footnotesize\textit{Note:} Arrows mark preferred directions ($\uparrow$/$\downarrow$). Tokens and latency (s) are means per query.\par}
\end{table*}

\begin{figure*}[!t]
    \centering
    \includegraphics[width=\textwidth]{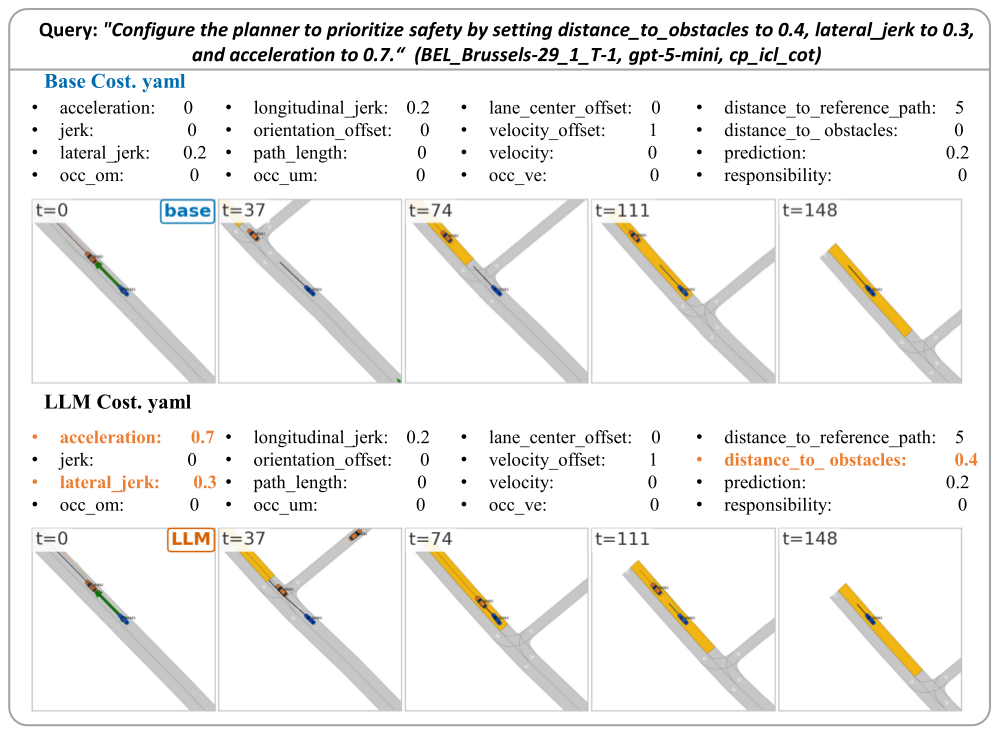}
    \caption{Pure-explicit Planner Testing and Enhancement example (\textsc{q003}, scenario \texttt{BEL\_Brussels-29\_1\_T-1}, \texttt{gpt-5.4-mini} $\times$ \texttt{cp\_icl\_cot}). The three weights modified by the LLM are highlighted in orange in the YAML; five matched-timestep frames are shown for the base and modified rollouts.}
    \label{fig:planner_q003}
\end{figure*}

Table~\ref{tab:planner-all-results} reports the full results matrix ($N{=}191$ scored queries per (model, condition) cell, with the remaining 9 out-of-vocab queries reported separately under \textit{refused \%}. All 50 cells are fully populated). The headline \textit{full-YAML \%} is 36\% across all baseline cells. This is the floor produced by leaving the input \texttt{cost.yaml} untouched, which matches \gls{GT} only on $\sim 36\%$ of queries that did not require a change. Any prompting (\texttt{cp} onwards) lifts almost every model to $\ge 98\%$ full-YAML match. The \textit{refused \%} column reveals that out-of-vocab refusals are the hardest sub-task: only the strongest models hit 100\% under \texttt{cp} alone. Adding \gls{ICL} exemplars (\texttt{cp\_icl}, \texttt{cp\_icl\_cot}) is what eliminates the remaining false-positive YAML emissions for those out-of-vocab queries. The Frenetix-validation footnote shows that 81.2\% of the YAMLs emitted by \texttt{gpt-5.4-mini} under \texttt{cp\_icl\_cot} produced a runnable planner configuration end-to-end. The remaining 18.8\% are scenarios in which the modified cost weights cause the planner to collide before reaching the goal, an additional planning-robustness signal that the YAML-match metric cannot detect. Figure~\ref{fig:planner_q003} shows a representative qualitative example.

\subsection{Cross-Planner Qualitative Comparison}
\label{app:cross-planner-qualitative}

\begin{figure*}[ht]
    \centering
    \begin{subfigure}[t]{0.46\linewidth}
        \includegraphics[width=\linewidth]{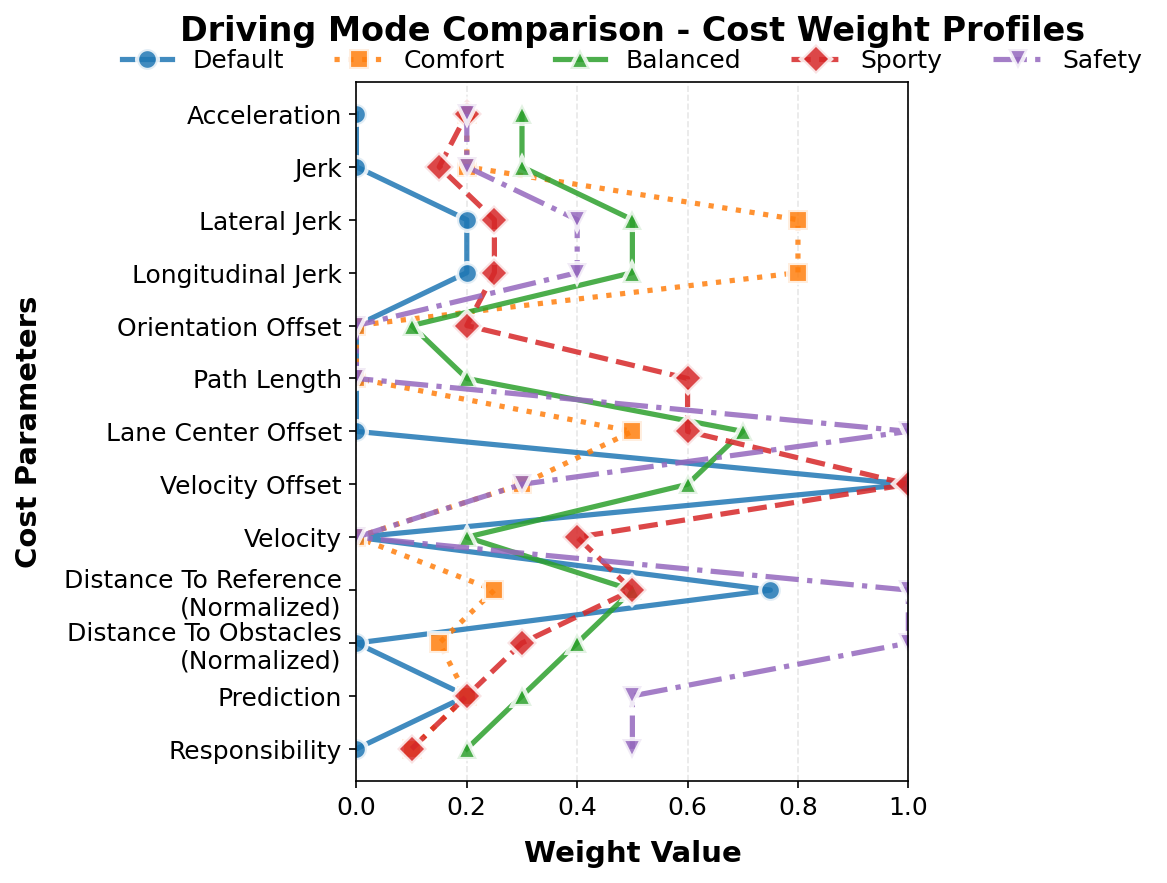}
        \caption{Frenetix --- thirteen cost weights across the five preset modes.}
        \label{fig:cross-planner-qualitative-a}
    \end{subfigure}\hfill
    \begin{subfigure}[t]{0.52\linewidth}
        \includegraphics[width=\linewidth]{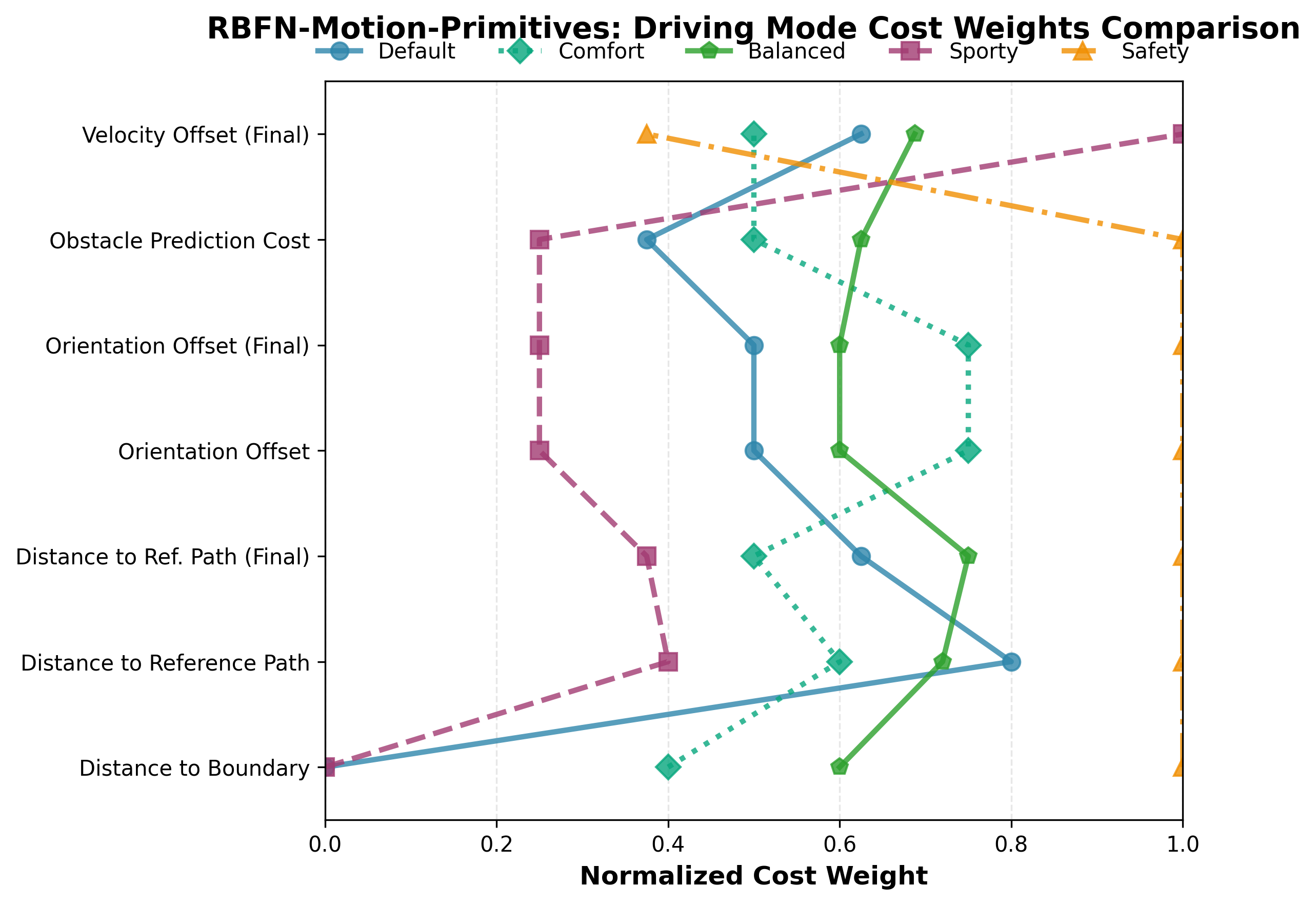}
        \caption{MP-RBFN --- seven normalised cost channels across the five preset modes.}
        \label{fig:cross-planner-qualitative-b}
    \end{subfigure}
    \caption{Qualitative cross-planner comparison of the five driving-mode presets exposed by PlannerForge. Despite the schema difference, both planners realise the same intent space (Sporty $\to$ velocity-up, lower obstacle weight; Safety $\to$ obstacle/prediction maxed; Comfort $\to$ jerk-and-orientation flattened) through their respective cost vocabularies.}
    \label{fig:cross-planner-qualitative}
\end{figure*}

PlannerForge exposes the same five driving-mode presets (\emph{Default}, \emph{Comfort}, \emph{Balanced}, \emph{Sporty}, \emph{Safety}) for both the sampling-based Frenetix planner and the learning-based MP-RBFN planner. Figure~\ref{fig:cross-planner-qualitative} compares each planner's per-mode cost-weight profile. The two planners operate on different cost vocabularies: Frenetix exposes thirteen named weights (acceleration, jerk axes, path-length, lane-centre offset, velocity offset, distances to reference path and to obstacles, prediction, responsibility), while MP-RBFN exposes seven normalised channels keyed on its radial-basis representation (distance-to-boundary, distance-to-reference-path, orientation/velocity offsets, obstacle prediction). Despite the schema differences, the five shared presets produce the same relative shape in both planners: \emph{Sporty} drives the velocity-offset weight up while loosening obstacle prediction. \emph{Safety} drives the obstacle and prediction weights to their maxima; \emph{Comfort} balances orientation and lateral-jerk weights. \emph{Balanced} lies between them. A head-to-head quantitative Frenetix vs MP-RBFN comparison on framework-generated scenarios is part of an ongoing extension. The qualitative correspondence in Figure~\ref{fig:cross-planner-qualitative} establishes that the framework's natural-language preset interface generalises across planner families.
\begin{figure}[ht]
    \centering
    \includegraphics[width=0.55\linewidth]{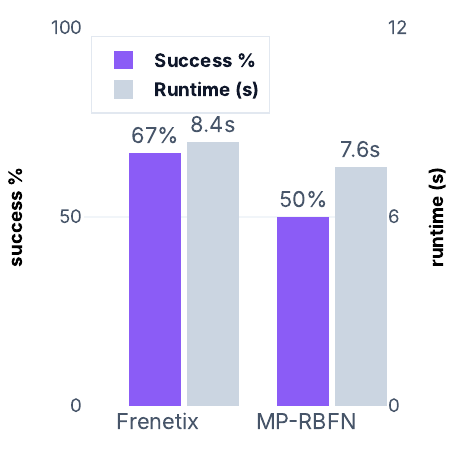}
    \caption{{Cross-planner batch comparison. A single prompt dispatches the
    same scenario batch to both Frenetix and MP-RBFN and returns success rate and
    runtime for each.}}
    \label{fig:cross-planner}
\end{figure}
Figure~\ref{fig:cross-planner} shows the chatbot workflow: a single prompt dispatches both Frenetix and MP-RBFN on $N{=}100$ shared scenarios from the \texttt{batch\_100} preset. The Analysis Module reports success rate and mean runtime side-by-side, and can further diagnose specific failure reasons (e.g., collisions vs.\ timeouts) for each planner. This confirms that the unified $\pi_P$ interface supports efficient, intuitive cross-planner benchmarking.

\subsection{Overview Prompts}
\label{app:overview-prompts}
\label{app:prompt-gen}
\label{app:prompt-sel}
\label{app:prompt-mod-T}
\label{app:prompt-mod-B}
\label{app:prompt-mod-P}
\label{app:prompt-mod-G}
\label{app:prompt-router}
\label{app:prompt-tune}
\label{app:prompt-batch-analysis}

{Each module uses a curated-prompt (\texttt{cp}) header that specifies the system role, the JSON/YAML/XML output schema, and the main constraints. The five prompt conditions (\texttt{baseline}, \texttt{cp}, \texttt{cp\_cot}, \texttt{cp\_icl}, \texttt{cp\_icl\_cot}) share this header; \gls{ICL} and \gls{CoT} variants add demonstrations and reasoning scaffolds on top. Full prompt files for Generation, Selection, Modification (T/B/P/G), Module Router, Planner Testing and Enhancement, and Batch Analysis are released with the code at \url{https://github.com/TUM-AVS/PlannerForge}.}

\end{document}